%% file: main.tex
\documentclass{article}
\usepackage{hyperref}
\usepackage[accepted]{icml2024}
\input{definitions/paper_commands}

\icmltitlerunning{Federated Full-Parameter Tuning of Billion-Sized Language Models with Communication Cost under 18 Kilobytes}

\input{definitions/math_commands}

\input{definitions/symbols}

\begin{document}

\twocolumn[
%\icmltitle{Federated Full-Parameter Fine-Tuning of Billion-sized Large Language Models with Communication Cost under 18 Kilobytes}
% too long, perhaps: 
\icmltitle{Federated Full-Parameter Tuning of Billion-Sized Language Models \\ with Communication Cost under 18 Kilobytes}

\icmlsetsymbol{intern}{*}

\begin{icmlauthorlist}
\icmlauthor{Zhen Qin}{zju,intern}
\icmlauthor{Daoyuan Chen}{ali}
\icmlauthor{Bingchen Qian}{ali}
\icmlauthor{Bolin Ding}{ali}
\icmlauthor{Yaliang Li}{ali}
\icmlauthor{Shuiguang Deng}{zju}

\end{icmlauthorlist}

\icmlaffiliation{zju}{College of Computer Science and Technology, Zhejiang University, Hangzhou, China}
\icmlaffiliation{ali}{Alibaba Group}

\icmlcorrespondingauthor{Yaliang Li}{yaliang.li@alibaba-inc.com}
\icmlcorrespondingauthor{Shuiguang Deng}{dengsg@zju.edu.cn}

\icmlkeywords{Federated Learning, Large Language Model, Instruction Tuning, Communication Cost, Full-Parameter Fine-Tuning}

\vskip 0.3in
]

\printAffiliationsAndNotice{\textsuperscript{*} Work done as an intern at Alibaba Group.} 

\input{sections/0_abstract}
\input{sections/1_introduction}
\input{sections/2_related}
\input{sections/3_formulation}
\input{sections/4_approach}
\input{sections/5_exp}
\input{sections/6_conclusion}

\newpage
\section*{Acknowledgements}
This work was supported in part by the National Science Foundation of China under Grants 62125206 and U20A20173 and in part by the Key Research Project of Zhejiang Province under Grant 2022C01145.

\section*{Impact Statement}
This paper presents work whose goal is to advance the field of Machine Learning. There are many potential societal consequences of our work, none which we feel must be specifically highlighted here.

\bibliography{main}
\bibliographystyle{icml2024}

\newpage
\appendix
\onecolumn

\input{sections/7_0_toc_appendix}
\input{sections/7_appendix_technical}
\input{sections/8_appendix_notations}
\input{sections/9_appendix_algorithm}
\input{sections/10_appendix_assumption}
\input{sections/11_appendix_detailed_proofs}
\input{sections/12_appendix_another_explanation}
\input{sections/13_appendix_implementation}
\input{sections/14_appendix_supplementary_exp}
\input{sections/15_appendix_calculation_overhead}
\input{sections/16_appendix_discussion}

\end{document}

%% file: definitions/paper_commands.tex
\usepackage{microtype}
\usepackage{graphicx}
\usepackage{subfigure}
\usepackage{booktabs} % for professional tables

\makeatletter
\def\UrlAlphabet{%
      \do\a\do\b\do\c\do\d\do\e\do\f\do\g\do\h\do\i\do\j%
      \do\k\do\l\do\m\do\n\do\o\do\p\do\q\do\r\do\s\do\t%
      \do\u\do\v\do\w\do\x\do\y\do\z\do\A\do\B\do\C\do\D%
      \do\E\do\F\do\G\do\H\do\I\do\J\do\K\do\L\do\M\do\N%
      \do\O\do\P\do\Q\do\R\do\S\do\T\do\U\do\V\do\W\do\X%
      \do\Y\do\Z}
\def\UrlDigits{\do\1\do\2\do\3\do\4\do\5\do\6\do\7\do\8\do\9\do\0}
\g@addto@macro{\UrlBreaks}{\UrlOrds}
\g@addto@macro{\UrlBreaks}{\UrlAlphabet}
\g@addto@macro{\UrlBreaks}{\UrlDigits}

\usepackage{amsmath}
\usepackage{amssymb}
\usepackage{mathtools}
\usepackage{amsthm}
\usepackage[linesnumbered,ruled,noend]{algorithm2e}
\usepackage{tabularx}
\usepackage{multirow}
\usepackage{rotating}
\usepackage{pifont}
\usepackage{bbding} % serial number
\usepackage{enumitem}
\usepackage{enumerate}
\usepackage{xspace}
\usepackage{makecell}
\usepackage{colortbl}

\usepackage{soul}
\definecolor{lightgray}{gray}{0.88}
\sethlcolor{lightgray}

\usepackage[capitalize,noabbrev]{cleveref}

\newtheorem{theorem}{Theorem}
\newtheorem{assumption}{Assumption}

\newtheorem{principle}{Principle}
\newtheorem{lemma}{Lemma}

\newtheorem{definition}{Definition}

\newcommand{\apppro}{FedKSeed-Pro\xspace}
\newcommand{\app}{FedKSeed\xspace}
\newcommand{\datani}{Natural Instructions\xspace}
\newcommand{\datadolly}{Dolly-15K\xspace}
\newcommand{\modeldatajuicer}{DataJuicer-1.3B\xspace}
\newcommand{\modelllama}{LLaMA-3B\xspace}

\newcommand{\texl}[1]{\underline{#1}}
\newcommand{\texb}[1]{\textbf{#1}}
\usepackage[textsize=tiny]{todonotes}

\newcommand{\para}[1]{\textbf{#1}}

%% file: definitions/math_commands.tex
\usepackage{bm, dsfont}

\let\hat\widehat

\newcommand{\gb}{\mathbf{g}}

\newcommand{\pb}{\mathbf{p}}

\newcommand{\wb}{\mathbf{w}}
\newcommand{\xb}{\mathbf{x}}

\newcommand{\zb}{\mathbf{z}}

\newcommand{\Ib}{\mathbf{I}}

\newcommand{\cA}{\mathcal{A}}

\newcommand{\cD}{\mathcal{D}}

\newcommand{\cG}{\mathcal{G}}

\newcommand{\cL}{\mathcal{L}}

\newcommand{\cN}{\mathcal{N}}
\newcommand{\cO}{\mathcal{O}}

\newcommand{\CC}{\mathbb{C}}

\newcommand{\EE}{\mathbb{E}}
\newcommand{\GG}{\mathbb{G}}
\newcommand{\HH}{\mathbb{H}}

\newcommand{\RR}{\mathbb{R}}
\newcommand{\SSS}{\mathbb{S}}

\newcommand{\ZZ}{\mathbb{Z}}

\newcommand{\F}{\mathrm{F}}

\newcommand*{\E}{\mathbb{E}}

\newcommand{\eps}{\epsilon}

\makeatletter
\newcommand{\ve}{\@ifnextchar\bgroup{\velong}{{\bm{e}}}}
\newcommand{\velong}[1]{{\bm{#1}}}
\makeatother

%% file: definitions/symbols.tex
\newcommand{\Z}{\mathbb{Z}}
\newcommand{\N}{\mathbb{N}}

%% file: sections/0_abstract.tex
\begin{abstract}
Pre-trained large language models (LLMs) need fine-tuning to improve their responsiveness to natural language instructions. 
Federated learning offers a way to fine-tune LLMs using the abundant data on end devices without compromising data privacy.
Most existing federated fine-tuning methods for LLMs rely on parameter-efficient fine-tuning techniques, which may not reach the performance height possible with full-parameter tuning. 
However, federated full-parameter tuning of LLMs is a non-trivial problem due to the immense communication cost.
This work introduces \app that employs zeroth-order optimization with a finite set of random seeds.
It significantly reduces transmission requirements between the server and clients to just a few random seeds and scalar gradients, amounting to only a few thousand bytes, making federated full-parameter tuning of billion-sized LLMs possible on devices.
Building on it, we develop a strategy enabling probability-differentiated seed sampling, prioritizing perturbations with greater impact on model accuracy.
Experiments across six scenarios with various LLMs, datasets and data partitions demonstrate that our approach outperforms existing federated LLM fine-tuning methods in both communication efficiency and zero-shot generalization. 
\end{abstract}

%% file: sections/1_introduction.tex
\section{Introduction}
\label{sec-introduction}
Large language models (LLMs) exhibit outstanding performance on various natural language tasks yet require fine-tuning to enhance their task responsiveness \cite{chen2023position,dong2023towards}. 
While existing open datasets contribute to LLM tuning \cite{supernaturalinstructions,wei2022FLAN}, the vast quantities of private data continuously generated at end devices present an untapped opportunity for further exploitation, especially as the reservoir of high-quality language data may become depleted in the future \cite{villalobos2022will}.
Federated learning (FL) \cite{mcmahan2017communication,Kairouz2021federated} offers a privacy-protected way to collaboratively tune LLMs with distributed data, which has been explored by recent parameter-efficient fine-tuning (PEFT) based works \cite{zhang2023fedit,babakniya2023slora,zhang2023-FedPETuning,che2023FedPepTAO}.
Nonetheless, PEFT is not a universal solution for LLM tuning, as it may not consistently match the accuracy of full-parameter tuning \cite{chen2022revisiting,pu2023empirical,sun2023comparative}, particularly in FL scenarios where the statistically heterogeneous client data diminish the effectiveness of PEFT \cite{babakniya2023slora,zhang2023-FedPETuning}. 
Considering full-parameter tuning's potential for higher accuracy, exploring its feasibility to LLMs with FL is promising.

\begin{table}[t]
  \renewcommand\arraystretch{1.06}
  \caption{Comparing federated tuning methods w.r.t. \hl{accuracy} and client-side costs, with \textit{computation cost} referring to that incurred by obtaining the latest model, 
  $d$ as the model parameter count,  
  $\nu$ as the ratio of trainable parameters in PEFT versus full-parameter tuning, 
  $\tau$ as the average number of local steps performed by each client per round,
  $r$ as the number of communication rounds, 
  and $m$ as the number of active clients in each round. 
  $M_{\text{infer}}$, $M_{\text{peft}}$ and $M_{\text{full}}$ are peak memory usage for inference, PEFT with BP, and full-parameter tuning with BP, respectively. 
  For simplicity, we denote $\xi=M_{\text{peft}}/M_{\text{infer}}$ and $\Xi=M_{\text{full}}/M_{\text{infer}}$.
  Generally, $\nu \ll  1 < \xi < \Xi \ll \tau rm$, and $d$ is in billions for LLMs. \app delivers top-tier performance across these aspects simultaneously.
  }
  \label{tab-qualitative-comparison}
  \setlength\tabcolsep{1.1pt}
  \centering
  \small
  \begin{tabularx}{\linewidth}{l|r|r|r|r}
    \toprule[1.0pt]
    Approach                                  &\cellcolor{gray!20}Acc.\small{$\uparrow $}  & Commu.\small{$\downarrow $} & Mem.\small{$\downarrow $}& Comput.\small{$\downarrow $} \\
    \midrule[1.0pt]
            PEFT with BP                                     &\cellcolor{gray!20}$\star $         &$\cO(\nu d)$    &$\cO(\xi d)$  &$\cO(d)$\\
    \cline{1-5}
        Full-param. with \small{BP}                    &\cellcolor{gray!20}$\star\star$&$\cO(d)$                    & $ \cO(\Xi d)$ & $\cO(d)$         \\
    \cline{1-5}
    Full-param. with \small{ZOO}                   &\cellcolor{gray!20}$\star \star$    &$\cO(d)$                    & $\cO(d)$         & $\cO(d)$         \\
    \ \small{ infinite seed-pool in uplink}&\cellcolor{gray!20}$\star\star$     &$\cO(d)$                    & $\cO(d)$         & $\cO(d)$         \\
    \ \small{ infinite seed-pool in bi-link}&\cellcolor{gray!20}$\star\star$     &$\cO(1)$                    & $\cO(d)$         & $\cO(\tau rm d)$      \\
    \midrule[1.0pt]
    \app ($|$seed-pool$|$=$K$)                                     &\cellcolor{gray!20}$\star\star $   & $\cO(1)$                    & $\cO(d)$         & $ \cO(d)$ \\
    \bottomrule[1.0pt]
  \end{tabularx}
\end{table}

However, full-parameter tuning of billion-sized LLMs with FL on devices is impractical with current technology, as backpropagation (BP) and most BP-free methods, such as zeroth-order optimization (ZOO) \cite{fang2022communication}, incur communication costs that scale with model size, as shown in Table \ref{tab-qualitative-comparison}. 
These costs become prohibitive for billion-sized LLMs. 
In particular, BP-based approaches also require significant memory that is not feasible for most end devices, e.g., tuning a full LLM with 1.3 billion parameters may consume over 20GB of memory \cite{malladi2023mezo}.

\begin{figure}
  \centering
  \begin{minipage}[t]{0.416\linewidth}
    \centering
    \includegraphics[width=\linewidth]{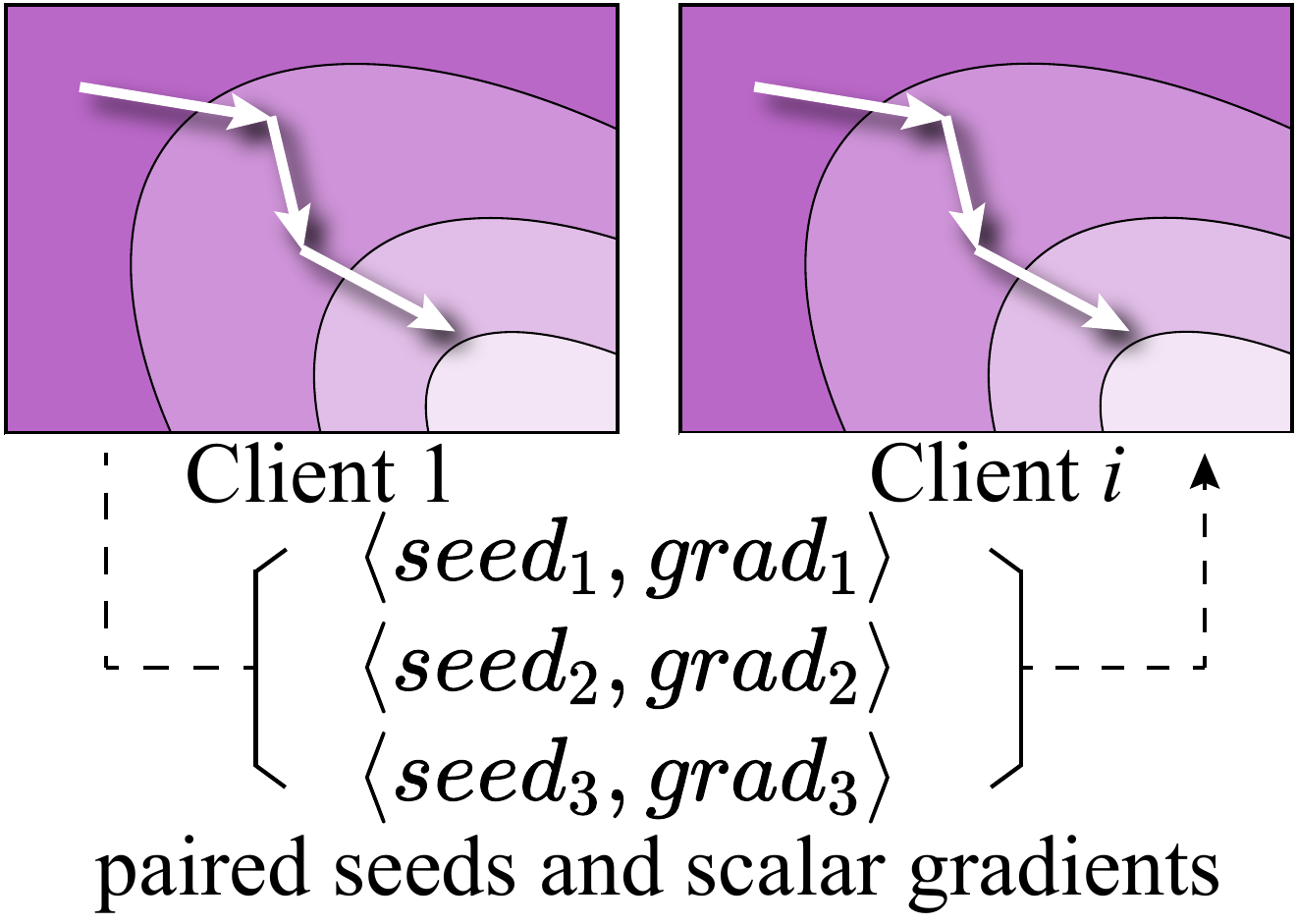}
    \caption{Each step of ZOO can be replicated by 1) a random seed that is used to generate a perturbation, and 2) a scalar gradient on it.}
    \label{pic-intro-seed-transmission}
  \end{minipage}
  \hspace{0.1cm}
  \begin{minipage}[t]{0.465\linewidth}
    \centering
    \includegraphics[width=\linewidth]{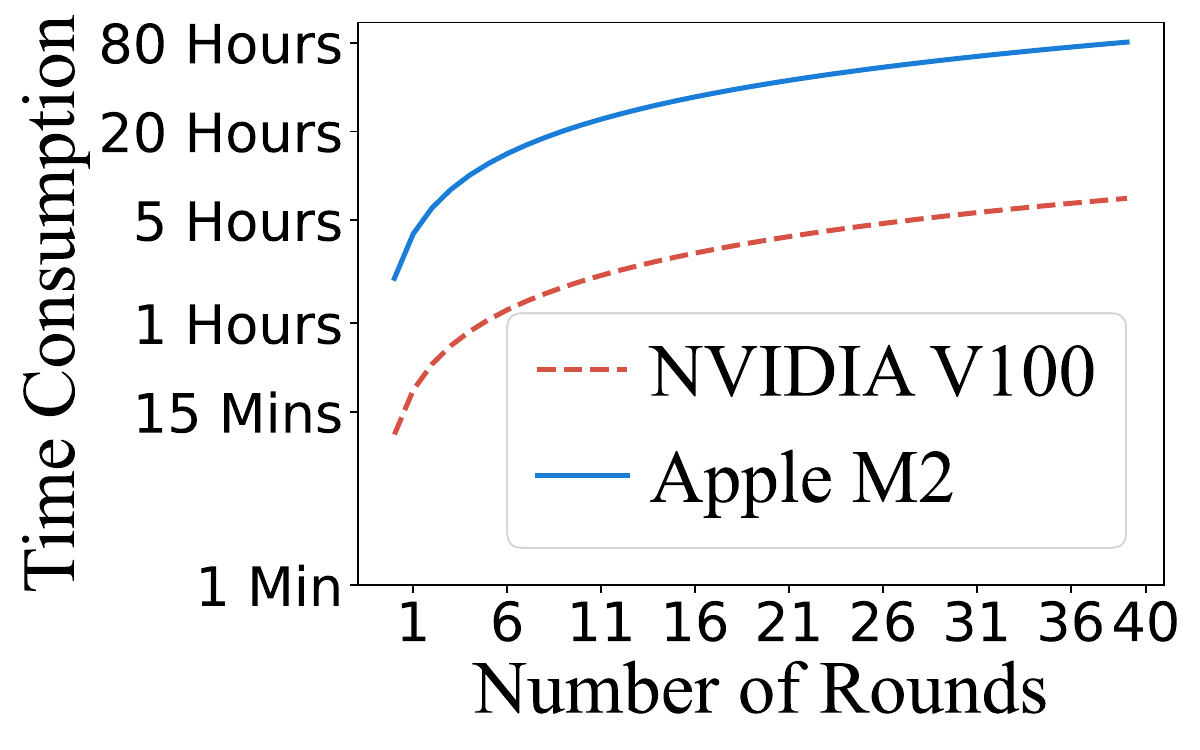}
    \caption{With more total steps, the time required to compute the latest global model by update replication grows rapidly (calculated with \modelllama).
    }
    \label{pic-intro-time-consumption}
  \end{minipage}
\end{figure}

We note an interesting property in ZOO: a parameter update step of ZOO can be replicated with just two values: a seed (with an identical random number generator) and the corresponding scalar gradient (the product of the scalar gradient and the perturbation yields the vector gradient).
Some recent ZOO-based FL methods \cite{zelikman2023onebyte,feng2023baffle,maritan2023fedzen} explore this property to reduce communication cost as shown in Figure \ref{pic-intro-seed-transmission}, however, they compromise other performance factors, making them still impractical for LLMs.
As outlined in Table \ref{tab-qualitative-comparison}, current methods either 
(1) optimize the uplink communication for clients but neglect the significant downlink cost to distribute the latest global model in each round \cite{feng2023baffle,maritan2023fedzen}, or 
(2) optimize bi-link communication but require each client to replicate all update steps from the others to synchronize the latest global model, leading to a computation cost that increases indefinitely with the number of rounds \cite{zelikman2023onebyte}, as shown in Figure \ref{pic-intro-time-consumption}.

To achieve the best of both worlds, i.e., avoiding the massive \textit{communication cost} associated with \textit{transmitting full model parameters} while limiting the ever-increasing \textit{computation cost} of \textit{syncing} to the \textit{latest global model}, this work introduces a novel federated full-parameter tuning approach for LLMs, based on ZOO with only $K$ random seeds (denoted as \app).
It employs a theoretically informed paradigm of seed reuse, implementing federated tuning with a finite set of seeds to generate perturbations, thus enabling full-parameter tuning of LLMs in FL with a communication cost of less than 18 kilobytes per round, and the memory footprint equivalent to inference requirements.
Building on \app, we introduce a strategy to assess the significance of perturbations, assigning varied sampling probabilities to candidate seeds. 
It narrows the seed pool to expedite the syncing to the latest global model, thereby further enhancing both computational efficiency and model accuracy.

Our main contributions are summarized as follows:
\begin{itemize}[leftmargin=1em]
    \item We propose a novel federated full-parameter tuning approach for LLM based on ZOO, \app, which transmits only $K$ seeds and corresponding scalar gradients between the server and clients.
    To the best of our knowledge, this is the first work to make full-parameter tuning of billion-sized LLMs feasible on federated devices, with a communication cost of less than 18 kilobytes per round. 
    \item We investigate the differentiated importance of ZOO perturbations, and propose a simple yet effective strategy that samples seeds with non-uniform probabilities. 
    It improves accuracy while reducing the cardinality of candidate seeds needed by \app, thereby accelerating the client-side synchronization with the latest global model.
    \item Experiments on 6 scenarios with various LLMs, datasets and data partitions show that \app with the proposed non-uniform seed sampling attains an average relative improvement of 7.26\% in Rouge-L over the best-performing practical baseline on held-out tasks and reduces communication costs by a factor of around a thousand. 
    Our codes are publicly available at \url{https://github.com/alibaba/FederatedScope/tree/FedKSeed}.
\end{itemize}

%% file: sections/2_related.tex
\section{Related Work}
\label{sec-related}
\textbf{Federated Fine-Tuning for LLMs.}
There are some studies exploring fine-tuning LLMs with FL based on PEFT techniques, e.g., \citet{zhang2023-FedPETuning} provide benchmarks for PEFT techniques in FL.
Among existing PEFT techniques, LoRA \cite{hu2022lora} is usually preferable. 
\citet{zhang2023fedit} proposes a federated instruction tuning approach based on LoRA. 
\citet{jiang2023low-parameter} design a low-parameter FL approach based on LoRA for text classification.
\citet{babakniya2023slora} experimentally demonstrate that when facing FL with non-IID data, LoRA is not as good as full-parameter tuning and propose a strategic initialization of LoRA weights based on SVD decomposition of full parameters fine-tuned with BP.
There are also some works contributing to the deployment of LLM tuning with FL, e.g., FederatedScope-LLM \cite{kuang2023federatedscope-LLM} and FATE-LLM \cite{fan2023-FATE-LLM}.
The computational bottlenecks have been thoroughly investigated by \citet{woisetschlager2023very-edge}.

\textbf{Federated Learning with Zeroth-Order Optimization.}
There are some researches using ZOO for non-differentiable problems \cite{li2021communication}.
\citet{shu2023Trajectory} boost the query efficiency of ZOO in FL by optimization trajectory. 
Some researches analyze the convergence and generalization of ZOO-based FL \cite{fang2022communication,chen2023fine}.
However, these approaches are only validated for small models with no more than 10 million parameters.

There are also some works leveraging random seeds to optimize communication efficiency. 
However, they are not suitable for full-parameter tuning of LLMs with FL due to 
(1) distributing the latest model parameters in each round \cite{xu2023billion-sized,maritan2023fedzen,feng2023baffle} that hinders the important download efficiency of clients \cite{dorfman2023docofl}, 
or 
(2) tremendous computation overhead for calculating the latest model \cite{zelikman2023onebyte} as in Figure \ref{pic-intro-time-consumption}, 
or 
(3) the reliance on BP which consumes a substantial amount of memory \cite{rahimi2023evofed}.

\para{Difference from Related Works.} 
A recent work FwdLLM \cite{xu2023billion-sized} conducts FL based on PEFT and ZOO, but with the goal and techniques different from \app.
FwdLLM uses quantization and PEFT to reduce memory cost, while we mainly focus on communication cost and enable full-parameter tuning of LLMs with FL. 
\app is orthogonal to quantization techniques \cite{xi2023-4bit,dettmers2023qlora}. 
FwdLLM requires a total of several hundred GB of communication cost to tune an LLM with only about 300 million parameters, as it only optimizes client-side uplink. 
Besides, although some BP-based methods optimize the communication \cite{rahimi2023evofed}, they are not tailored and are not suitable for LLM tuning on end devices due to the tremendous memory footprint.

In a nutshell, existing works mainly focus on tuning partial LLM parameters, while our method enables full-parameter tuning of LLMs with FL, obtaining higher accuracy (Table \ref{tab-performance}).
\app significantly cuts the communication and memory costs by eliminating model parameter transmission and BP, outperforming existing approaches tailored for federated LLM tuning (Table \ref{tab-overhead}).  
Further technical comparisons between \app and existing works are in Appendix \ref{sec-appendix-related}. 

%% file: sections/3_formulation.tex
\section{Problem Formulation}
\label{sec-formulation}
Consider an FL system with $N$ clients, each with a private dataset $\cD_i$, federated fine-tuning aims at collaboratively tuning model $\wb \in \RR^d$ with the pre-trained weight $\wb^0 \in \RR^d$ at initialization, which can be formulated as
\begin{equation}
  \min_{\wb \in \RR^d} f(\wb) \triangleq \sum_{i=1}^N c_i \cdot \EE_{\xb \sim \cD_i}\left[ \cL(\wb; \xb)\right],
  \label{eq-fl-optimization}
\end{equation}
where $\cL(\wb;\xb)$ is the loss evaluated at model $\wb$ on a data instance $\xb$ drawn from $\cD_i$ and $c_i \geq 0$ is the aggregate weight with $\sum_{i=1}^N\! c_i \!=\! 1$. 
Here we utilize $\xb$ since we set the batch size to 1 to lower memory cost as \citet{malladi2023mezo}.
The fundamental distinction between federated fine-tuning and vanilla FL \cite{mcmahan2017communication} is that it begins optimization from a pretrained weight $\wb^0$ rather than from scratch. 
Equation \eqref{eq-fl-optimization} is solved in several rounds of local training and aggregation. 
In round $r$ of BP-based FL \cite{mcmahan2017communication}, each client $i$ performs several steps of gradient descent algorithms on its local model $\wb^r_{i}$ initialized by weight $\wb^r$ downloaded from the server, as
\begin{equation}
  \wb^r_{i, t + 1}= \wb^r_{i, t} - \eta \cdot \gb^r_{i,t}
  \label{eq-local-sgd}
\end{equation}
where $\wb^r_{i, t}$ is the local model of client $i$ at local step $t$, $\eta$ is the learning rate, and $\gb^r_{i,t}$ represents the gradient computed as$ \nabla_{\wb^r_{i, t}} \cL_i(\wb^r_{i, t}; \xb), \forall \xb \in \cD_i$.
After local training, the server aggregates all received $\wb^r_{i}$ for subsequent rounds.

The main difference between ZOO-based FL and BP-based FL lies in the obtaining of the gradient.
ZOO-based FL estimates the gradient by forward propagations instead of direct calculation.
Our work uses the ZOO with a two-point gradient estimator proposed by \citet{malladi2023mezo}, as
\begin{equation}
  \hat\gb^r_{i,t} \!\triangleq\! \frac{\cL(\wb^r_{i,t} \!+\! \epsilon\zb;\xb) \!-\! \cL(\wb^r_{i,t} \!-\! \epsilon\zb;\xb)}{2\epsilon}\zb \approx \zb\zb^\top \gb^r_{i,t},
  \label{eq-mezo}
\end{equation}
where $\hat{\gb}_{i,t}$ is the estimated gradient, $\zb \in\RR^d$ is a random perturbation that follows $\zb\sim\cN(\mathbf{0},\Ib_d)$ and $\epsilon$ is the scale of perturbations. 
When $\hat\gb^r_{i,t}$ is estimated, client $i$ updates its local model as Equation \eqref{eq-local-sgd}.
For symbol convenience, we denote $\hat{\gb} = \hat{\varrho} \cdot \zb$ with omitted scripts, where $\hat{\varrho} \triangleq \frac{\cL(\wb + \epsilon\zb;\xb) - \cL(\wb - \epsilon\zb;\xb)}{2\epsilon}$ is termed as \textbf{\emph{scalar gradient}}.

%% file: sections/4_approach.tex
\section{The proposed \app}
\label{sec-Approach}
\subsection{Overview}
\begin{figure*}[t]
  \centering
  \includegraphics[width=0.83\linewidth]{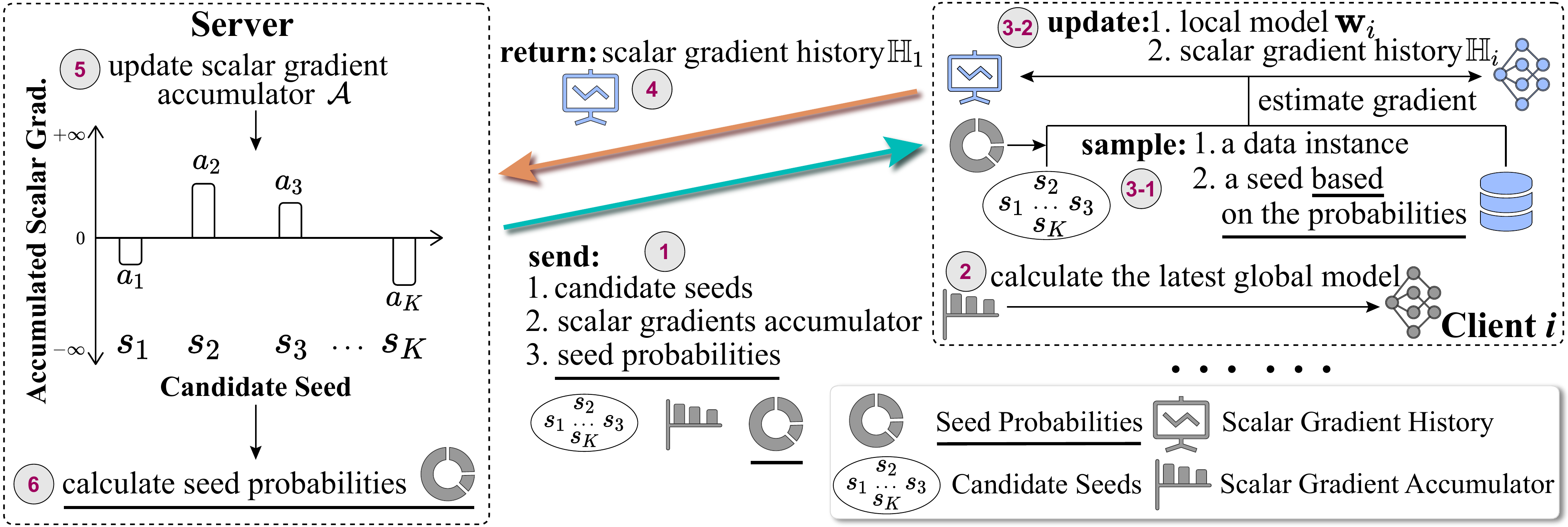}
  \caption{
    Overview of \app, where the serial numbers indicate processes in each round. 
    Gray components share identical values among all clients.
    The \underline{underlined} components are only required by an enhanced version of it, i.e., \underline{\apppro} (Section \ref{subsec-approach-bias}).
  }
  \label{pic-framework}
\end{figure*}

\app is designed with the following goals: (1) to avoid the massive communication cost for transmitting full-model parameters, and (2) to avoid the tremendous memory footprint caused by BP. 
We design \app based on ZOO, and propose a theoretically-informed paradigm that enables seed reuse to limit the ever-increasing computational cost of clients to catch up to the latest global model.

Figure \ref{pic-framework} outlines \app, where the server maintains $K$ unique candidate seeds $\SSS \in \ZZ^K$ and a scalar gradient accumulator $\cA\in \RR^K$ recording the sum of received scalar gradients grouped by corresponding candidate seeds.
Note that the server holds no model parameters, and each client possesses a pretrained LLM $\wb^0$.
The processes are as follows: 
\ding{192} At the start of each round, the server sends $\SSS$ and $\cA$ to active clients. 
\ding{193} Each client $i$ calculates the latest global model as its local model $\wb_i$ based on $\cA$.
\ding{194} A loop of local training, where in each step, the client samples a seed $s_j$ from $\SSS$ and a data instance, then computes the scalar gradient $\hat{\varrho}_j$.
Next, $\wb_i$ is updated based on $\hat{\varrho}_j$ and $s_j$, and $(s_j, \hat{\varrho}_j)$ is staged to the scalar gradient history $\HH_i$.
\ding{195} Each client sends $\HH_i$ to the server after several steps of local training.
\ding{196} The server updates $\cA$ based on all received scalar gradient histories. 
We summarize the processes in Algorithm \ref{algo:main_algo} in Appendix \ref{sec-appendix-algorithm} and detail them in the subsequent sections.

\subsection{Federated Full-Parameter Tuning by Limited Seeds}
\label{subsec-enabling-seed-pool}
Recall Equation \eqref{eq-mezo}, if clients use the same pseudo number generator, a perturbation can be encoded by a random seed (one integer), so $t$ steps of update can be replicated with a scalar gradient history $\HH = \left\{( s_j, \hat{\varrho}_{j} ) \right\}^t$ containing $t$ pairs of seeds and scalar gradients. 
Therefore, an intuitive solution to alleviate model transmission is to have the server track the scalar gradients of all clients \cite{zelikman2023onebyte}.
Assuming $m$ clients participate in FL in each round, and each one conducts average $\tau$ steps of local training. 
After $r$ rounds, a client has to perform $ \tau r m $ steps of model updating to get the latest global model from initial weight $\wb^0$.
From Figure \ref{pic-intro-time-consumption}, when $m = 50$, $\tau = 200$, and $r = 30$, this operation requires over 60 hours with an Apple M2, and over 5 hours even with an NVIDIA V100.
Besides, just 1000 steps of updating consume 8\% of a MacBook M2's battery (battery health at 92\%). 
Thus, it is very important to reduce the computational overhead of computing the latest model.

\para{Restrict the Cardinality of Candidate Seeds: from Infinite to $\bm{K}$.}
If seeds are reused, the update steps needed to get the latest model can be merged. 
If we select only $K$ candidate seeds and accumulate the scalar gradients corresponding to the same seed, \emph{each client only needs to perform at most $K$ iterations to get the latest global model}, unlike the solutions with infinite seeds (as outlined in Table \ref{tab-qualitative-comparison}).

At the start, the server randomly samples $K$ unique candidate seeds $\SSS = \left\{ s_1, s_2, \ldots, s_K\right\}$, and initializes a scalar gradient accumulator $\cA = \left\{a_1, \ldots, a_K\right\} \in \RR^K$, where $a_j = \sum_{\hat{\varrho} \in \cG_j} \hat{\varrho}$, and $\cG_j$ collects all scalar gradients $\hat{\varrho}_{j}$ in $\HH$ on the perturbation of $s_j$.
Each client downloads $\cA$ and gets the latest global model as its local model $\wb_i$ by performing
\begin{equation}
  \wb_i = \wb^0 - \eta \cdot \sum_{j=1}^K a_j \cdot \zb_{j}.
  \label{eq-obtain-latest-model}
\end{equation}
Then, the latest global model $\wb$ is treated as the local model $\wb_i$. 
During each step of local training, the client samples a data instance $\xb$ and a seed $s_j \in \SSS$, and calculates $\hat{\varrho}_{j}$ as
\begin{equation}
  \hat{\varrho}_{j} = \frac{\cL(\wb_i + \epsilon\zb_j;\xb) - \cL(\wb_i - \epsilon\zb_j;\xb)}{2\epsilon}.
  \label{eq-local-calculate-grad}
\end{equation}
Then, the local model $\wb_i$ is updated as
\begin{equation}
  \wb_i \leftarrow \wb_i - \eta \hat{\varrho}_{j} \cdot \zb_j,
  \label{eq-local-update-model}
\end{equation}
and $s_j$ and $\hat{\varrho}_{j}$ are tracked in $\HH_i = \left\{(s_j, \hat{\varrho}_j), \ldots\right\}$.
After $\tau$ steps of local training, $\HH_i$ is sent to the server.
Then, for each $(s_j, \hat{\varrho}_j) \in \HH_i$, the server conducts 
\begin{equation}
  a_j = a_j + c_i \cdot \hat{\varrho}_j,
  \label{eq-aggregate-history}
\end{equation}
to aggregate the gradient history of client $i$ into $\cA$.

Considering that $K$ is a predefined constant, this paradigm shift reduces the computation complexity of obtaining the latest global model to $\cO(d)$, which remains constant throughout the progression of FL. 
Note that the server, just like any client, can obtain the latest global model using Equation \eqref{eq-obtain-latest-model}.
We refer to the above approach as \app. 

\subsubsection{Theoretical Support for Seed Reuse}
\label{subsubsec-theoretical-support}
The effectiveness of seed reuse lies in the minimal impact on convergence when the size of the seed pool ($K$) is not too small, and the total number of update steps is relatively large.
To support this point, we analyze \emph{the convergence similarity between \app and Federated Zeroth-order Optimization (FedZO)} \cite{fang2022communication}, an FL approach based on ZOO without limited range on seed sampling, whose convergence has been proved. 
\begin{definition}
  (Gradient estimation of FedZO).
  Given $\zb$ as i.i.d. random perturbation with $\cN(\mathbf{0},\Ib_d)$ distribution, FedZO estimates gradients in a mini-batch manner as
  \begin{equation}
   \!\hat{\gb}^r_{i,t} \!=\! \frac{1}{b_1 b_2} \!\sum_{b=1}^{b_1} \!\sum_{j=1}^{b_2} \!\frac{\left[\cL(\wb^r_{i,t} \!+\! \epsilon\zb_{j};\xb_{b}) \!-\! \cL(\wb^r_{i,t};\xb_{b})\right]}{\epsilon} \!\zb_{j}. \!\!
    \label{eq-zeroth-one-point}
  \end{equation}
\end{definition}

\begin{figure}
  \centering
  \begin{minipage}[t]{0.48\linewidth}
    \centering
    \includegraphics[width=\linewidth]{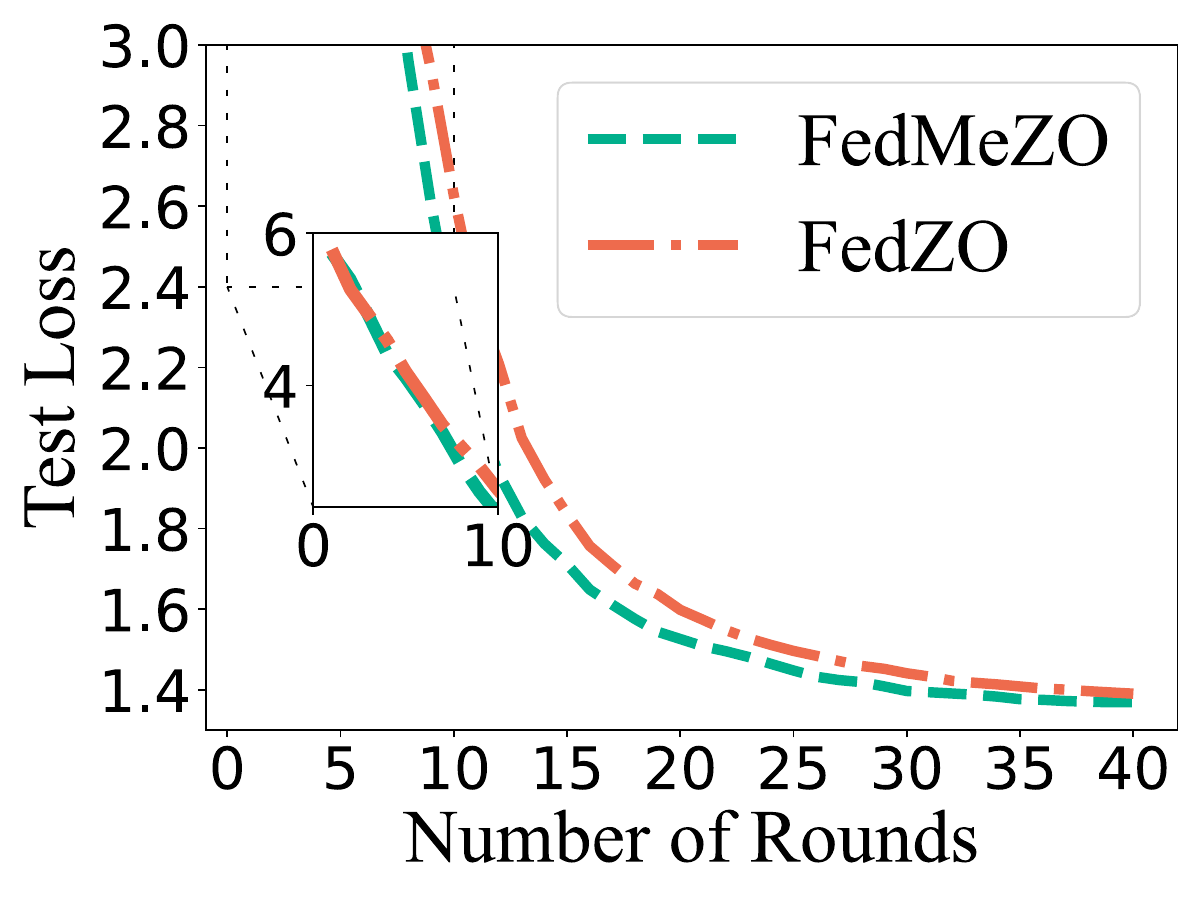}
    \caption{Full-parameter tuning convergence of \modelllama on \datani by FedZO ($b_1 \!\!=\!\! b_2 \!\!=\!\! 1$) and FedMeZO.}
    \label{pic-converge-fedzo-fedmezo}
  \end{minipage}
  \hspace{0.1cm}
  \begin{minipage}[t]{0.475\linewidth}
    \centering
    \includegraphics[width=\linewidth]{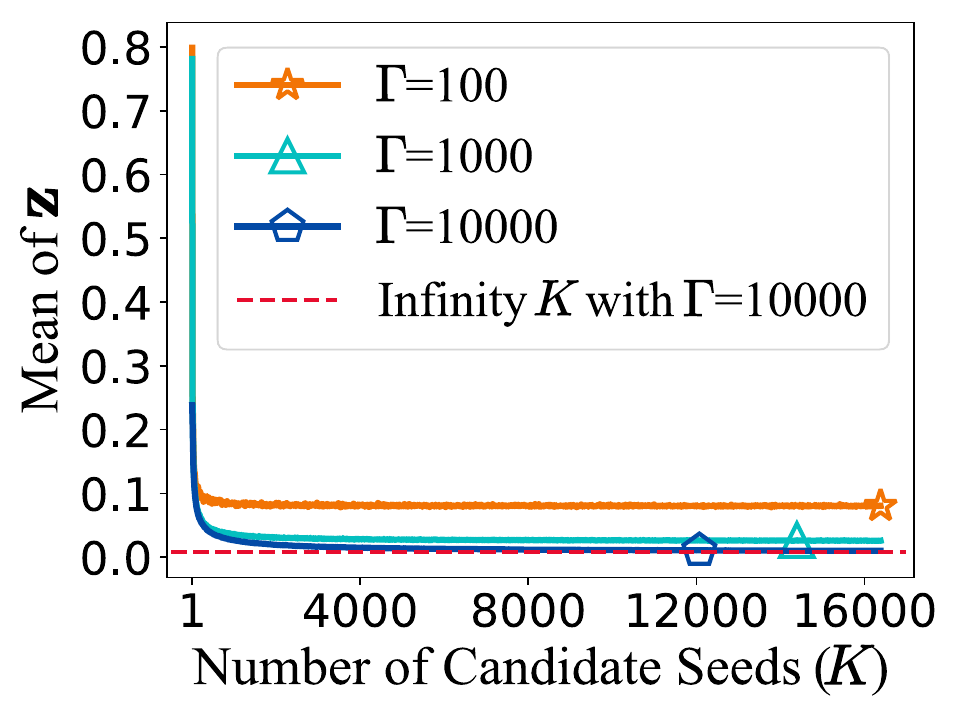}
    \caption{The mean (absolute value) of the $\Gamma$ perturbations randomly sampled with the $K$ candidate random seeds.}
    \label{pic-mean-perturbation}
  \end{minipage}
\end{figure}

\begin{theorem}
  (Convergence of FedZO.) 
  With the assumptions made by \citet{fang2022communication} on (\ref{assumption-loss-boundary}) loss boundary, (\ref{assumption-L-smooth}) $L$-smoothness of objective and loss functions, (\ref{assumption-second-order-moment}) the second-order gradient moment boundary and (\ref{assumption-gradient-dissimilarity-boundary}) local-global gradient dissimilarity boundary, FedZO satisfies
  \begin{equation}
    \min_{r\in \left\{1, \ldots, T \right\}} \EE \! \left \| \nabla f(\wb^r) \right \|^2 \!\leq\! \cO \!\left( \sqrt{\frac{d}{\tau mTb_1 b_2}} \!+\! \sqrt{\frac{b_1b_2\tau}{dmT}} \right),
    \label{eq-zoo-one-point-convergence}
  \end{equation}
  where $\tau$ is the average number of local iterations within one round for each client, and $T$ is the number of total rounds.
  \label{theorem-zoo-one-point-convergence}
\end{theorem}
Assumptions \ref{assumption-loss-boundary}-\ref{assumption-gradient-dissimilarity-boundary} are detailed in Appendix \ref{sec-appendix-assumption}. 
Theorem \ref{theorem-zoo-one-point-convergence} has been proved by \citet{fang2022communication}.
To compare the convergence of \app and FedZO, we introduce FedMeZO by replacing BP in FedAvg with MeZO \cite{malladi2023mezo}, which utilizes the two-point estimator in Equation \eqref{eq-mezo}.
\begin{assumption}
FedMeZO converges similarly or faster compared to FedZO with $b_1=b_2=1$.
\label{assumption-fedmezo-fedzo}
\end{assumption}
The gradient estimated by the two-point estimator resembles the mean of two gradients by the one-point estimator (Equation \eqref{eq-zeroth-one-point}) with two opposing perturbations, as 
\begin{equation*}
\begin{matrix}
  \zb\! \left[\cL(\wb \!+\! \epsilon\zb;\xb) \!-\! \cL(\wb;\xb)\right] \!+\!(-\zb)\!\left[\cL(\wb \!-\! \epsilon\zb;\xb) \!-\! \cL(\wb;\xb)\right] \\
  = \zb \left[\cL(\wb + \epsilon\zb;\xb) - \cL(\wb - \epsilon\zb;\xb)\right].
\end{matrix}
\end{equation*}
Besides, the one-point estimator generally suffers from a higher estimation variance than the two-point estimator \cite{liu2018zeroth}. 
Thus, FedMeZO should exhibit a convergence trend not inferior to FedZO with $b_1 = b_2 = 1$.
This is also experimentally demonstrated by their test loss on \datani \cite{supernaturalinstructions} in Figure \ref{pic-converge-fedzo-fedmezo} (settings aligned with Section \ref{subsec-exp-setup}). 
Note that a recent work also proves the convergence of FL based on MeZO \cite{ling2024convergence}.

\para{Impact by seed restriction.}
Compared to FedMeZO, \app imposes additional restrictions on seed sampling. 
Perturbation sampling in \app can be formalized in two stages:
(1) Randomly and uniformly sampling $K$ distinct random seeds $\left \{s_1, s_2, \ldots, s_K \right \}$ from an infinity space, thus obtaining $K$ perturbations $\mathbb{Z} = \left \{\mathbf{z}_1, \mathbf{z}_2, \ldots, \mathbf{z}_K \right \}$, where $\mathbf{z}_j \in \mathbb{R}^d$ is sampled from $\mathcal{N}(\mathbf{0}, \mathbf{I}_d)$.
(2) Across the $T$ rounds of FL, randomly and uniformly sampling a total of $\Gamma = m \cdot \tau \cdot T$ perturbations from $\mathbb{Z}$ with replacement.
From the formal separation, the $\Gamma$ perturbations are still i.i.d.
\begin{lemma}
    (Mean of Perturbations). Given $\mathcal{Z}'= \{Z'_1, Z'_2, \ldots, Z'_K\}$ randomly sampled from $\mathcal{N}(0,1)$, $\Gamma$ variables $Z_1, Z_2, \ldots, Z_\Gamma$ randomly and uniformly sampled from $\mathcal{Z}'$ with replacement are i.i.d, and $\bar{S}_\Gamma = \frac{1}{\Gamma}\sum_{i=1}^\Gamma Z_i$ satisfy
{
\small
\begin{align}
    \operatorname{Pr}[|\bar{S}_\Gamma-0|\ge \varepsilon] & \le \\
     \frac{4}{K\varepsilon^2} &+ 2\exp\left(-\frac{\Gamma\varepsilon^2}{2[\max(\mathcal{Z}') - \min(\mathcal{Z}')]^2}\right). \nonumber
\end{align}
}
\label{lemma-mean-of-perturbations}
\end{lemma}
\emph{Proof.}
The proof is provided in Appendix \ref{subsec-appendix-proof-mean-perturbations}.

From Lemma \ref{lemma-mean-of-perturbations}, there exists a gap $\varepsilon$ between zero and the mean of $\mathbf{z}$ based on these seeds, which decreases with the increase of $K$ and $\Gamma$.
In our experiments, $K$ should be in the thousands, and $\Gamma$ increases by ten thousand per round on \datani. 
So both of them are usually large values.
Next, we analyze the impact of this mean shift ($\varepsilon$).
\begin{theorem}
    (Gradient Estimation Error). 
    With assumption on $L$-smoothness of $\cL$ (Assumption \ref{assumption-L-smooth}), for $\mathbf{z}\sim \mathcal{N}(\mathbf{0}, \mathbf{I}_d)$, the gradient $\hat{\mathbf{g}}_1$ estimated by the two-point estimator with seed restrictions and $\hat{\mathbf{g}}_0$ estimated by the two-point estimator without seed restrictions satisfy
    \begin{equation}
        \left \|\hat{\mathbf{g}}_1 - \hat{\mathbf{g}}_0\right \| \leq 2L\left\| \mathbf{z}\right\|\left\| \overrightarrow{\varepsilon}\right\| + L\left\| \overrightarrow{\varepsilon}\right\|^2.
    \end{equation}
    \label{theorem-gradient-estimation-error}
\end{theorem}
\emph{Proof.}
The proof is provided in Appendix \ref{subsec-appendix-proof-gradient-estimation-error}.

Intuitively, the estimation error $\hat{\mathbf{g}}_1 - \hat{\mathbf{g}}_0$ has a slight impact, because:
1) it is symmetrical in terms of the direction with the two-point gradient estimator,
and 
2) it is bounded by $\varepsilon$, which is usually very small.
We demonstrate the estimation error with a toy example for a more intuitive understanding.

\emph{Empirical Observation}.
We generate $K$ seeds, then sample $\Gamma$ perturbations and calculate the mean (absolute value) of them.
Note that the above toy example is performed with $\zb \in \RR^{10000}$, with $K$ and $\Gamma$ ranged in $[1, 16384]$ and $\left\{100, 1000, 10000\right\}$, respectively. 
With each combination of $(K, \Gamma)$, we perform the example 100 times and record the average results. 
To have a comparison with the situation without any restriction on seed sampling, we present a dotted line to indicate the situation with $\Gamma=10000$ and no restriction on $K$. 
The results are presented in Figure \ref{pic-mean-perturbation}, and we can find that:
The mean of $\mathbf{z}$ converges to zero with the increase of $K$, and when $K$ is over 1024 (the least adopted value in Section \ref{subsec-exp-setup}), it converges to a value very similar to that without seed restriction.
When fixing $K$, larger $\Gamma$ helps to lower $\varepsilon$.
Note that in our experiments, there are $50 * 200$ steps of updates generated in each round, thus, $\Gamma$ is significantly larger than this toy example, bringing a much smaller $\varepsilon$ than that in Figure \ref{pic-mean-perturbation}. 
Thus, the impact of seed restrictions is greatly limited in our settings.

Considering both: 
(1) the non-inferiority of the two-point estimator to the one-point estimator (Assumption \ref{assumption-fedmezo-fedzo}), and 
(2) the minimal nature of the gradient estimation error (Theorem \ref{theorem-gradient-estimation-error}), we can conclude that \emph{\app does not differ significantly from FedZO on convergence}.
\begin{figure}[t]
  \centering
  \subfigure[\modeldatajuicer]{
    \includegraphics[width=0.4725\linewidth]{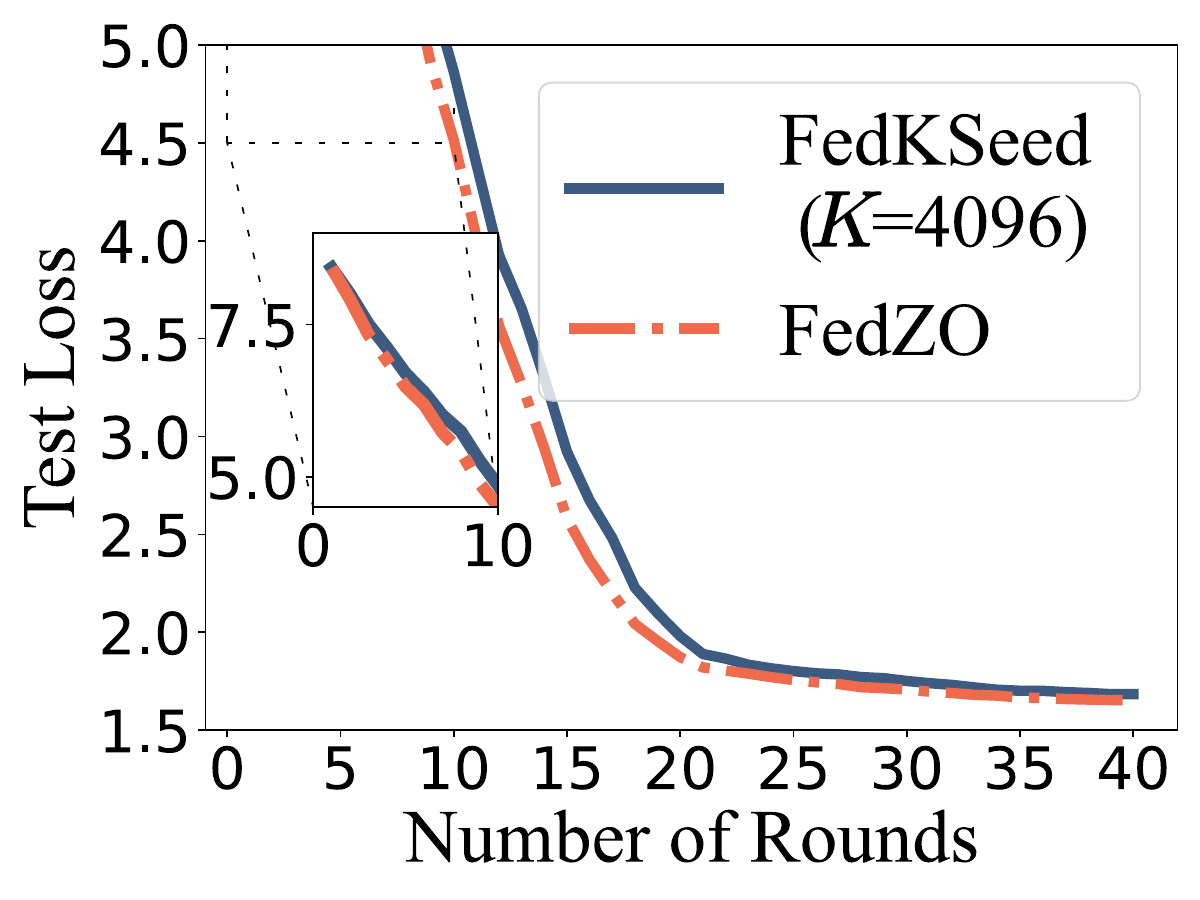}
  }
  \subfigure[\modelllama]{
    \includegraphics[width=0.4725\linewidth]{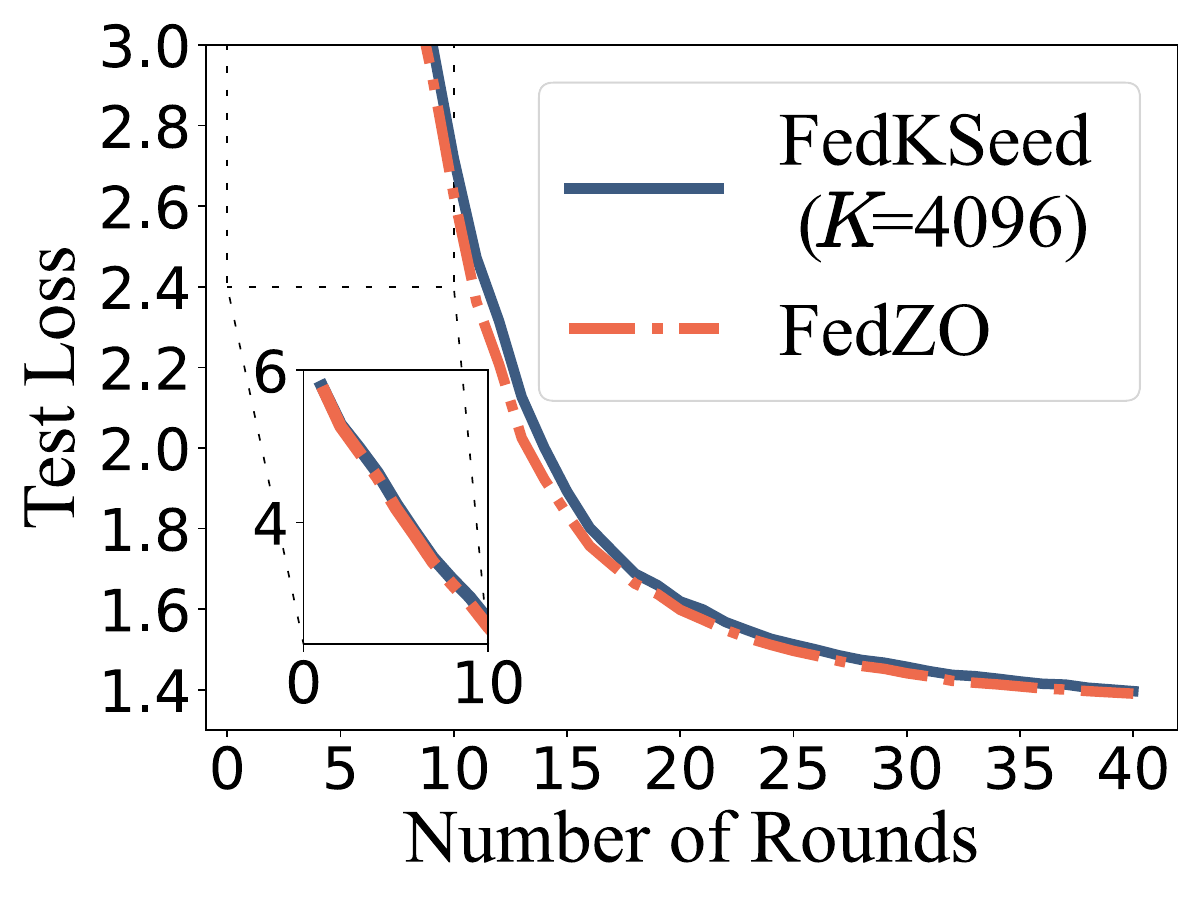}
  }
  \caption{Convergence of full-parameter tuning by \app and FedZO ($b_1 = b_2 = 1$) on \datani.}
  \label{pic-zo-convergence}
\end{figure}
We experimentally demonstrate it by presenting the test loss on \datani \cite{supernaturalinstructions} dataset in Figure \ref{pic-zo-convergence} (settings aligned with Section \ref{subsec-exp-setup}), showing that \app and FedZO ($b_1 = b_2 = 1$) share similar convergence trends. 

\subsubsection{Selection of $K$}
\label{subsubsec-approach-selection-K}
The selection of $K$ can be guided by the intrinsic dimension theory \cite{li2018measuring-intrinsic,aghajanyan2021intrinsic}.
Given $\GG = \begin{bmatrix} \sum_{\hat{\varrho}\in \cG_1} \hat{\varrho}, \ldots, \sum_{\hat{\varrho}\in \cG_K} \hat{\varrho} \end{bmatrix}^\top$, in which the $j$-th element is the summation of all scalar gradients corresponding to the perturbation with $s_j$. 
Equation \eqref{eq-fl-optimization} can be transformed to 
\begin{equation}
  \!\!\!\!\!\min_{\GG \in \RR^K} \!\sum_{i=1}^N c_i \cdot \EE_{\xb \sim \cD_i} \!\left[ \!\cL(\!\underbrace{\wb^0 \!+\! \begin{bmatrix}\zb_1, \ldots, \zb_K\end{bmatrix}\GG}_{\text{The difference from Equation \eqref{eq-fl-optimization}}}; \xb)\!\right].\!\!
  \label{eq-fl-optimization-intrinsic-transformed}
\end{equation}
Thus, \app actually performs federated tuning in a $K$-dimensional random subspace.
Equation \eqref{eq-fl-optimization-intrinsic-transformed} matches the form that trains a neural network in a subspace with the dimensionality equivalent to the intrinsic dimension \cite{li2018measuring-intrinsic}, with $\begin{bmatrix}\zb_1, \ldots, \zb_K\end{bmatrix}$ as the randomly generated projection matrix. 
Both $\wb^0$ and $\begin{bmatrix}\zb_1, \ldots, \zb_K\end{bmatrix}$ are frozen during training.
Thus, we can determine $K$ in the vicinity of the LLM's intrinsic dimension, which may approximately fall between $10^3$ and $10^4$ following \citet{aghajanyan2021intrinsic}.

\begin{principle}
  (Seed Insufficiency.)
  There exists a threshold $\overset{\leftharpoonup }{K}$ such that when $K \leq \overset{\leftharpoonup }{K}$, the accuracy of the model decreases with the reduction of $K$. 
  \label{principle-not-too-few-seeds}
\end{principle}
According to \citet{li2018measuring-intrinsic}, when $K$ is less than the codimension of the solution, solutions will almost not be found in the subspace, or the founded solution is low-quality, since low-dimensional subspaces do not possess sufficient complexity to embed the solution manifold. 
We also provide additional theoretical support for Principle \ref{principle-not-too-few-seeds} from the perspective of solving optimization problems in Appendix \ref{subsec-appendix-supports-principles-not-too-few}. 
From the analyses, $\overset{\leftharpoonup }{K}$ should exist, and a value approximates the intrinsic dimension typically serves as a good empirical estimate of it, such as 1024, as shown in Figure \ref{pic-performance-number-of-seeds}.

\begin{principle}
  (Seed Excessiveness.)
  There exists a threshold $\overset{\rightharpoonup }{K}$ such that given the total number of local training steps fixed, when $K \geq \overset{\rightharpoonup }{K}$, there is no upward trend in the accuracy of the model with the increase of $K$. 
  \label{principle-not-too-much-seeds}
\end{principle}
When $K$ surpasses the intrinsic dimension, the marginal gain in accuracy becomes increasingly smaller with the increase of $K$, since further increasing $K$ does not increase the ability to approximate the solution manifold. 
The redundancy of $K$ affects \app similar to that reported by \citet{li2018measuring-intrinsic,aghajanyan2021intrinsic} but with slight differences.
In \app, only one element of $\GG$ is optimized in each training step.
Intuitively, each element of $\GG$ requires several steps to be accurately estimated.
Given the fixed total number of update steps $\tau r m$, each element of $\GG$ consumes $\frac{\tau r m}{K}$ steps averagely.
Thus, increasing $K$ reduces the number of training steps for each element of $\GG$. 
When $K$ has reached an ample magnitude, this increment may induce adverse effects, as shown in Figure \ref{pic-performance-number-of-seeds}. 
We provide additional theoretical support to Principle \ref{principle-not-too-much-seeds} from the perspective of batch size in Appendix \ref{subsec-appendix-supports-principles-not-too-much}, and experimental support from the marginal gain on accuracy in terms of seed quantity in Appendix \ref{subsec-appendix-exp-marginal-gain}.
From these analyses, $\overset{\rightharpoonup }{K}$ should exist thus Principle \ref{principle-not-too-much-seeds} holds. 
From Figure \ref{pic-performance-number-of-seeds}, an empirically estimated value of it is 4096, which lies around the intrinsic dimension and tends to be slightly larger than it.

It is hard to map a specific LLM to a precise value of $K$ due to the complex architecture of LLMs. 
From the analyses, we can choose $K$ in $[\overset{\leftharpoonup}{K}, \overset{\rightharpoonup}{K}]$, i.e., an integer slightly larger than the intrinsic dimension.
Section \ref{subsec-exp-performance} experimentally demonstrates that for models with 1B and 3B parameters, $K$ can be several thousand so that \app performs well.

\subsection{Sampling Seeds with Non-uniform Probabilities}
\label{subsec-approach-bias}
This section enhances \app through enabling non-uniform \textbf{\underline{pro}}babilities for seed sampling to further reduce $K$ and boost the model accuracy, termed as \apppro.

The gradient $\gb$ indicates the direction of the steepest descent for loss function $\cL$ at a given point. 
However, in \app, $\gb$ is not available due to the removal of BP. 
The scalar gradient $\hat{\varrho}$ can be regarded as the estimated directional derivative of $\cL$ along $\zb$.
The similarity between different directional vectors and the gradient varies, affecting the rate of change in the objective and thus contributing differently to the descent of the loss.
The scalar gradient is determined by both the model, data instance and the similarity between true gradient $\gb$ and $\zb$.
Given the model and data instances equivalent in expectation for all perturbations, the average amplitude of scalar gradient $\psi_j = \frac{1}{\left | \cG_j \right |} \sum_{\hat{\varrho} \in \cG_j} \left | \hat{\varrho} \right |$ can quantify the importance of $\zb_j$. 
To avoid excessive probability differences in $\SSS$, we perform min-max normalization on $\left\{\psi_1, \ldots, \psi_K\right\}$, where $\psi_j^{\text{norm}}$ denotes the normalized amplitude. 
Then, we compute the probability $p_j$ of each seed $s_j \in \SSS$ as
\begin{equation}
  p_j = \frac{\exp(\psi^{\text{norm}}_j)}{\sum_{k=1}^K \exp(\psi^{\text{norm}}_k)}.
  \label{eq-seed-importance}
\end{equation}
The probabilities $\left\{p_1, \ldots, p_K \right\}$ are updated and sent to active clients in each round to guide the seed sampling of local training. 
In Section \ref{subsec-exp-performance}, we experimentally find that when significant seeds are sampled with higher probabilities, we can reduce the cardinality of seeds without sacrificing, and sometimes enhancing, model accuracy.

%% file: sections/5_exp.tex
\section{Experiments}
\label{sec-exp}

\subsection{Experimental Setup}
\label{subsec-exp-setup}
\para{Baselines.}
We choose 4 \emph{practical methods} tailored for federated LLM tuning as the baselines, including: 
(1) FedPTuning \cite{kuang2023federatedscope-LLM} with P-Tuning \cite{liu2023gpt} as the PEFT technique, trained by SGD; 
(2) FedPrompt \cite{kuang2023federatedscope-LLM} with Prompt Tuning \cite{lester2021power} as the PEFT technique, trained by SGD; 
(3) FedIT: a federated instruction tuning approach proposed by \citet{zhang2023fedit}, with LoRA as the PEFT technique and Adam \cite{adam} as the optimizer; 
and 
(4) FedIT-SGD: a variation of FedIT that replaces Adam with SGD.

To clarify the impact of the restriction on seed sampling in \app, we introduce 3 full-parameter tuning approaches as references, including:
(1) FedAvg \cite{mcmahan2017communication}, 
(2) FedZO \cite{fang2022communication}, 
and
(3) FedMeZO (synthetic of FedAvg and MeZO \cite{malladi2023mezo}).
Note that these three approaches incur excessive communication overheads due to the transmission of full LLM parameters in each round, thus may not be practical in real world. 
Their inclusion is purely for reference purposes.

\para{Datasets \& Evaluation.}
We use \datani (NI) \cite{supernaturalinstructions} and \datadolly \cite{DatabricksBlog2023DollyV2} datasets, following task held-out setups. 
Each of the 738 training tasks in NI is assigned to a unique client for local training while the 119 test tasks in NI are used for global evaluation, building a non-IID scenario with feature distribution skew \cite{tan2022towards}. 
The last task of \datadolly is used for global evaluation while the rest are split to 200 clients via Dirichlet distribution with $\alpha$=\{0.5, 5.0\} for local training, providing label distribution skew with varying degrees \cite{chen2023efficient}.
Rouge-L \cite{lin2004rouge} is used as the evaluation metric following \citet{dettmers2023qlora}, which correlates with the trend of accuracy on classification tasks \cite{supernaturalinstructions}. 
Considering the limited device resources, we take \modeldatajuicer \cite{chen2023datajuicer} and \modelllama \cite{touvron2023llama} as the foundation models.

\para{Implementations.}
We randomly sample 5\% of the clients to participate in FL in each round. 
The total number of communication rounds is set to 40 for NI and 60 for \datadolly. 
BP-based baselines conduct local training for one epoch, and \app and \apppro conduct local training for 200 steps.
Unless stated otherwise, we set $K$ to 4096 for \app, 1024 for \apppro with \modeldatajuicer, and 2048 for \apppro with \modelllama.
Note that from Figure \ref{pic-performance-number-of-seeds}, these settings are not tailored for best values of $K$ in corresponding scenarios, ensuring fair comparisons.
Please refer to Appendix \ref{sec-appendix-implementations} for more implementation details.
\begin{table*}[t]
  \renewcommand\arraystretch{0.85}
  \caption{Rouge-L (\%) comparisons. 
  Each cell presents the average Rouge-L in the last round of four runs with different random seeds. 
  Methods with excessive communication costs that may be \hl{impractical} in real world are introduced just as references and \hl{marked in gray}.
  \texb{Bold} and \texl{underlined} numbers are the \texb{best} and \texl{second-best} values among practical approaches (middle and bottom sections), respectively.
  }
  \label{tab-performance}
  \setlength\tabcolsep{5.67pt}
  \centering
  \begin{tabularx}{0.996\linewidth}{l|cc|cc|cc}
    \toprule[1.0pt]
    \multirow{2}{*}{Approach}         & \multicolumn{2}{c|}{\textbf{\datani}}                             & \multicolumn{2}{c|}{\textbf{\datadolly ($\alpha=0.5$)}}            & \multicolumn{2}{c}{\textbf{\datadolly ($\alpha=5.0$)}} \\
    \cmidrule{2-7}  
                                      & \modeldatajuicer                & \modelllama                     & \modeldatajuicer                & \modelllama                      & \modeldatajuicer                & \modelllama \\
    \midrule[1.0pt]
    \rowcolor{gray!20}FedAvg          & 22.08 \small{$\pm$ 1.52}        & 27.88 \small{$\pm$ 0.75}        & 32.30 \small{$\pm$ 1.23}	    & 34.27 \small{$\pm$ 0.45}         & 33.38 \small{$\pm$ 1.43}        & 33.95 \small{$\pm$ 0.79}\\
    \rowcolor{gray!20}FedZO           & 21.74 \small{$\pm$ 1.91}        & 29.46 \small{$\pm$ 0.38}        & 32.91 \small{$\pm$ 0.67}        & 36.34 \small{$\pm$ 0.39}         & 33.28 \small{$\pm$ 0.42}        & 	36.72 \small{$\pm$ 0.18}\\
    \rowcolor{gray!20}FedMeZO         & 21.71 \small{$\pm$ 1.26}        & 30.18 \small{$\pm$ 0.69}        & 33.32 \small{$\pm$ 0.14}        & 35.66 \small{$\pm$ 1.06}         & 33.07 \small{$\pm$ 0.47}        & 36.21 \small{$\pm$ 0.15}\\
    \midrule[1.0pt]
    FedPTuning                        & 19.61 \small{$\pm$ 2.71}        & 25.41 \small{$\pm$ 1.14}        & 23.98 \small{$\pm$ 3.23}        & 30.30 \small{$\pm$ 1.16}         & 25.33 \small{$\pm$ 2.48}        & 29.08 \small{$\pm$ 1.33} \\
    FedPrompt                         & \ \ 6.04 \small{$\pm$ 0.12}     & \ \ 8.95 \small{$\pm$ 2.47}     & 32.73 \small{$\pm$ 0.87}        & 24.50 \small{$\pm$ 4.78}         & 32.51 \small{$\pm$ 1.31}        & 23.94 \small{$\pm$ 4.15} \\
    FedIT-SGD                         & 19.40 \small{$\pm$ 1.83}        & 28.14 \small{$\pm$ 0.85}        & 27.23 \small{$\pm$ 0.68}        & 29.28 \small{$\pm$ 0.50}         & 27.28 \small{$\pm$ 1.35}        & 29.19 \small{$\pm$ 0.89} \\
    FedIT                             & 22.30 \small{$\pm$ 0.42}        & 28.13 \small{$\pm$ 0.50}        & 30.80 \small{$\pm$ 0.98}        & 33.23 \small{$\pm$ 1.51}         & 30.97 \small{$\pm$ 0.43}        & 33.68 \small{$\pm$ 1.07} \\
    \cmidrule{1-7}
    \app                       & \texl{22.33 \small{$\pm$ 1.72}} & \texl{29.77 \small{$\pm$ 0.75}} & \texl{32.90 \small{$\pm$ 0.37}} & \texl{35.64 \small{$\pm$ 0.83}}  & \texb{33.12 \small{$\pm$ 0.31}} & \texl{35.93 \small{$\pm$ 1.35}} \\
    \apppro                    & \texb{23.50 \small{$\pm$ 1.35}} & \texb{30.19 \small{$\pm$ 1.10}} & \texb{33.18 \small{$\pm$ 0.68}} & \texb{36.29 \small{$\pm$ 0.63}}  & \texl{33.00 \small{$\pm$ 0.34}} & \texb{35.95 \small{$\pm$ 1.41}} \\
    \bottomrule[1.0pt]
  \end{tabularx}
\end{table*}
\begin{table}[t]
  \centering
  \renewcommand\arraystretch{0.82}
  \caption{Per-round communication overhead and Peak GPU memory footprint of the approaches, where ``B'' denotes ``Bytes''.}
  \label{tab-overhead}
  \setlength\tabcolsep{2.3pt}
  \begin{tabularx}{\linewidth}{l|rr|rr}
    \toprule[1.0pt]
    \multirow{2}{*}{Approach} & \multicolumn{2}{c|}{\modeldatajuicer}                         & \multicolumn{2}{c}{\modelllama}  \\
    \cmidrule{2-5}
                              & Commun.         & Memory     & Commun.       & Memory    \\
    \midrule[1.0pt]
    FedAvg                    & 5.01 GB         & 17.8 GB    & 12.76 GB      & 39.1 GB  \\
    FedZO                     & 5.01 GB         & 3.4 GB     & 12.76 GB      & 7.6 GB   \\
    FedMeZO                   & 5.01 GB         & 3.5 GB     & 12.76 GB      & 7.8 GB  \\
    \midrule[1.0pt]
    FedPTuning                & 96.36 MB        & 11.9 GB    & 234.9 MB      & 16.3 GB   \\
    FedPrompt                 & 320.0 \ KB      & 11.8 GB    & 500.0 \ KB    & 19.0 GB   \\
    FedIT-SGD                 & 12.00 MB        & 12.4 GB    & 20.31 MB      & 18.2 GB   \\
    FedIT                     & 12.00 MB        & 12.4 GB    & 20.31 MB      & 18.3 GB   \\
    \cmidrule{1-5}
    \app                      & 17,988 B    & \ \ 3.5 GB & 17,988 B  &\ \ 7.8 GB \\
    \apppro                   & \ \ 9,796 B & \ \ 3.5 GB & 17,988 B  &\ \ 7.8 GB \\
    \bottomrule[1.0pt]
  \end{tabularx}
\end{table}

\subsection{Comparisons on Accuracy Performance}
\label{subsec-exp-performance}
\para{Comparisons with practical baselines.}
From Table \ref{tab-performance}, \app and \apppro achieve the top two performances among the practical approaches across all six scenarios.
In particular, on \datadolly ($\alpha=0.5$) with \modelllama, \apppro outperforms the best practical baseline, FedIT, by 3.06\%. 
These improvements can be attributed to the benefits of full-parameter tuning, where the number of trainable parameters is significantly larger compared to PEFT techniques as Figure \ref{pic-training-efficiency}.
Further, the gains achieved by \apppro over the best practical baseline, FedIT, are generally larger with \modelllama than \modeldatajuicer, since with the same LoRA configuration, the model size increase does not proportionally affect the number of trainable parameters in FedIT as much as it does in our approaches. 

\para{Comparisons with full-parameter tuning approaches.}
From Table \ref{tab-performance}, FedAvg, FedZO and FedMeZO do not exhibit higher accuracy than \app and \apppro. 
However, from Table \ref{tab-overhead}, \app and \apppro have significant advantages in communication and memory costs over FedAvg, and communication advantages over FedZO and FedMeZO. 
Thus, the seed sampling restriction method we proposed does not introduce noticeable negative effects.

\begin{figure*}[t]
  \centering
  \subfigure[\modeldatajuicer \datani]{
    \includegraphics[width=0.282\linewidth]{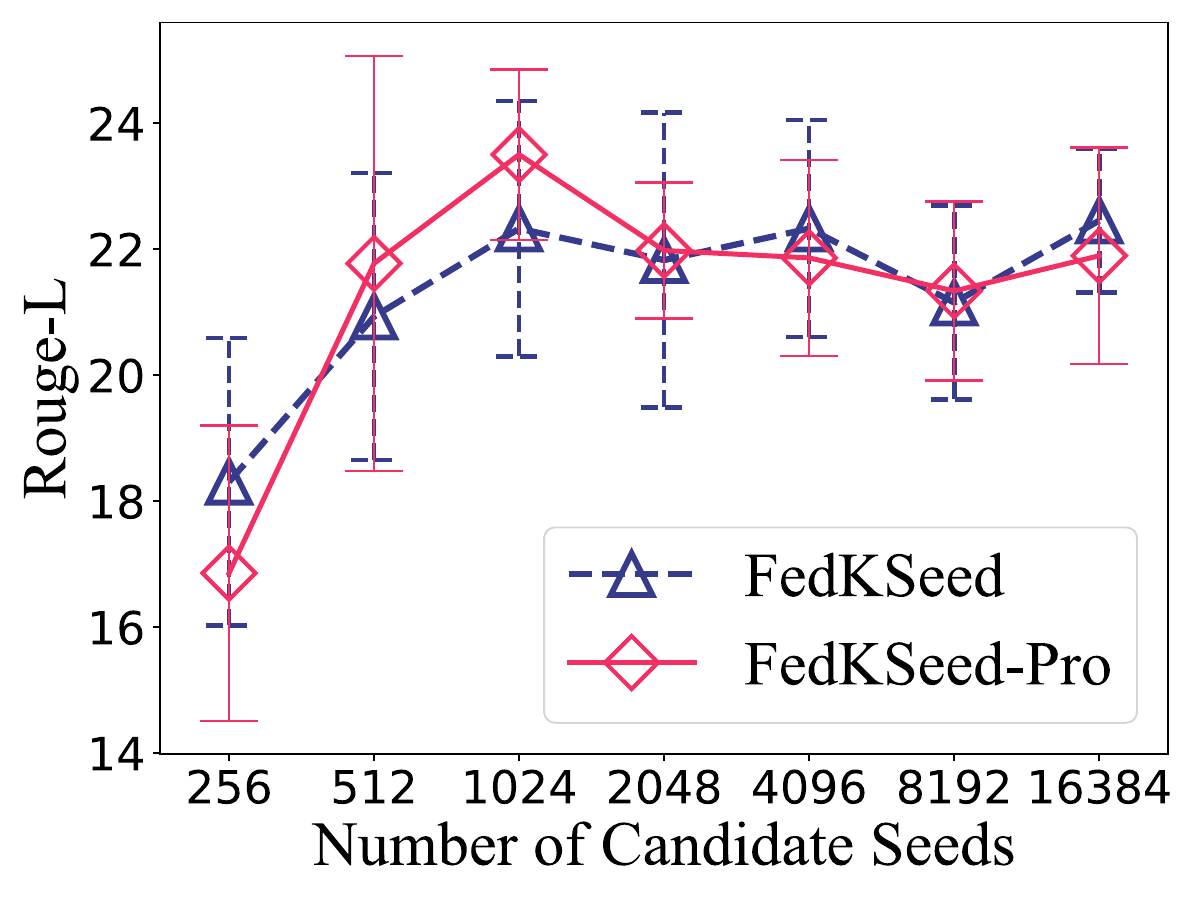}
  }
  \setcounter{subfigure}{2}
  \subfigure[\modeldatajuicer \datadolly ($\alpha$=0.5)]{
    \includegraphics[width=0.282\linewidth]{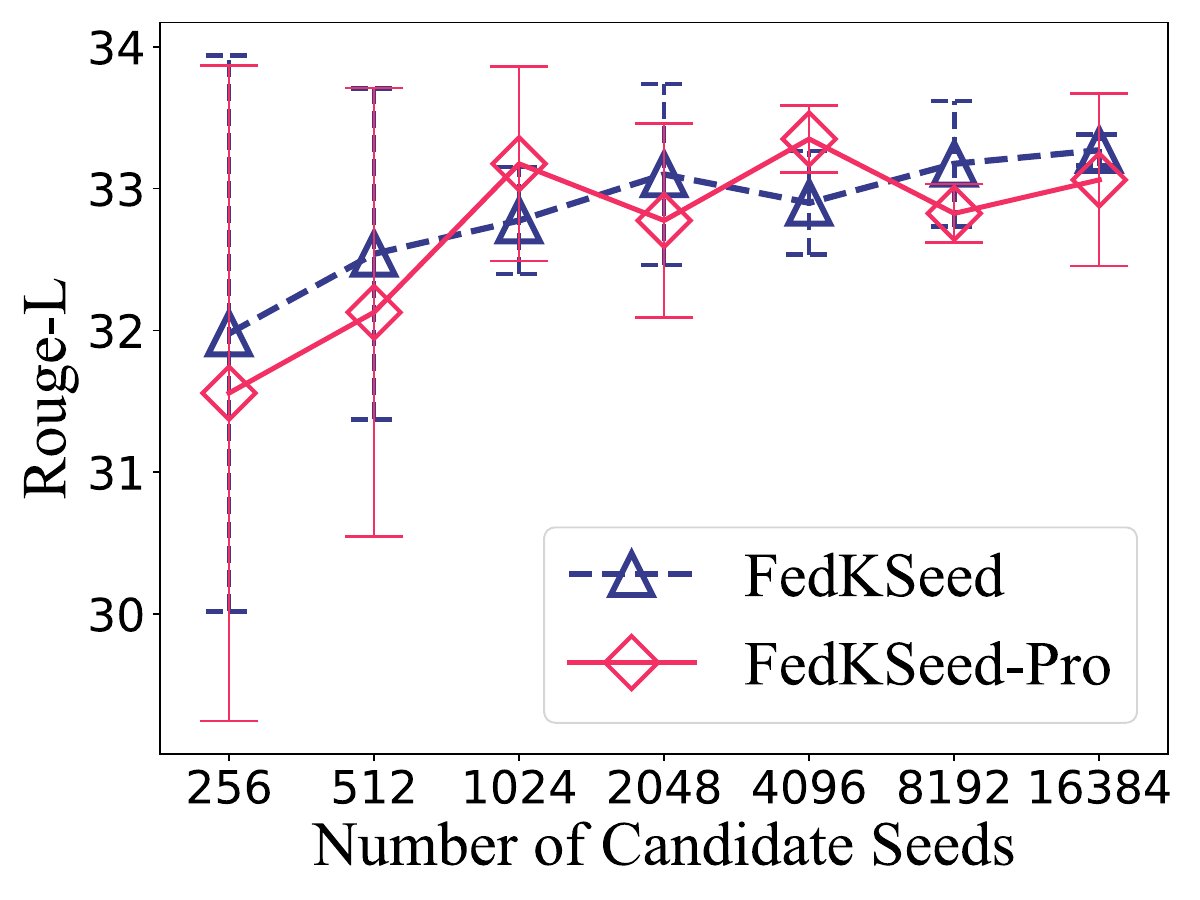}
  }
  \setcounter{subfigure}{4}
  \subfigure[\modeldatajuicer \datadolly ($\alpha$=5.0)]{
    \includegraphics[width=0.282\linewidth]{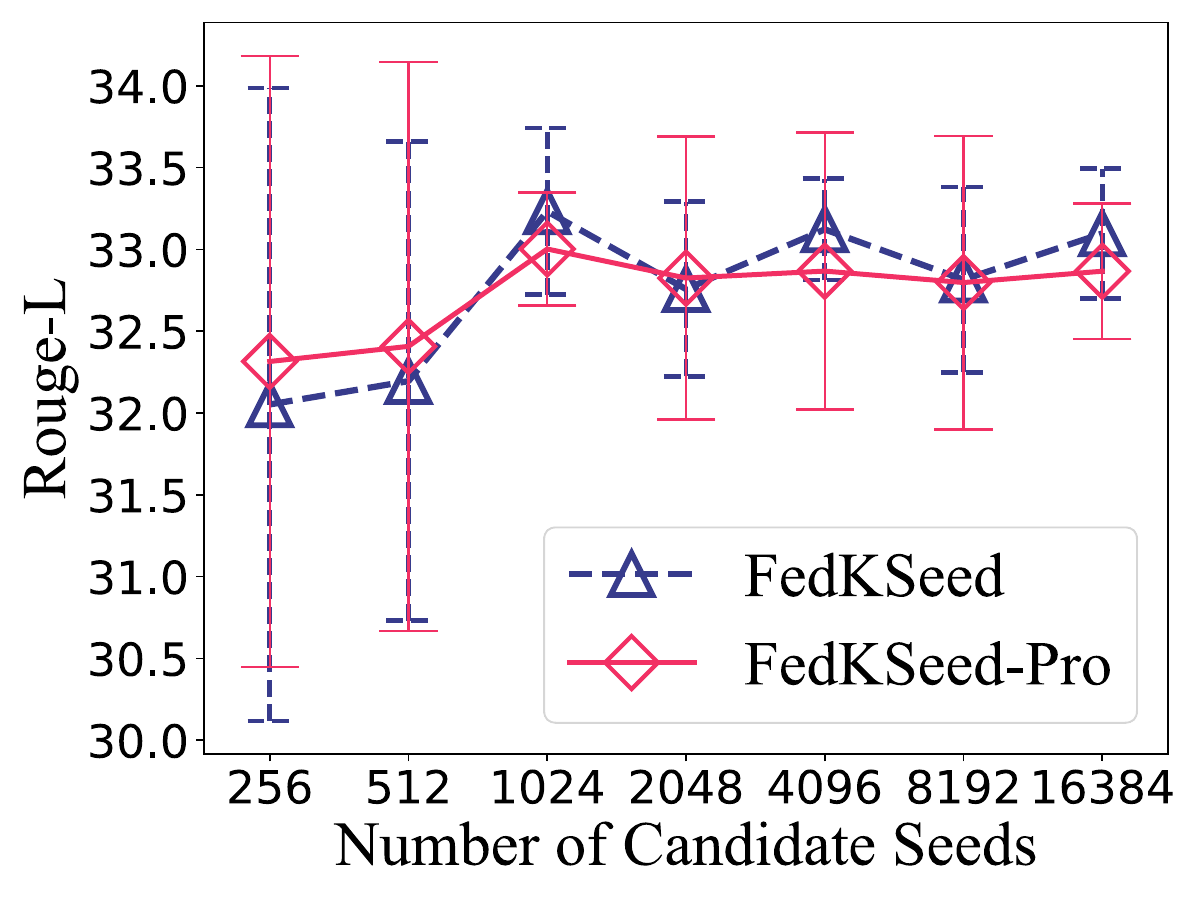}
  }
  \setcounter{subfigure}{1}
  \subfigure[\modelllama \datani]{
    \includegraphics[width=0.282\linewidth]{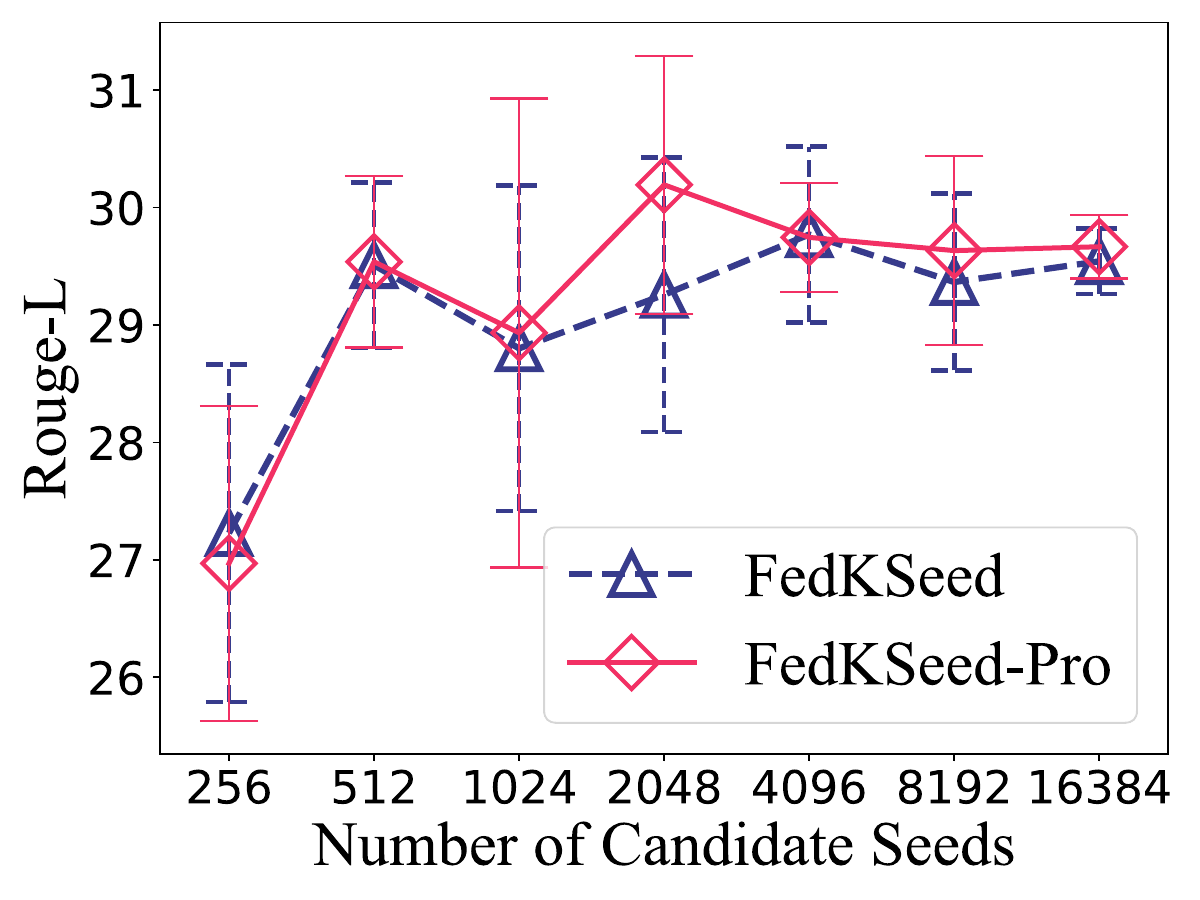}
  }
  \setcounter{subfigure}{3}
  \subfigure[\modelllama \datadolly ($\alpha$=0.5)]{
    \includegraphics[width=0.282\linewidth]{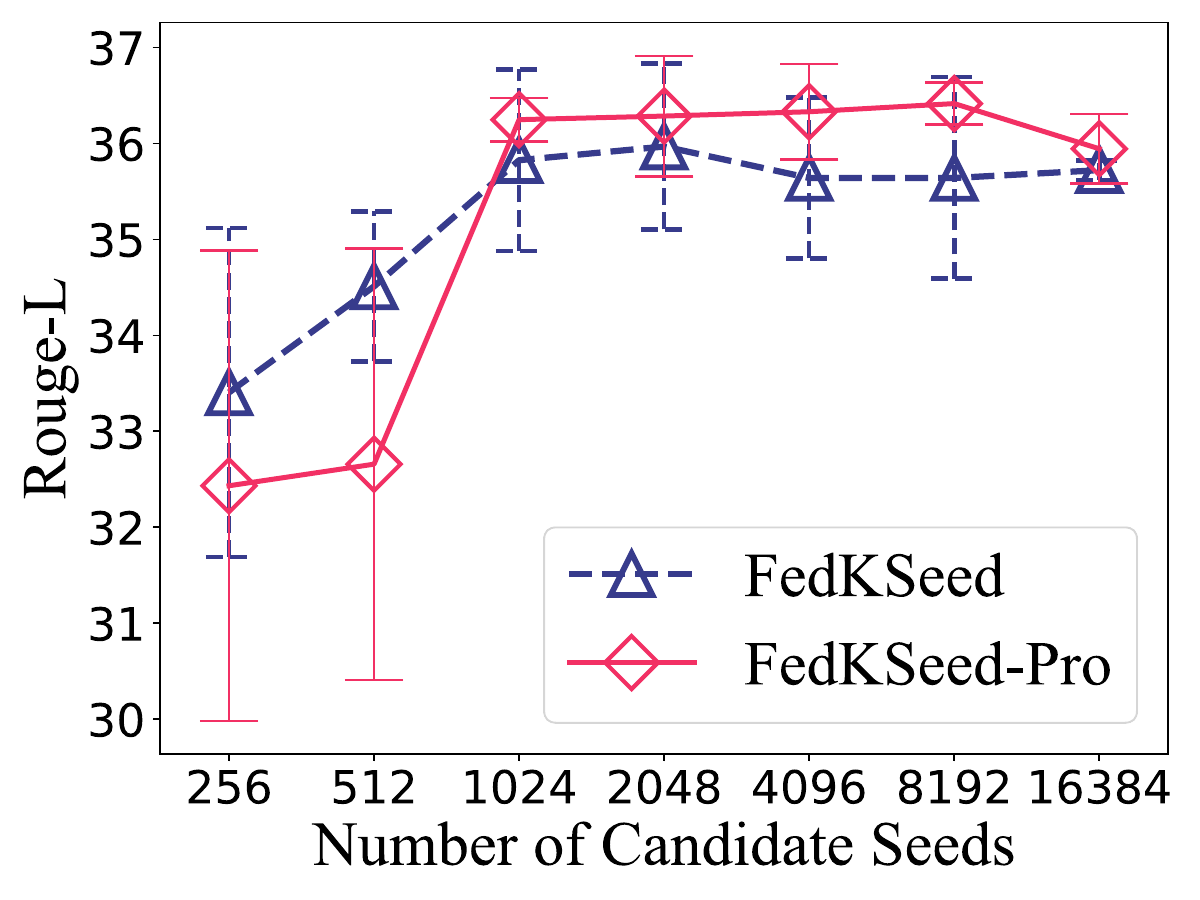}
  }
  \setcounter{subfigure}{5}
  \subfigure[\modelllama \datadolly ($\alpha$=5.0)]{
    \includegraphics[width=0.282\linewidth]{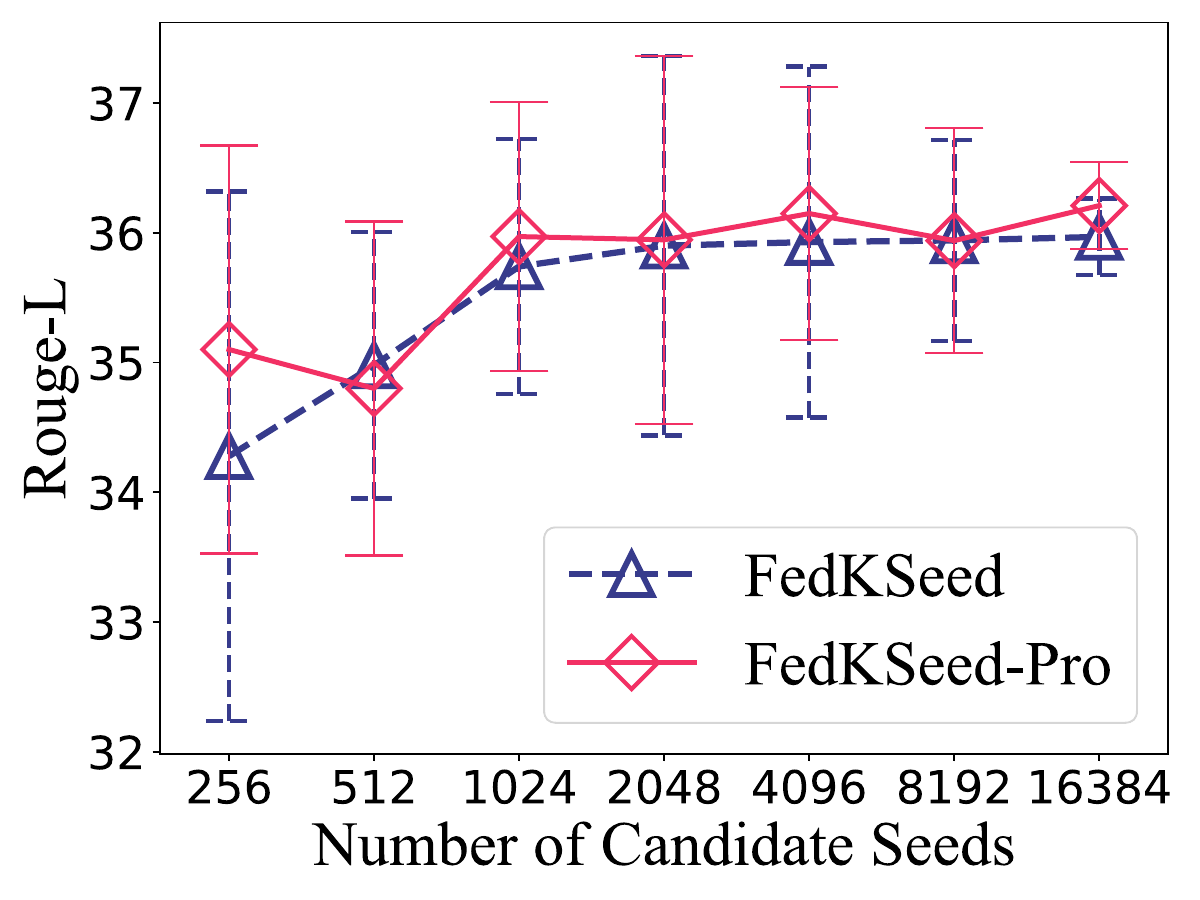}
  }
  \caption{Performance on Rouge-L of \app and \apppro with different cardinality of candidate seeds.}
  \label{pic-performance-number-of-seeds}
\end{figure*}

\para{Effect of $\bm{K}$.}
To validate Principles \ref{principle-not-too-few-seeds} and \ref{principle-not-too-much-seeds}, and to understand the relationship between the number of perturbation seeds ($K$) and the accuracy of \app and \apppro, we examine their performance with varying $K$, as depicted in Figure \ref{pic-performance-number-of-seeds}.
We observe that when the $K$ exceeds the recommended range specified in Section \ref{subsec-exp-setup}, the accuracy does not improve and may occasionally decline. 
Because the total number of optimization steps is constant, with more seeds, the likelihood that each seed consumes sufficient data to determine its step size is reduced. 
Conversely, with too few seeds, the performance of both \app and \apppro deteriorates due to the limited expressiveness resulting from an insufficient number of perturbations.
Thus, the value of $K$ should be balanced as discussed in Section \ref{subsubsec-approach-selection-K}: not too high to waste computational costs, nor too low to restrict the model's expressiveness. 
Our experimental results indicate that for models with 1B to 3B parameters, setting $K$ in the range of [1024, 4096] is preferable.

\para{Effect of seed probabilities.}
\apppro gains superior performance in five out of six scenarios and comparable results on \datadolly ($\alpha=5.0$) with \modeldatajuicer compared to \app. 
This highlights the effectiveness of the non-uniform seed sampling proposed in Section \ref{subsec-approach-bias}.
\apppro makes the probabilities of each seed being sampled differ by several multiples, as shown in Appendix \ref{subsec-appendix-exp-seed-probabilities}.
Thus, by preferentially sampling seeds with greater importance (those with larger scalar gradient magnitudes), the accuracy of \app can be further enhanced.

\subsection{Comparisons on Overheads}
\label{subsec-exp-cost}
Table \ref{tab-overhead} shows that \app and \apppro incur the least communication and memory costs, where the $K$ randomly selected seeds are encoded with one single seed that only occupies 4 Bytes. 
The time costs are illustrated in Appendix \ref{subsec-appendix-exp-training-efficiency}.
The calculation of communication costs is detailed in Appendix \ref{sec-appendix-calculation-overhead}. 
\app and \apppro enhance communication efficiency by removing the transmission of trainable parameters, and memory efficiency by omitting BP with the in-place ZOO \cite{malladi2023mezo}.
Thus, they can be effectively applied to tune full LLMs on end devices with limited communication and memory budgets.
Besides, \apppro requires fewer seeds to achieve the same accuracy compared to \app, as shown in Figure \ref{pic-performance-number-of-seeds}.
It accelerates the synchronization to the latest model relative to \app, since computing the latest model takes longer with more candidate seeds, as shown in Figure \ref{pic-time-pulling}.

\subsection{Hyper-parameter Sensitivity}
\begin{figure}
  \centering
  \begin{minipage}[t]{0.43\linewidth}
    \centering
    \includegraphics[width=\linewidth]{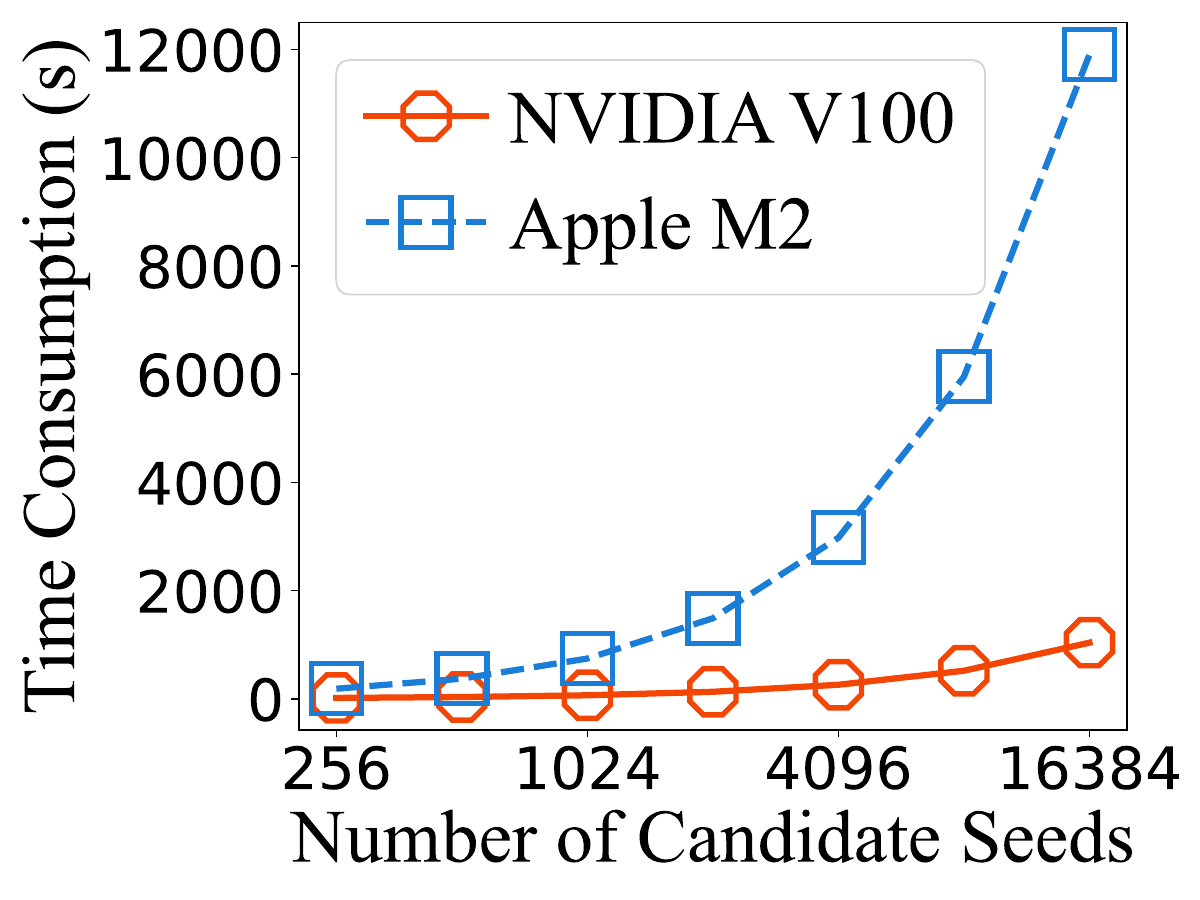}
    \caption{Time consumed to calculate the latest model (\modelllama) by \app.}
    \label{pic-time-pulling}
  \end{minipage}
  \hspace{0.1cm}
  \begin{minipage}[t]{0.526\linewidth}
    \centering
    \includegraphics[width=\linewidth]{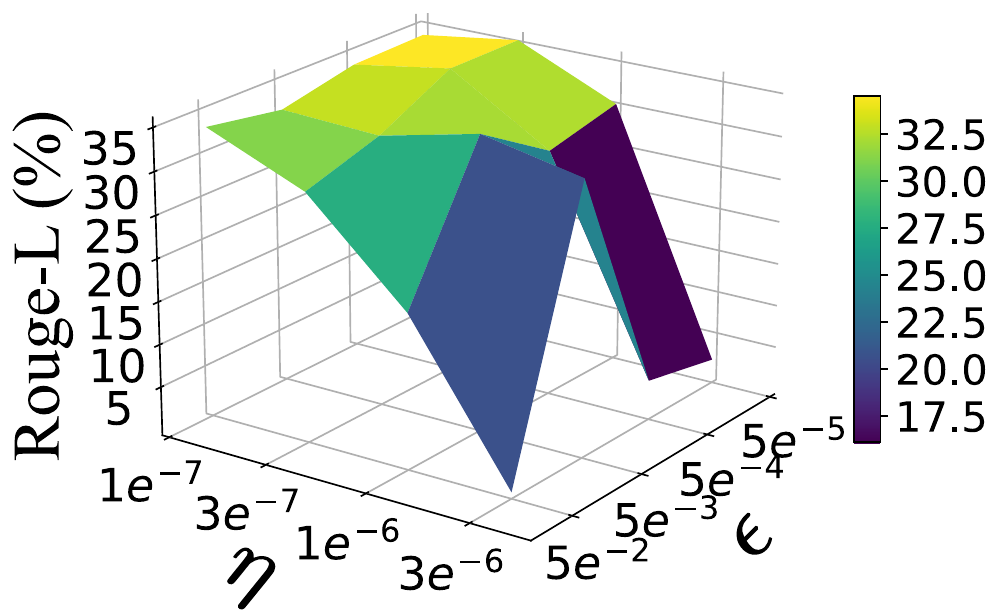}
    \caption{Rouge-L of \apppro on \datadolly ($\alpha=0.5$) with different combinations of $\eta$ and $\epsilon$.}
    \label{pic-hyper-mezo}
  \end{minipage}
\end{figure}

\label{subsec-exp-parameter}
Given that ZOO serves as the training technique, we investigate the impact of key hyperparameters in ZOO, i.e., the learning rate ($\eta$) and the perturbation scale ($\eps$), with \apppro for \modelllama on \datadolly ($\alpha=0.5$) as a case study.
From Figure \ref{pic-hyper-mezo}, both $\eta$ and $\eps$ should not be excessively large. 
Since $\eps$ determines the magnitude of perturbations applied during gradient estimation, a smaller $\eps$ leads to a more accurate gradient approximation. 
However, too small $\eps$ may result in numerical underflow when using half-precision floating-point numbers.
An overly large value of $\eta$ can cause too aggressive update steps, potentially causing significant deviation from the optimum or even diverge.

\subsection{Comparisons in Various Federated Scenarios}
\label{subsec-exp-fl-scenarios}
\begin{figure}[t]
  \centering
  \subfigure[Impact of total client count $N$ (on \datani).]{
    \includegraphics[width=0.456\linewidth]{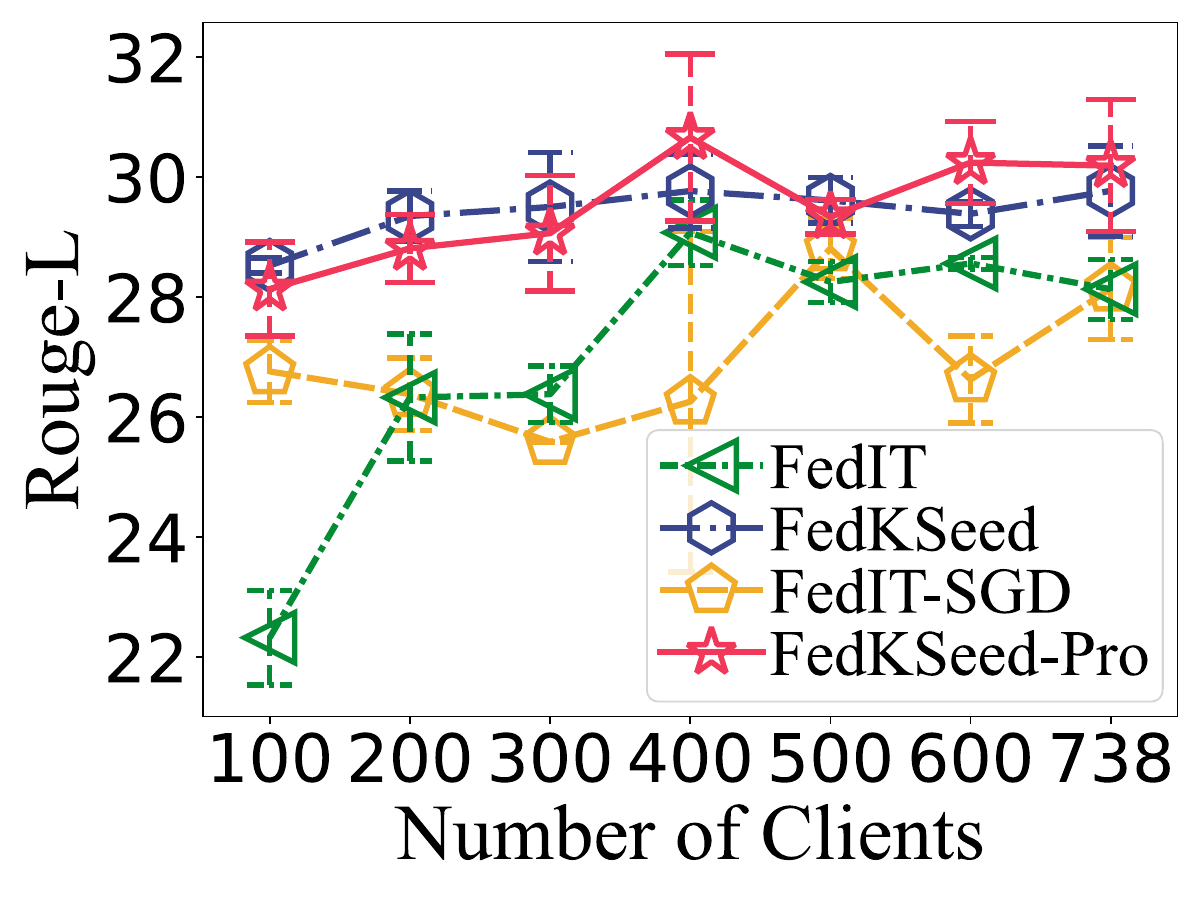}
    \label{pic-FL-scenrios-number-of-client}
  }
  \hspace{0.01cm}
  \subfigure[Impact of activate client ratio $m/N$ (on \datadolly $\alpha$=0.5).]{
    \includegraphics[width=0.456\linewidth]{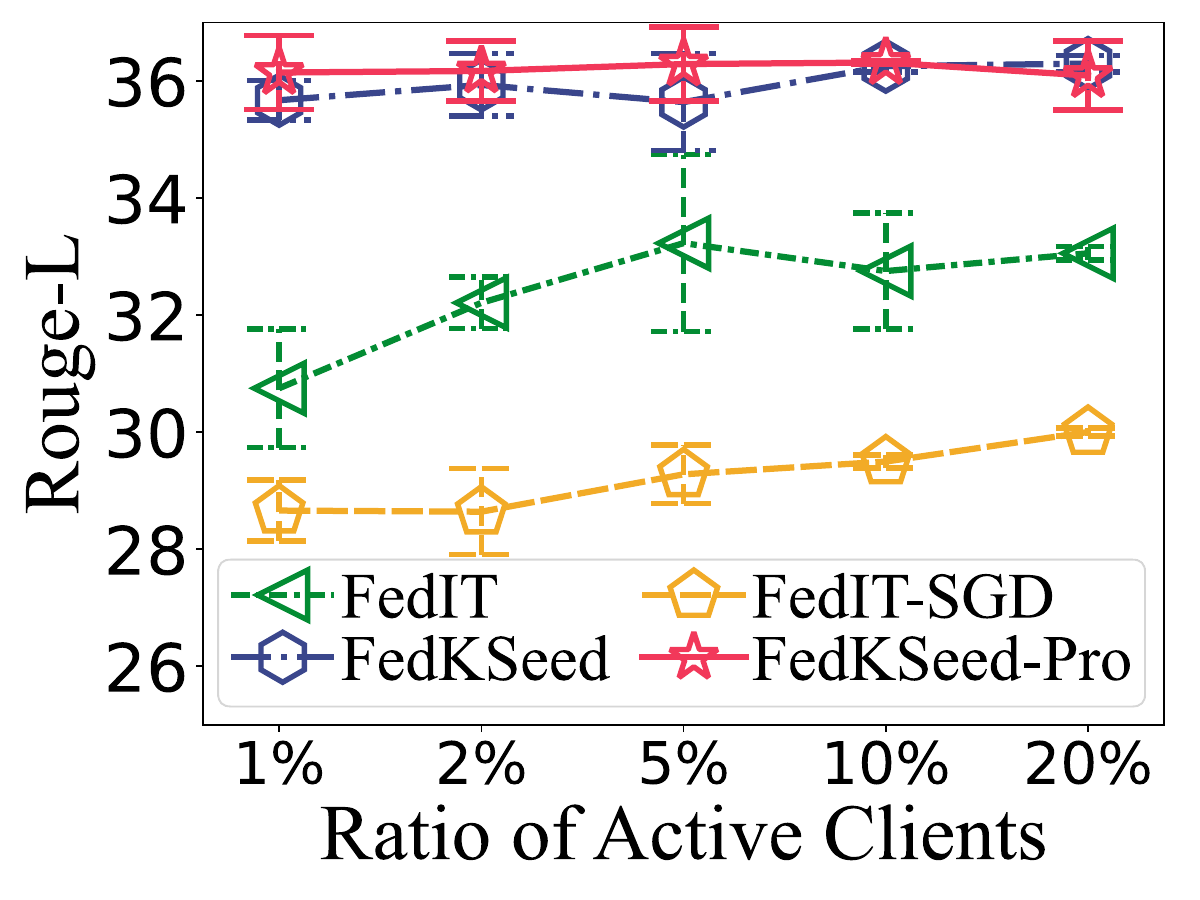}
    \label{pic-FL-scenrios-active-ratio}
  }
  \caption{Performances in various FL scenarios with \modelllama. $K$=4096 in \app and 2048 in \apppro, as Section \ref{subsec-exp-setup}.}
  \label{pic-performance-various-FL-scenrios}
\end{figure}

We further evaluate the two strongest baselines and our approaches under different FL settings.
Figure \ref{pic-FL-scenrios-number-of-client} shows that benefiting from more valuable data, both our approaches and the baselines gain better accuracy as $N$ increases, although not always monotonically. 
It further confirms the importance of federated tuning since it can leverage a broader range of data sources. 
Figure \ref{pic-FL-scenrios-active-ratio} presents that the accuracy of these approaches increases and stabilizes as more clients participate in each round of FL. 
Furthermore, our approaches consistently outperform the baselines in various FL scenarios, further demonstrating their superiority.

%% file: sections/6_conclusion.tex
\section{Conclusion}
Existing federated fine-tuning approaches for LLMs usually rely on PEFT techniques. 
Considering PEFT still falls short in FL scenarios compared to full-parameter tuning, we focus on enabling full-parameter tuning of billion-sized LLMs on devices with FL. 
To fulfill this, we design \app characterized by a theoretically-informed seed-reuse paradigm, where only a limited number of candidate seeds and corresponding scalar gradients need to be transmitted between the server and clients.
It enables federated full-parameter tuning of LLMs with per-round communication costs lower than 18 kilobytes.
Based on \app, inspired by the fact that the scalar gradient of a perturbation is the directional derivative of the true gradient, we propose a strategy to quantify the importance of seeds and grant differentiated sampling probabilities to them. 
It reduces the number of required seeds, thus speeding up the obtaining of the latest model while achieving higher accuracy than \app. 
Extensive experiments conducted on real-world datasets demonstrate our approaches surpass FL baselines tailored for LLM tuning on the accuracy of unseen tasks, communication cost and memory footprint at the same time.

Our work may raise some new potential research directions, such as decentralized federated fine-tuning since the communication cost is more critical in this context. 
More benefits brought by this work are discussed in Appendix \ref{sec-appendix-discussions}.

%% file: sections/7_0_toc_appendix.tex
\section*{Appendix}
We provide more discussions and experiments of this work in the appendix and organize them as follows:
\begin{itemize}
    \item Appendix \ref{sec-appendix-related}: we discuss detailed technical distinctions of our approach to existing approaches related to our work.
    \item Appendix \ref{sec-appendix-notations}: we provide Table \ref{tab-notations} to list the used notations along with their corresponding meanings.
    \item Appendix \ref{sec-appendix-algorithm}: we summarize the processes of \app and \apppro in Algorithm \ref{algo:main_algo}.
    \item Appendix \ref{sec-appendix-assumption}: we provide detailed assumptions relied upon Theorem \ref{theorem-zoo-one-point-convergence}.
    \item Appendix \ref{sec-appendix-detailed-proofs}: We provide the detailed proofs of Lemma \ref{lemma-mean-of-perturbations} and Theorem \ref{theorem-gradient-estimation-error}.
    \item Appendix \ref{sec-appendix-supports-principles}: we offer additional analytical support for Principles \ref{principle-not-too-few-seeds} and \ref{principle-not-too-much-seeds} from a perspective distinct from Section \ref{subsubsec-approach-selection-K}.
    \item Appendix \ref{sec-appendix-implementations}: we provide implementation details of our approach for better reproducibility.
    \item Appendix \ref{sec-appendix-exp}: we provide supplementary experiments, including (1) the marginal improvement on Rouge-L gained per extra seed across varied seed quantities by \app and \apppro in Appendix \ref{subsec-appendix-exp-marginal-gain}; (2) the convergence illustrations in Appendix \ref{subsec-appendix-exp-convergence}; (3) illustrations of seed probabilities calculated by \apppro in Appendix \ref{subsec-appendix-exp-seed-probabilities}; and (4) the time consumption for local training in Appendix \ref{subsec-appendix-exp-training-efficiency}.
    \item Appendix \ref{sec-appendix-calculation-overhead}: we provide the detailed calculation of the communication overhead of our approach and the baselines, to explain how to derive the numerical communication overheads of these approaches in Table \ref{tab-overhead}.
    \item Appendix \ref{sec-appendix-discussions}: we discuss the extended benefits brought by our proposed approach in real-world applications.
\end{itemize}

%% file: sections/7_appendix_technical.tex
\section{Detailed Technical Comparisons}
\label{sec-appendix-related}
In this section, we compare our approach with existing approaches that have a certain relationship with our approach from technical, objective, and performance perspectives. 

In recent years, there have been some researches that utilize the universal random seeds to lower the communication cost of transmitting model parameters or gradients \cite{xu2023billion-sized,feng2023baffle,zelikman2023onebyte,rahimi2023evofed}. 
These approaches can respectively achieve unidirectional $\cO(d)$ \cite{xu2023billion-sized,feng2023baffle}, bidirectional $\cO(1)$ \cite{zelikman2023onebyte,rahimi2023evofed} communication cost.

FwdLLM \cite{xu2023billion-sized}, BAFFLE \cite{feng2023baffle} and FedZeN \cite{maritan2023fedzen} do not restrict the number of candidate seeds. 
Therefore, they still need to distribute the latest trainable model parameters to clients in each round. 
If they are adopted for full-parameter tuning of LLMs with FL, clients have to consume tremendous communication resources to download the latest global LLM in each round, thus, it may prevent many clients from participating in FL since the cost and quality of a wireless connection can vary greatly between different countries \cite{dorfman2023docofl}. 

The approach proposed by \citet{zelikman2023onebyte} achieves bidirectional $\cO(1)$ communication cost, optimizing the communication efficiency to the utmost. 
However, there is no such thing as a free lunch. 
As we have presented in Figure \ref{pic-intro-time-consumption}, \citet{zelikman2023onebyte} sample seeds from a vast space that is almost infinite, causing each client must replicate the update steps performed by all other clients to obtain the latest model. 
Consequently, the overhead of calculating the latest model from $\wb^0$ also grows indefinitely as the rounds of FL continue, quickly reaching a level that is unsustainable for end devices.
Since the fine-tuning of large models typically requires many steps over a large instruction dataset, the approach proposed by \citet{zelikman2023onebyte} is not suitable for full-parameter tuning of LLMs with FL on devices. 
Moreover, \citet{zelikman2023onebyte} conduct experiments only on a small sentiment classification dataset, i.e., SST2 \cite{socher2013recursive}, and train a model for a total of only 16,000 steps.
However, for complex datasets, such as \datani \cite{supernaturalinstructions}, this number of update steps may not be sufficient for the convergence of LLMs.

One recent work, EvoFed \cite{rahimi2023evofed}, also achieves bidirectional $\cO(1)$ communication cost. 
However, EvoFed is fundamentally different from \app, and it is not specifically designed for fine-tuning LLMs: 
(1) During local training, EvoFed first conducts BP to compute the true gradient. 
Then it generates $K$ noise-perturbed model populations and tries to represent the true gradient by the summation of these noise perturbations.
The weight coefficient of each population is determined by the $l_2$ norm of its pairwise differences with the true gradient.
Thus, EvoFed still relies on the BP process to get the true gradient, which is not practical for fine-tuning LLMs with billions of parameters. 
As reported by \citet{malladi2023mezo}, fine-tuning full LLMs imposes a tremendous memory burden. 
For example, full-parameter tuning of an LLM with 2.7B parameters with an average sequence length of 400 tokens can consume up to 55GB of memory.
Such a level of memory consumption exceeds the capabilities of a single graphics card such as NVIDIA V100 (32GB), not to mention an end device. 
(2) The calculation of $l_2$ norm of the pairwise differences between a perturbation and the true gradient also consumes tremendous computation resources for billion-sized LLMs. 
This operation must be repeated $K$ times each round, further increasing the computational load.
(3) As highlighted in Equation (12) presented by \citet{rahimi2023evofed}, when facing the scenario of partial client participation, a client that has not participated in FL for $\varsigma $ rounds has to perform $\varsigma \cdot K$ model updates to calculate the latest global model, while \app and \apppro still only need to perform $K$ steps.
In the experiments conducted by \citet{rahimi2023evofed}, EvoFed is evaluated on small visual datasets with small models, i.e., containing at most 2.3 million parameters, while our approach is evaluated on LLMs with at most 3.43 billion parameters.

Note that there are also some BP-based FL approaches that optimize the communication overhead by encoding the gradient into low-dimensional spaces, such as model sketch \cite{melas2021intrinisic,Rothchild2020FetchSGD}. 
As we have discussed in Section \ref{sec-related}, these approaches are not tailored for LLMs and are not suitable for LLM tuning on end devices due to the tremendous memory footprint, especially for full-parameter tuning of billion-sized LLMs.
Besides, they are only evaluated by small models instead of billion-sized LLMs. 

Based on the technical comparisons presented above, \app emerges as the first approach that enables the possibility of federated full-parameter tuning of billion-sized LLMs on devices.
It achieves this by reducing the communication cost to a constant of less than 18 kilobytes and minimizing memory consumption to the level required for inference.

%% file: sections/8_appendix_notations.tex
\section{Summarization of Notations}
\label{sec-appendix-notations}
\begin{table}[!h]
\centering
\renewcommand\arraystretch{1.16}
\caption{Notations and corresponding meanings.}
\label{tab-notations}
\begin{tabular}{ll}
\toprule[1.0pt]
Notation &
Meaning \\
\midrule[1.0pt]
$N$ & Number of total clients. \\
$m$ & Average number of active clients in each round. \\
$\tau$ & Number of steps during local training in each round. \\
$T$ & Number of total rounds. \\
$r$ & Number of rounds that have already been elapsed. \\
$\Gamma$ & $\Gamma = m \cdot \tau \cdot T$, denoting the total steps of updates across the $T$ rounds of FL. \\
$\cD_i$ & Local dataset of client $i$. \\
$\xb$ & A data sample in $\cD_i$, such as a Q-A pair in \datani. \\
$c_i$ & The weight of aggregation for client$i$. \\
$\wb_0$ & Initial weights of an LLM, $\wb_0 \in \RR^d$. \\
$d$ & The dimensionality of the LLM. \\
$\wb^r_{i, t}$ & Model weights of client $i$ after the $t$-th step of local training in round $r$. \\
$\cL(\wb;\xb)$ & The loss evaluated at model $\wb$ on a data instance $\xb$. \\
$\SSS$ & The set of distinct candidate seeds. \\
$K$ & The size of $\SSS$, i.e., $\left | \SSS \right |$. \\
$s_j$ & One random seed in $\SSS$, indexed by $j$. \\
$\zb_j$ & A randomly sampled perturbation drawn from $\cN(\mathbf{0}, \Ib_d)$, indexed by $j$. \\
$\gb$ & True gradient computed with stochastic gradient descent algorithms. \\
$\hat{\gb}_j$ & The gradient estimated with zeroth-order optimization, on the perturbation $\zb_j$. \\
$\hat{\varrho}_j$ & The scalar gradient corresponding to $\hat{\gb}_j$, where $\hat{\gb}_j = \hat{\varrho}_j \cdot \zb_j$. \\
$(s_j,\hat{\varrho}_j)$ & A (seed, scalar gradient) pair. \\
$\eta$ & Learning rate during local training. \\
$\epsilon$ & Scale of a random perturbation. \\
\bottomrule[1.0pt]
\end{tabular}
\end{table}

%% file: sections/9_appendix_algorithm.tex
\section{The Algorithm of the Proposed \app and \apppro}
\label{sec-appendix-algorithm}
\begin{algorithm}
  \SetAlgoLined
  \DontPrintSemicolon
  \SetNoFillComment
  \SetKwFunction{FClientTraining}{ClientTraining}
  \SetKwFunction{FUpdateModel}{UpdateModel}
  \SetKwProg{Fn}{Function}{:}{end} 
  \setlength{\abovedisplayskip}{3pt}
  \setlength{\belowdisplayskip}{3pt}
  \setlength{\abovedisplayshortskip}{3pt}
  \setlength{\belowdisplayshortskip}{3pt}
  \caption{The processes of \textbf{\app}, where the \underline{underlined} components and processes are only required by the enhanced version of it that samples seeds during local training with non-uniform probabilities, i.e., \underline{\apppro}.}
  \label{algo:main_algo}
  \KwIn{$N$, $K$, $\wb^0$, $\eta$, $\left\{c_1, \dots, c_N\right\}$, $T$ and $\tau$.}
  \KwOut{The global model $\wb^T$ that has been fine-tuned for $T$ rounds.}
  \vspace{0.2cm}
  \textbf{Server Executes:} initialize $K$ candidate seeds $\SSS$, scalar gradient accumulator $\cA$, \underline{and their probabilities $\pb$}. \;
  \For{each round $r = 1, 2, \ldots, T$}{
      \For{each client $i \in$ activate clients $\CC$ \textbf{in parallel}}{
        $\HH_i \!\leftarrow\!$ \FClientTraining{$\SSS$, $\cA$, \underline{$\pb$}, $i$}  \tcp*{\ding{192} in Figure \ref{pic-framework}}
        \For{$( s_j, \hat{\varrho}_j ) \ \in \HH_i$ }{
          $a_j = a_j + c_i \cdot \hat{\varrho}_j$  \tcp*{\ding{196} in Figure \ref{pic-framework}}
        }
      }
      \underline{compute the seed importance and then the probability $\pb$ as Equation \eqref{eq-seed-importance}} \tcp*{\ding{197} in Figure \ref{pic-framework}}
  }
  \KwRet{the fine-tuned global model $\wb^T$, which is calculated with $\wb^0$ as the initial point based on $\cA$, as Equation \eqref{eq-obtain-latest-model}}
  
  % \vspace{0.05cm}\hspace{-2em}
  \vspace{0.3cm}
  \Fn{\FClientTraining{$\SSS$, $\cA$, \underline{$\pb$}, $i$}}{
  calculate the latest global model with $\wb^0$ as the initial point based on $\cA$, as Equation \eqref{eq-obtain-latest-model}  \tcp*{\ding{193} in Figure \ref{pic-framework}}
  \For{each local step $t = 1, 2, \ldots, \tau$}{
    sample a data instance $\xb$ from local dataset $\cD_i$, a seed $s_j$ from $\SSS$ \underline{based on $\pb$}, then generate a perturbation $\zb_j$ with $s_j$ as the random seed \tcp*{\ding{194}-1 in Figure \ref{pic-framework}}
    $\hat{\varrho}_j = \frac{\cL(\wb + \epsilon\zb_j;\xb) - \cL(\wb - \epsilon\zb_j;\xb)}{2\epsilon}$ \;
    $\wb_{t+1} = $ \FUpdateModel{$\wb_t$, $s_j$, $\hat{\varrho}_j$}  \;
    stage $( s_j, \hat{\varrho}_j )$ into $\HH_i$ \tcp*{\ding{194}-2 in Figure \ref{pic-framework}}
  }
    \KwRet{$\HH_i$ to the server}  \tcp*{\ding{195}-1 in Figure \ref{pic-framework}}
  }
  % \vspace{0.05cm}\hspace{-2em}
  \vspace{0.3cm}
  \Fn{\FUpdateModel{$\wb$, $s$, $\hat{\varrho}$}}{
    sample perturbation $\zb \in \RR^{d}$ with random seed $s$ \;
    \KwRet{$\wb - \eta \cdot \hat{\varrho} \cdot \zb$}
  }
\end{algorithm}
Algorithm \ref{algo:main_algo} summarizes the main processes of \app. 
For ease of comparison, we also include the processes and components that are only required by \apppro in Algorithm \ref{algo:main_algo}, which are underlined and freely detachable as needed. 

%% file: sections/10_appendix_assumption.tex
\section{Detailed Assumptions of Theorem \ref{theorem-zoo-one-point-convergence}}
\label{sec-appendix-assumption}
We detail the assumptions made by \citet{fang2022communication} which are necessary conditions for deriving the convergence of ZOO-based FL with a one-point estimator, as claimed in Theorem \ref{theorem-zoo-one-point-convergence}.
We have also made variable substitutions to facilitate understanding within the context of our work.

\begin{assumption}
  (Loss Boundary.)
  The global loss $f(\wb)$ defined in Equation \eqref{eq-fl-optimization} is lower bounded by $f_*$, thus we have
  \begin{equation*}
      f(\wb) \geq f_* > -\infty.
  \end{equation*}
  \label{assumption-loss-boundary}
\end{assumption}

Before presenting Assumption \ref{assumption-L-smooth} and Assumption \ref{assumption-second-order-moment}, we define the expected loss $f_i(\wb)$ of model $\wb$ on the $i$-th client's local dataset $\cD_i$ as $f_i(\wb) \triangleq \EE_{\xb \sim \cD_i}\left[ \cL_i(\wb; \xb)\right]$.
\begin{assumption}
  (Objective Smoothness.)
  $\cL_i(\wb;\xb)$, $f_i(\wb)$ and $f(\wb)$ are all $L$-smooth, i.e., for any $\wb \in \RR^d$ and $\wb' \in \RR^d$, we have
  \begin{equation*}
      \left\| \nabla f_i(\wb') - \nabla f_i(\wb)  \right\|  \leq L\left\| \wb' - \wb \right\|, \forall i,
  \end{equation*}
  \begin{equation*}
      f(\wb') \leq f(\wb) + \langle \nabla f(\wb), \wb' - \wb \rangle + \frac{L}{2} \left\| \wb' - \wb \right\|^2.
  \end{equation*}
  \label{assumption-L-smooth}
\end{assumption}

\begin{assumption}
  (Boundary of the Second-Order Gradient Moment.)
  The second-order moment of stochastic gradient $\nabla_{\wb} \cL(\wb, \xb)$ satisfies 
  \begin{equation*}
      \EE_{\xb}\left \| \nabla_{\wb} \cL_i(\wb;\xb) \right \|^2 \leq c_g \left\| \nabla f_i(\wb) \right \|^2 + \sigma_g^2 , \forall \wb \in \RR^d,
  \end{equation*}
  where $c_g\geq 1$.
  \label{assumption-second-order-moment}
\end{assumption}

\begin{assumption}
  (Local-Global Gradient Dissimilarity Boundary.)
  The gradient dissimilarity between the local loss evaluated at each client and the global loss defined in Equation \eqref{eq-fl-optimization} is bounded as 
  \begin{equation*}
      \left \| \nabla f(\wb) - \nabla f_i(\wb)  \right \|^2 \leq c_h \left\| \nabla f(\wb) \right \|^2\ + \sigma_h^2, \forall \wb \in \RR^d,
  \end{equation*}
  where $c_h$ is a positive constant.
  \label{assumption-gradient-dissimilarity-boundary}
\end{assumption}

Assumptions \ref{assumption-loss-boundary}, \ref{assumption-L-smooth} and \ref{assumption-second-order-moment} are commonly employed in the analysis of stochastic optimizations \cite{fang2022communication}, and Assumption \ref{assumption-gradient-dissimilarity-boundary} can capture the extent of statistical heterogeneity inherent in the client-side data distribution within FL settings.
Similar assumptions have also been employed in existing research to analyze FL convergence in non-IID scenarios \cite{li2020fedprox,li2020convergence,ling2024convergence}. 
Theorem \ref{theorem-zoo-one-point-convergence} corresponds to Equation (12) in the work of \citet{fang2022communication}, with certain lower-order terms omitted for brevity.

%% file: sections/11_appendix_detailed_proofs.tex
\section{Detailed Proofs}
\label{sec-appendix-detailed-proofs}
\subsection{Proof of Lemma \ref{lemma-mean-of-perturbations}}
\label{subsec-appendix-proof-mean-perturbations}
Recall the two formal sampling stage of \app in Section \ref{subsubsec-theoretical-support}, in the first sampling stage, we have $K$ random variables $\mathcal{Z}' = {Z'_1,Z'_2,Z'_3,\ldots,Z'_K}$ that randomly sampled from $\mathcal{N}(0, 1)$. 
By Chebyshev’s inequality, $\forall \varepsilon>0$, we have
\begin{equation}
    \text{Pr}[|\frac{1}{K}(Z'_1+Z'_2+\cdots+Z'_K)|\ge\varepsilon]\le\frac{4}{K\varepsilon^2}.
\end{equation}

In the second sampling stage, we have $Z_1,Z_2,\ldots,Z_\Gamma$ that randomly and uniformly sampled from $\mathcal{Z}'$.
Thus $\forall Z_i$, we have $\min(\mathcal{Z}') \leq Z_i \leq \max(\mathcal{Z}')$. 
Let $\mu = \mathbb{E}[Z_i]$, according to Hoeffding's inequality, $\forall \varepsilon'>0$, $S_\Gamma = Z_1 + Z_2 + \ldots + Z_\Gamma$ satisfies
\begin{equation}
    \text{Pr}[|S_\Gamma-\mu\Gamma|\ge \varepsilon']\le 2\exp(-\frac{2\varepsilon'^2}{\sum_{i=1}^\Gamma [\max(\mathcal{Z}') - \min(\mathcal{Z})]^2}).
\end{equation}

Let $\bar{S}_\Gamma=\frac{Z_1 + Z_2 + \ldots + Z_\Gamma}{\Gamma}$. 
Thus $\forall \varepsilon>0$, we have 
\begin{equation}
    \text{Pr}[|\bar{S}_\Gamma-\mu|\ge \varepsilon]\le 2\exp(-\frac{2\Gamma\varepsilon^2}{[\max(\mathcal{Z}') - \min(\mathcal{Z}')]^2}).
\end{equation}

Finally, $\forall \varepsilon>0$, we have
\begin{equation}
\begin{aligned} 
    \text{Pr}[|\bar{S}_\Gamma-0|\ge \varepsilon] & \le \text{Pr}[|\mu-0|\ge\frac{\varepsilon}{2}] + \text{Pr}[|\bar{S}_\Gamma-\mu|\ge\frac{\varepsilon}{2}]\\ &\le \frac{4}{K\varepsilon^2} + 2\exp(-\frac{\Gamma\varepsilon^2}{2[\max(\mathcal{Z}') - \min(\mathcal{Z}')]^2}). 
\end{aligned}
\end{equation}
$\blacksquare$

\subsection{Proof of Theorem \ref{theorem-gradient-estimation-error}}
\label{subsec-appendix-proof-gradient-estimation-error}
From Lemma \ref{lemma-mean-of-perturbations}, the perturbations follow $\varepsilon$-mean in \app.
We denote the gradient estimated by the two-point gradient estimator by $\hat{\mathbf{g}}_1$, calculated as
\begin{equation}
    \hat{\mathbf{g}}_1 = \frac{\mathbf{z} + \overrightarrow{\varepsilon}}{2\epsilon}[\mathcal{L}(\mathbf{w}+ \epsilon \mathbf{z} + \epsilon \overrightarrow{\varepsilon}) - \mathcal{L}(\mathbf{w} - \epsilon \mathbf{z} - \epsilon \overrightarrow{\varepsilon})],
\end{equation}
and the gradient estimated by the two-point gradient estimator without seed restrictions is defined as
\begin{equation}
    \hat{\mathbf{g}}_0 = \frac{\mathbf{z}}{2\epsilon}[\mathcal{L}(\mathbf{w} + \epsilon \mathbf{z}) - \mathcal{L}(\mathbf{w} - \epsilon \mathbf{z})].
\end{equation}
Thus,
\begin{equation}
\begin{aligned} 
    \hat{\mathbf{g}}_1 - \hat{\mathbf{g}}_0 = & \frac{\mathbf{z}}{2\epsilon}[\mathcal{L}(\mathbf{w} + \epsilon\mathbf{z} + \epsilon\overrightarrow{\varepsilon}) - \mathcal{L}(\mathbf{w} + \epsilon\mathbf{z})] - \frac{\mathbf{z}}{2\epsilon}[\mathcal{L}(\mathbf{w} - \epsilon \mathbf{z} - \epsilon \overrightarrow{\varepsilon}) - \mathcal{L}(\mathbf{w} - \epsilon \mathbf{z}) ] + \\ & \frac{\overrightarrow{\varepsilon}}{2\epsilon}[\mathcal{L}(\mathbf{w}+ \epsilon \mathbf{z} + \epsilon \overrightarrow{\varepsilon}) - \mathcal{L}(\mathbf{w} - \epsilon \mathbf{z} - \epsilon \overrightarrow{\varepsilon})].
\end{aligned}
\end{equation}

According to the $L$-smoothness of objective function $\mathcal{L}$ (Assumption \ref{assumption-L-smooth}), we have
\begin{equation}
    \left\|\mathcal{L}(\mathbf{w} + \epsilon \mathbf{z} + \epsilon \overrightarrow{\varepsilon}) - \mathcal{L}(\mathbf{w} + \epsilon \mathbf{z}) \right\| \leq L\left\| \epsilon \overrightarrow{\varepsilon} \right\|,
\end{equation}
and
\begin{equation}
    \left\|\mathcal{L}(\mathbf{w} - \epsilon \mathbf{z} - \epsilon \overrightarrow{\varepsilon}) - \mathcal{L}(\mathbf{w} - \epsilon \mathbf{z}) \right\| \leq L\left\| \epsilon \overrightarrow{\varepsilon} \right\|.
\end{equation}
Therefore,
\begin{equation}
\begin{aligned} 
    \left\|\hat{\mathbf{g}}_1 - \hat{\mathbf{g}}_0\right\| \leq & \left\|\frac{\mathbf{z}}{2\epsilon}\right\|L\left\|\epsilon\overrightarrow{\varepsilon}\right\| + \left\|\frac{\mathbf{z}}{2\epsilon}\right\|L\left\|\epsilon\overrightarrow{\varepsilon}\right\| + \left\|\frac{\overrightarrow{\varepsilon}}{2\epsilon}\right\|L\left\|2\epsilon\mathbf{z} + 2\epsilon\overrightarrow{\varepsilon} \right\| \\ = & L\left\| \mathbf{z}\right\|\left\| \overrightarrow{\varepsilon}\right\| + L\left\| \overrightarrow{\varepsilon}\right\|\left\| \mathbf{z} + \overrightarrow{\varepsilon}\right\| \\ \leq & 2L\left\| \mathbf{z}\right\|\left\| \overrightarrow{\varepsilon}\right\| + L\left\| \overrightarrow{\varepsilon}\right\|^2. 
\end{aligned}
\end{equation}
$\blacksquare$

%% file: sections/12_appendix_another_explanation.tex
\section{Additional Analytical Supports to Principles on the Selection of $K$}
\label{sec-appendix-supports-principles}

\subsection{Additional Analytical Support to Principle \ref{principle-not-too-few-seeds}}
\label{subsec-appendix-supports-principles-not-too-few}
The federated fine-tuning process can be formally modeled as an optimization problem that seeks a model variation from $\wb^0$ to an ideally optimal model $\wb^*$, with the combination of $K$ perturbations, as
\begin{equation}
  \min_{\GG} \ \ \left \| \wb^0 - \eta \cdot 
  \begin{bmatrix}\zb_1, \ldots, \zb_K\end{bmatrix}\GG - \wb^* \right \|.
  \label{eq-definition-combination-optimization}
\end{equation}
It is important to note that this definition serves to provide an alternative perspective on FL, and it is not a formulation that can be solved outright because the ideally optimal model weights $\wb^*$ are not known.
From Equation \eqref{eq-definition-combination-optimization}, FL processes can be viewed as advancing the model towards an approximate optimal solution in an iterative manner.

With this formulation, matrix $\ZZ = \left[\zb_1, \ldots, \zb_K\right]$ can be regarded as the constraints of this problem. 
When the constraints are insufficient to uniquely determine a solution, i.e., the rank of the system is low, the solution space becomes larger and there are multiple or even infinitely many possible solutions, causing greater difficulty in finding the optimal solution.
Since high-dimensional vectors sampled from a Gaussian distribution are typically orthogonal, considering the dimension $d$ of LLM $\wb$ is usually very high such that $d \gg K$, the rank of $\ZZ = \left[\zb_1, \ldots, \zb_K\right]$ is typically $K$.
Therefore, usually the larger the value of $K$ is, the better the optimization problem defined in Equation \eqref{eq-definition-combination-optimization} could be finally solved.
Taking an extreme example, if $K=1$, the optimization problem defined in Equation \eqref{eq-definition-combination-optimization} may be fundamentally unoptimizable. 
Thus, $\overset{\leftharpoonup }{K}$ theoretically exists so that Principle \ref{principle-not-too-few-seeds} holds. 

\subsection{Additional Analytical Support to Principle \ref{principle-not-too-much-seeds}}
\label{subsec-appendix-supports-principles-not-too-much}
Recall Equation \eqref{eq-zeroth-one-point} and Theorem \ref{theorem-zoo-one-point-convergence}, each $\zb_{j}$ is independently randomly generated from $\cN\left(\mathbf{0}, \Ib_d\right)$.
Without introducing system error, Equation \eqref{eq-zeroth-one-point} can be rewrite as
\begin{equation}
  \hat{\gb}^r_{i,t} = \frac{1}{b_2 b_1} \sum_{j=1}^{b_2} \sum_{b=1}^{b_1} \frac{\left[\cL(\wb^r_{i,t} + \epsilon\zb_{j};\xb_{b}) - \cL(\wb^r_{i,t};\xb_{b})\right]}{\epsilon} \zb_{j}.
  \label{eq-zeroth-one-point-rewrite}
\end{equation}
In this new formulation, the FL process of \app and \apppro can be regarded as: 
for each perturbation $\zb$, computing the step size that guides the model's progress in the direction of $\zb$. 
This is achieved through local training on several data instances.
Given that the total number of update steps is fixed at $\tau r m$, each of the $K$ seeds is sampled on average $\frac{\tau r m}{K}$ times in \app.
When there are fewer candidate seeds, more data instances are used to determine the step size on the direction of each perturbation, magnifying the batch size probabilistically. 
Besides, when the cardinality of candidate seeds further increases, it does not change the optimal solution area, but enlarges the optimization space and thus increases the difficulty of random searching. 
From the above analysis, $\overset{\rightharpoonup }{K}$ theoretically exists so that Principle \ref{principle-not-too-much-seeds} holds. 

%% file: sections/13_appendix_implementation.tex
\section{Implementation Details}
\label{sec-appendix-implementations}
For better reproducibility, in this section, we provide the detailed implementations of our approaches and the baselines.
Some of the experimental settings have already been mentioned in Section \ref{subsec-exp-setup} and are not reiterated here.

\subsection{Datasets \& Evaluation Metrics}
\datani contains a large collection of tasks and their natural language instructions, and provides the splits for training and test, respectively. 
We utilize version \texttt{v2.8} of the dataset and adopt its \texttt{default} split, which includes 756 tasks for training and 119 tasks for testing, each with a unique task definition. 
Since \datani is very large, we conduct our experiments on a subset of it.
Specifically, we randomly sample 20\% of the data instances for each training task and 2\% of the data instances for each test task. 
After the above subsampling, each training task with no less than 20 data instances is treated as a unique client, forming an FL system with 738 clients. 
The test tasks are retained on the server for evaluation purposes.

\begin{table*}[t]
\begin{minipage}{\textwidth}
    \centering
    \caption{Prompt template for \datani.}
    \begin{tabularx}{0.9\linewidth}{p{4.6cm}X}
    \toprule
    \textbf{Attributes of data instances} & \textbf{Prompt} \\ 
    \midrule
    1. \texttt{Definition} 
    \newline
    2. \texttt{input}
    & Below is an instruction that describes a task, paired with an input that provides further context. Write a response that appropriately completes the request. \newline\newline \#\#\# Instruction: \{\texttt{Definition}\} \newline\newline \#\#\# Input: \{\texttt{input}\} \newline\newline \#\#\# Response: \\ 
    \addlinespace
    \bottomrule
    \end{tabularx}
    \label{tab-prompt-template-instruction}
\end{minipage}
\vspace{2em}
\begin{minipage}{\textwidth}
    \centering
    \caption{Prompt templates for \datadolly, which vary slightly depending on whether the data instance has \texttt{context}.}
    \begin{tabularx}{0.9\linewidth}{p{4.6cm}X}
    \toprule
    \textbf{Attributes of data instances} & \textbf{Prompt} \\ 
    \midrule
    Data instances with context:
    \newline
    1. \texttt{instruction} 
    \newline
    2. \texttt{context}
    & Below is an instruction that describes a task, paired with an input that provides further context. Write a response that appropriately completes the request. \newline\newline \#\#\# Instruction: \{\texttt{instruction}\} \newline\newline \#\#\# Input: \{\texttt{context}\} \newline\newline \#\#\# Response: \\ 
    \addlinespace
    \hline
    \addlinespace
    Data instances without context:
    \newline
    1. \texttt{instruction}
    & Below is an instruction that describes a task, paired with an input that provides further context. Write a response that appropriately completes the request. \newline\newline \#\#\# Instruction: \{\texttt{instruction}\}\newline\newline \#\#\# Response: \\ 
    \addlinespace
    \bottomrule
    \end{tabularx}
    \label{tab-prompt-template-dolly}
\end{minipage}
\end{table*}

\datadolly provides 15,015 data instances within 8 tasks. 
For \datadolly, we reserve the last task for evaluation and use the remaining tasks for training.
Experiments on \datadolly are conducted with 200 clients.
Note that each task in \datadolly contains a \texttt{category} attribute with a different value. 
Thus, we can allocate data instances to the 200 clients via Dirichlet distribution \cite{li2022federated} with the \texttt{category} as the labels. 
To build non-IID scenarios with varying degrees of label distribution skew \cite{chen2023efficient}, we perform data partitioning via Dirichlet distribution with $\alpha=0.5$ and $\alpha=5.0$, respectively, where a lower $\alpha$ indicates a higher degree of label distribution skew.

When feeding data instances into LLMs, we directly adopt the prompt template from Alpaca \cite{alpaca} following \citet{kuang2023federatedscope-LLM,zhang2023fedit}. 
The utilization of the prompt template is detailed in Appendix \ref{subsec-appendix-implementations-prompt-template}.
The maximum token length is set to 1,024, and data instances exceeding this length are ignored.
To reduce the impact of extraneous variables on the experiment results as much as possible, we uniformly employ greedy decoding to generate the responses during evaluation following \citet{malladi2023mezo,borzunov2023distributed}.

\subsection{Prompt Template}
\label{subsec-appendix-implementations-prompt-template}
In our experiments, data instances are wrapped to prompts before processing by LLMs. 
We directly apply the template provided by Alpaca \cite{alpaca} to the datasets in our experiments.
For better reproducibility, we present how we fill the fields in the template with the attributes of data instances in Tables \ref{tab-prompt-template-instruction} and \ref{tab-prompt-template-dolly}.

\subsection{Experimental Platforms}
\label{sec-appendix-implementations-environments}
We implement these approaches by PyTorch \cite{paszke2019pytorch} \texttt{v2.0.1} with \texttt{PEFT} \cite{peft} \texttt{v0.4.0} and \texttt{Transformers} \cite{wolf-etal-2020-transformers} \texttt{v4.31.0}. 
Extensive performance evaluations on Rouge-L scores with \modeldatajuicer and \modelllama are conducted on a platform with an NVIDIA RTX 3090 GPU and a platform with an NVIDIA A100 GPU, respectively, with the pre-trained LLMs loaded in 16-bit floating numbers. 
The devices used for time-related experimental results (Figure \ref{pic-intro-time-consumption}, Figure \ref{pic-time-pulling} and Figure \ref{pic-training-efficiency}) are all specified in the corresponding captions or sections.
Note that from Table \ref{tab-overhead}, \app and \apppro do not require as much memory as these platforms can provide, unlike the baselines based on BP. 
Therefore, this experimental setup is adopted to ensure consistency in the experimental environments among different approaches.

\subsection{Implementations}
Following \citet{kuang2023federatedscope-LLM} and \citet{malladi2023mezo}, all approaches perform local training with the batch size set to $1$ to reduce memory consumption.
Following \citet{kuang2023federatedscope-LLM}, BP-based approaches conduct local training with learning rate $\eta$ of $3\times 10^{-4}$, where the selected learning rate is searched from $[3\times 10^{-3}, 3\times 10^{-4}, 3\times 10^{-5}]$.
Among them, the number of virtual tokens in FedPTuning and FedPrompt are both set to 20, the type of reparameterization is set to ``MLP'' for FedPTuning following \citet{kuang2023federatedscope-LLM}, and the \texttt{rank} and \texttt{alpha} of LoRA adapters for both FedIT and FedIT-SGD are set to 8 and 16 respectively, following \citet{zhang2023fedit}. 
Following \citet{malladi2023mezo}, $\eta$ and $\epsilon$ of \app and \apppro are set to $3\times 10^{-7}$ and $5\times 10^{-4}$, respectively, unless stated otherwise. 
The selected learning rate of \app and \apppro is searched from $[3\times 10^{-5}, 3\times 10^{-6}, 3\times 10^{-7}, 1\times 10^{-7}]$.
The impacts of the two hyperparameters of \app and \apppro have been discussed in Section \ref{subsec-exp-parameter}.

Before starting the federated tuning, the server initializes the $K$ candidate seeds with integers uniformly and randomly sampled from [0, $10^{11}$). 
The aggregation weights of participating clients in each round are proportional to the scale of their private training set.

\subsection{Further Optimization on Updating Models}
We highly thanks an anonymous reviewer for providing a more efficient implementation.
The main differences between this implementation and ours lie in (1) the generation of perturbations ($\zb$), and (2) the updating of model parameters.
These differences are presented in Table \ref{tab-new-implementation}.
For the reproducibility of the Rouge-L scores in this work, we left the integration of this implementation in the future.
\begin{table}[!h]
\centering
\caption{Main differences between our implementation and a more efficient implementation.}
\label{tab-new-implementation}
\begin{tabularx}{\linewidth}{p{1.85cm}p{5.5cm}X}
    \toprule[1.0pt]
    Functionality & 
    Our Implementation &
    More Efficient Implementation \\
    \midrule[1.0pt]
    Perturbation \newline Generation & 
    \texttt{z = torch.normal(mean=0, std=1, size=param.size())} \newline 
    &
    1. \texttt{gen = torch.Generator()} \newline
    2. \texttt{z = torch.empty()} \newline
    3. \texttt{z.resize\_(param.size())} \newline
    4. \texttt{z.normal\_(mean=0, std=1, generator=gen)}
    \\
    \cmidrule{1-3}
    Updating \newline Model \newline Parameters 
    &
    \texttt{param.data = param.data - (lr * scalar\_grad) * z}
    &
    1. \texttt{scalar = lr * scalar\_grad} \newline
    2. \texttt{param.data.add\_(z, alpha=-scalar)}
    \\
    \bottomrule[1.0pt]
\end{tabularx}
\end{table}

%% file: sections/14_appendix_supplementary_exp.tex
\section{Supplementary Experiments}
\label{sec-appendix-exp}

\subsection{Experimental Support for Principle \ref{principle-not-too-much-seeds}}
\label{subsec-appendix-exp-marginal-gain}
\begin{figure*}[h]
  \centering
  \subfigure[\modeldatajuicer on \datani]{
    \includegraphics[width=0.316\linewidth]{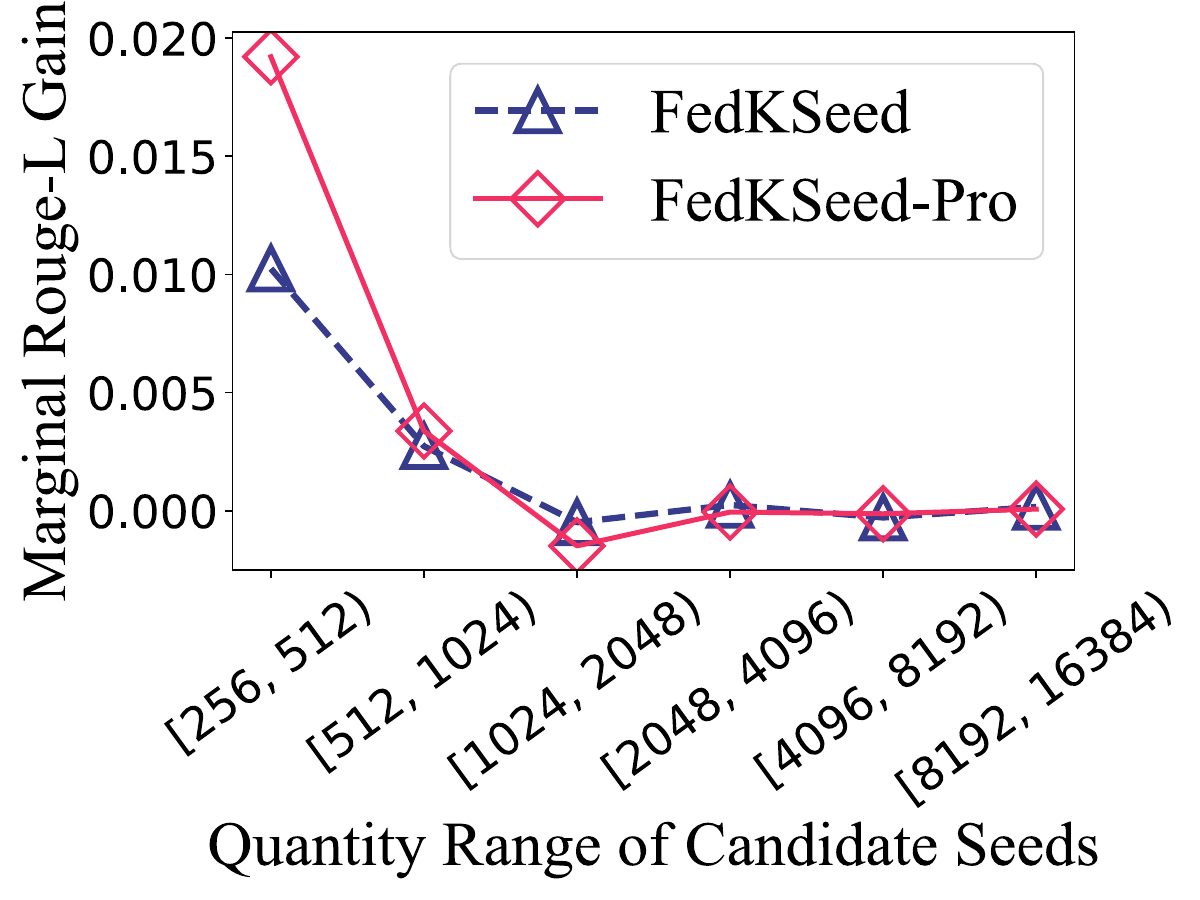}
  }
  \setcounter{subfigure}{2}
  \subfigure[\modeldatajuicer on \datadolly ($\alpha=0.5$)]{
    \includegraphics[width=0.316\linewidth]{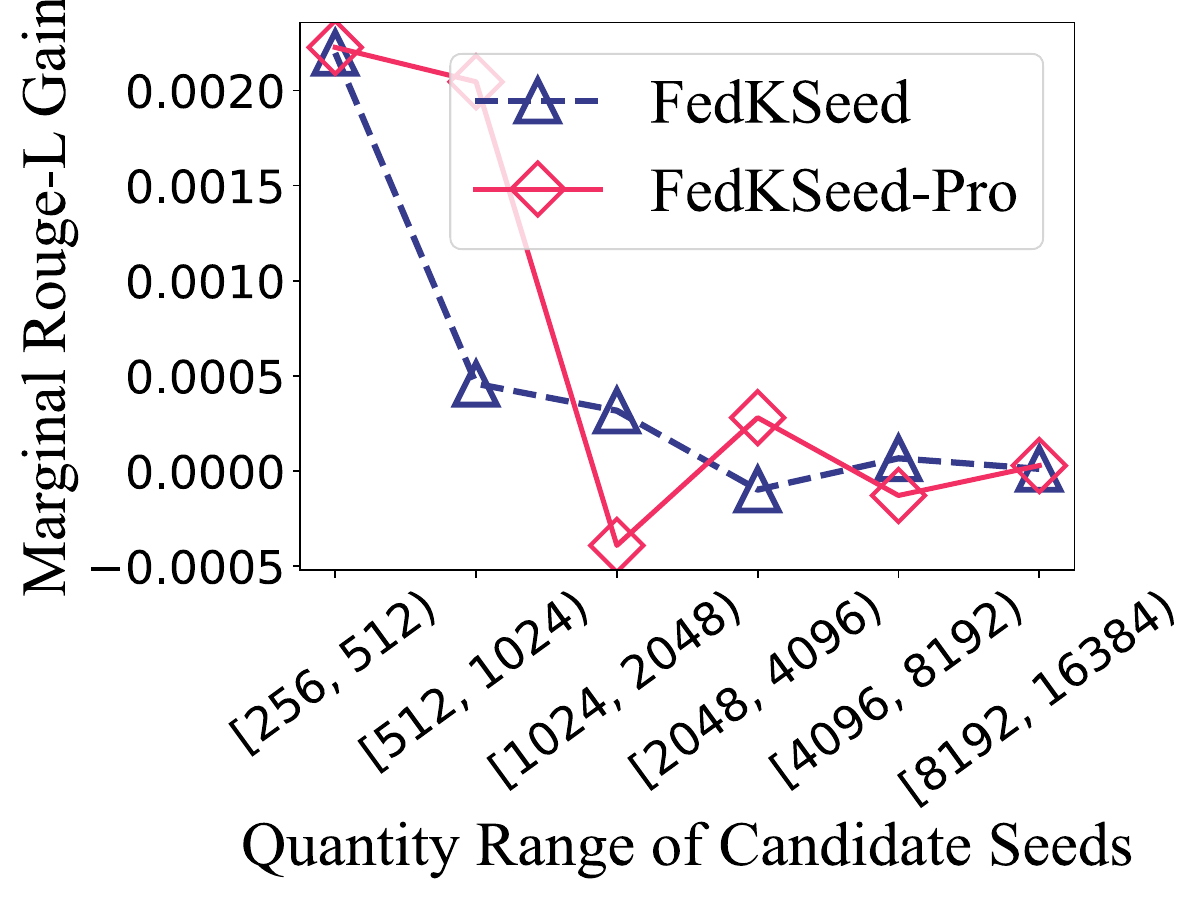}
  }
  \setcounter{subfigure}{4}
  \subfigure[\modeldatajuicer on \datadolly ($\alpha=5.0$)]{
    \includegraphics[width=0.316\linewidth]{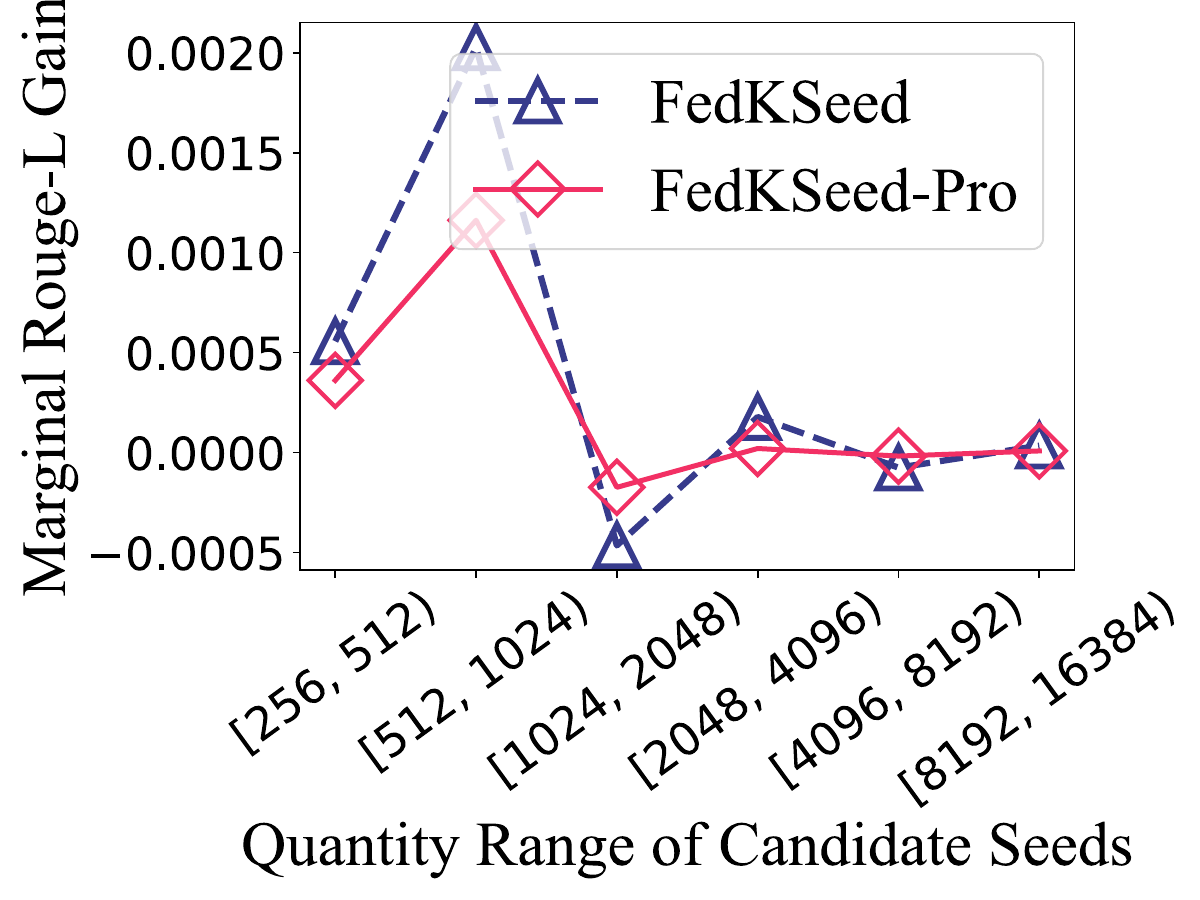}
  }
  \setcounter{subfigure}{1}
  \subfigure[\modelllama on \datani]{
    \includegraphics[width=0.316\linewidth]{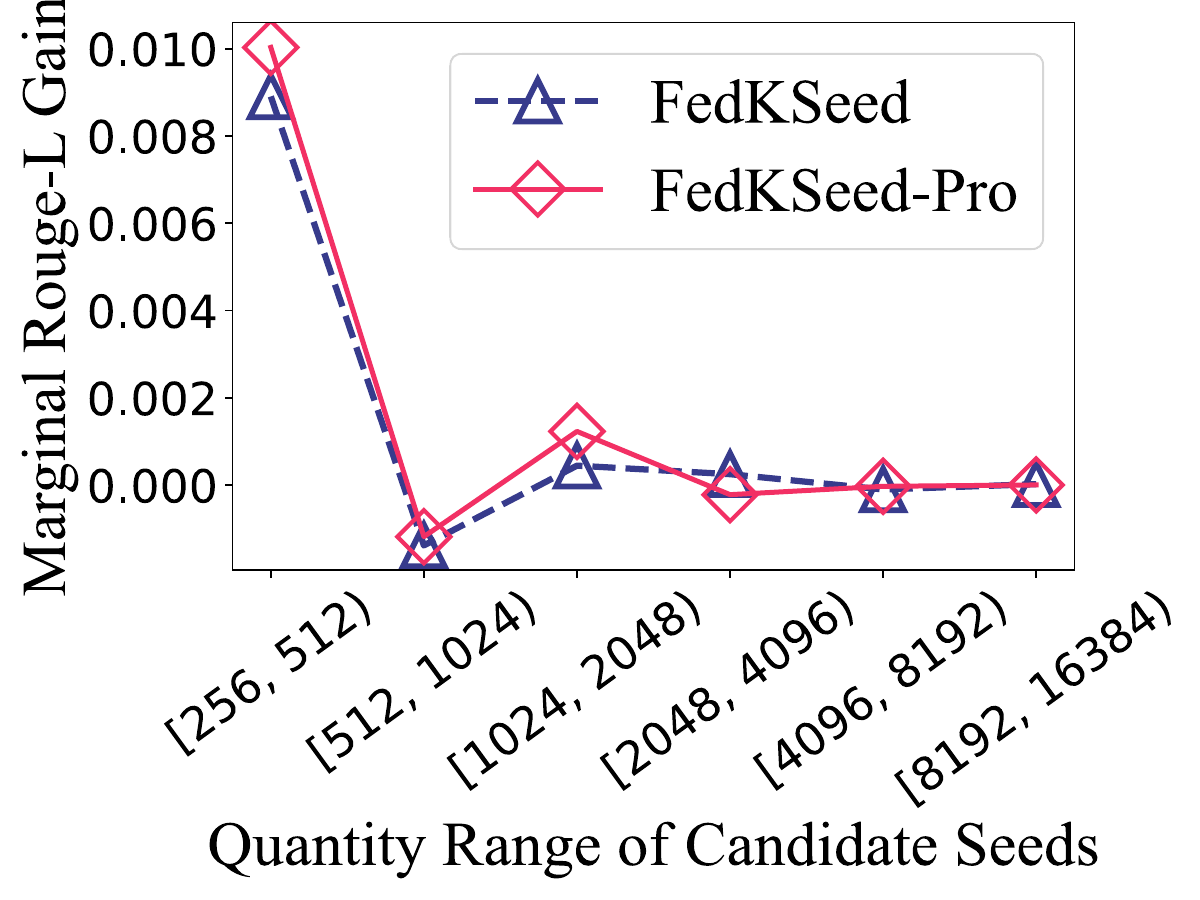}
  }
  \setcounter{subfigure}{3}
  \subfigure[\modelllama on \datadolly ($\alpha=0.5$)]{
    \includegraphics[width=0.316\linewidth]{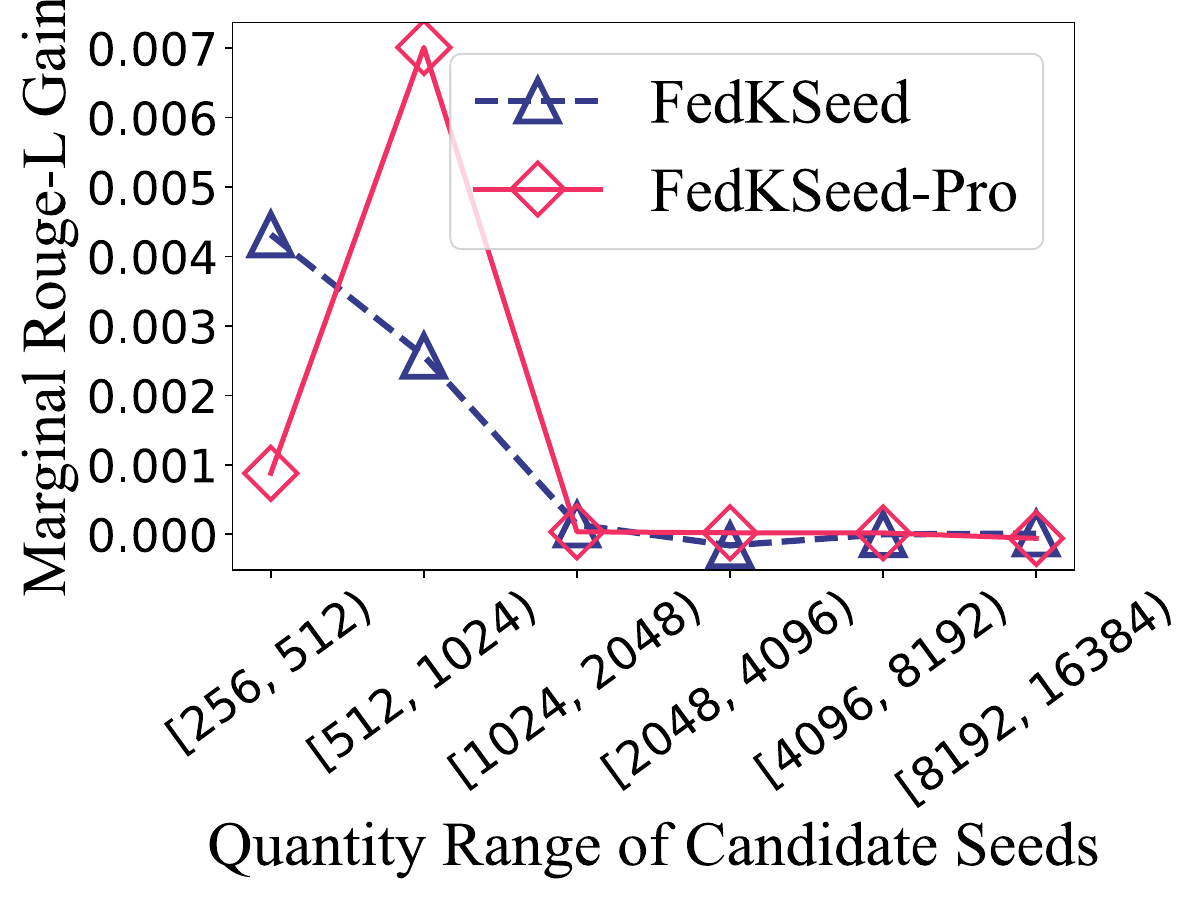}
  }
  \setcounter{subfigure}{5}
  \subfigure[\modelllama on \datadolly ($\alpha=5.0$)]{
    \includegraphics[width=0.316\linewidth]{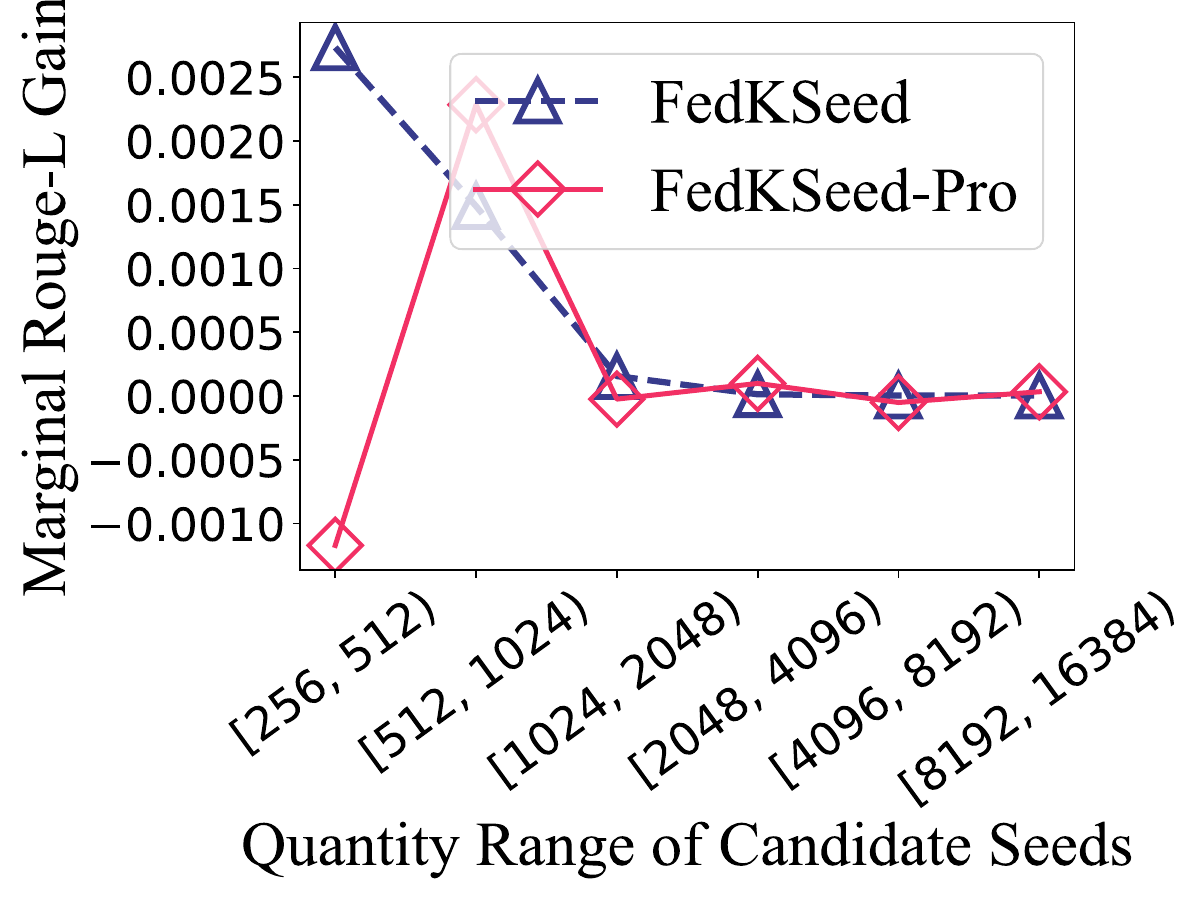}
  }
  \caption{Marginal improvement on Rouge-L obtained with each additional seed across different ranges of seed quantity by \app and \apppro, respectively.}
  \label{pic-marginal-gain-number-of-seeds}
\end{figure*}

In Section \ref{subsubsec-approach-selection-K}, we have provided Principles \ref{principle-not-too-few-seeds} and \ref{principle-not-too-much-seeds} to guide the determination of $K$. 
From Figure \ref{pic-performance-number-of-seeds} in Section \ref{subsec-exp-performance}, it can be observed that once the number of seeds exceeds the threshold $\overset{\rightharpoonup }{K}$, which is around 4096, additional seeds do not lead to improved accuracy. 
However, the Y-axis range in Figure \ref{pic-performance-number-of-seeds} may obscure the visibility of the trend outlined in Principle \ref{principle-not-too-much-seeds}. 
For better clarity, in Figure \ref{pic-marginal-gain-number-of-seeds}, we present the marginal improvement on Rouge-L obtained with each additional seed within different ranges of seed quantity by \app and \apppro in the six scenarios, respectively.

From Figure \ref{pic-marginal-gain-number-of-seeds}, we can find that when $K$ is lower than 1024, each additional seed yields a substantial and consistently greater than zero improvement in Rouge-L accuracy on average. 
In other words, within this range, reducing $K$ would result in a noticeable decrease in Rouge-L accuracy of the global model, which can also serve as empirical evidence for Principle \ref{principle-not-too-few-seeds}.
When $K$ falls within the range of [1024, 4096), the marginal gain in average accuracy for each additional seed is negligible in certain scenarios.
When $K$ exceeds 4096, it becomes evident that additional seeds yield negligible marginal gains in accuracy across various scenarios.
Thus, we can find that $\overset{\rightharpoonup }{K}$ for \app should be 4096 such that when $K > \overset{\rightharpoonup }{K}$, there is no upward trend in the Rouge-L of the global model with the increase of $K$, therefore Principle \ref{principle-not-too-much-seeds} holds.
Considering that increasing $K$ will incur additional time costs for clients to synchronize the global model as Equation \eqref{eq-obtain-latest-model}, we should choose a smaller $K$ value within a reasonable range.

Based on the experimental results and analysis presented above, and in conjunction with the analytical support for Principle \ref{principle-not-too-much-seeds} provided in Section \ref{subsubsec-approach-selection-K} and Appendix \ref{subsec-appendix-supports-principles-not-too-much}, Principle \ref{principle-not-too-much-seeds} holds. 

It should be noted that Figure \ref{pic-marginal-gain-number-of-seeds} demonstrates the marginal improvements, not the actual accuracy itself. 
In Figure \ref{pic-marginal-gain-number-of-seeds}, the marginal gains for each seed of \app and \apppro are relatively similar. 
But as illustrated in Figure \ref{pic-performance-number-of-seeds}, \apppro usually requires fewer candidate seeds to achieve the same level of accuracy compared to \app. 
Therefore, if we are not solely pursuing the highest accuracy, \apppro can achieve comparable results with fewer candidate seeds. 
It is for this purpose that when we record the results in Table \ref{tab-performance}, we make \apppro to use fewer candidate seeds.

\subsection{Convergence Study}
\label{subsec-appendix-exp-convergence}
\begin{figure*}[h]
  \centering
  \includegraphics[width=0.68\linewidth]{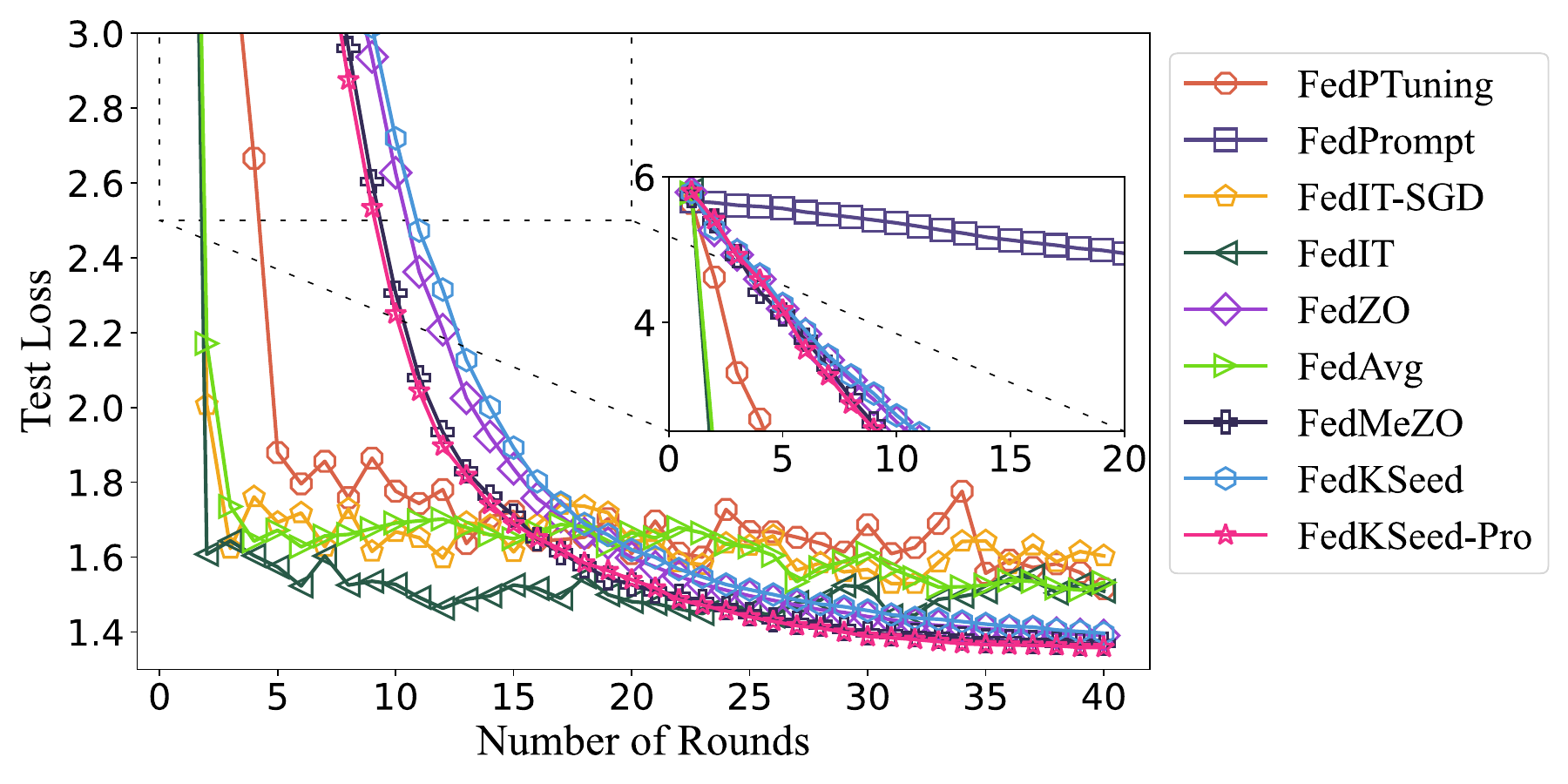}
  \caption{
    Convergence of the loss value on the test tasks obtained by \app, \apppro and the baselines with \modelllama on \datani.
  }
  \label{pic-convergence-all-instruct}
\end{figure*}

We illustrate the convergence curves obtained by \app, \apppro and the baselines with \modelllama on \datani in Figure \ref{pic-convergence-all-instruct}, where the experimental setup is aligned with that described in Section \ref{subsec-exp-setup} and Appendix \ref{sec-appendix-implementations}. 
Note that the loss values are calculated on the test set that is held by the server as described in Section \ref{subsec-exp-setup}.

\begin{figure}[h]
  \centering
  \subfigure[\modeldatajuicer on \datani]{
    \includegraphics[width=0.32\linewidth]{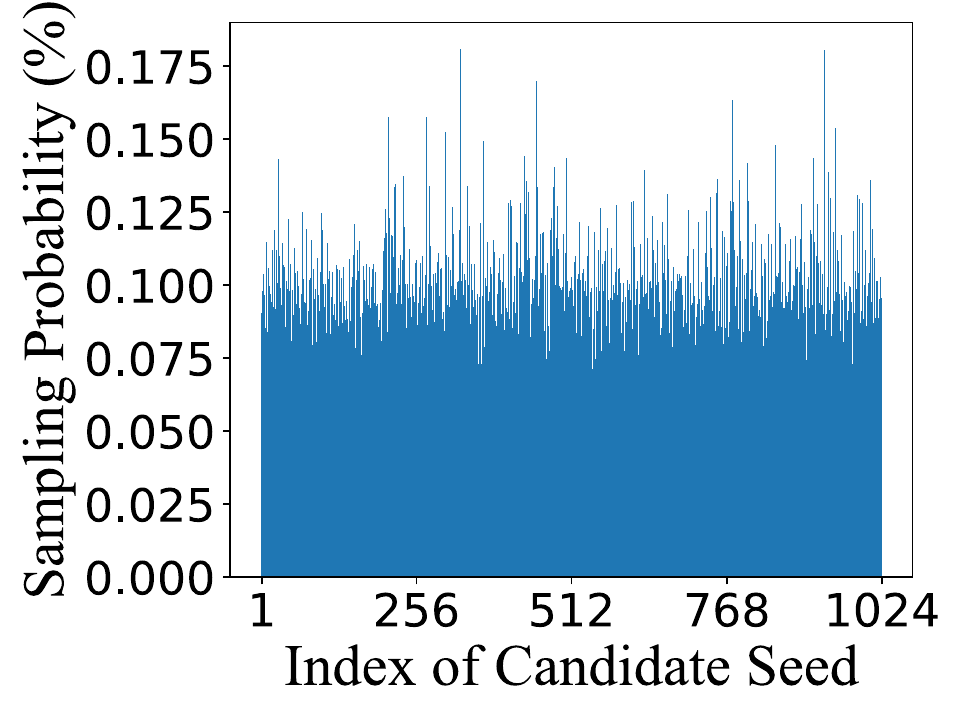}
  }
  \hspace{0.5cm}
  \subfigure[\modeldatajuicer on \datadolly ($\alpha=0.5$)]{
    \includegraphics[width=0.32\linewidth]{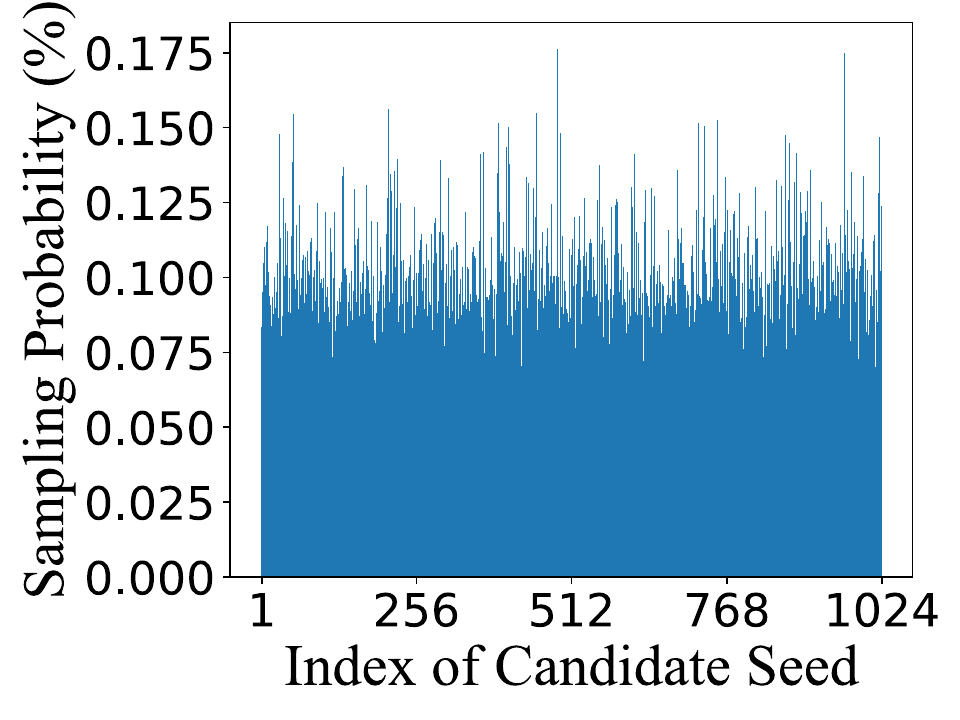}
  }
  \subfigure[\modelllama on \datani]{
    \includegraphics[width=0.32\linewidth]{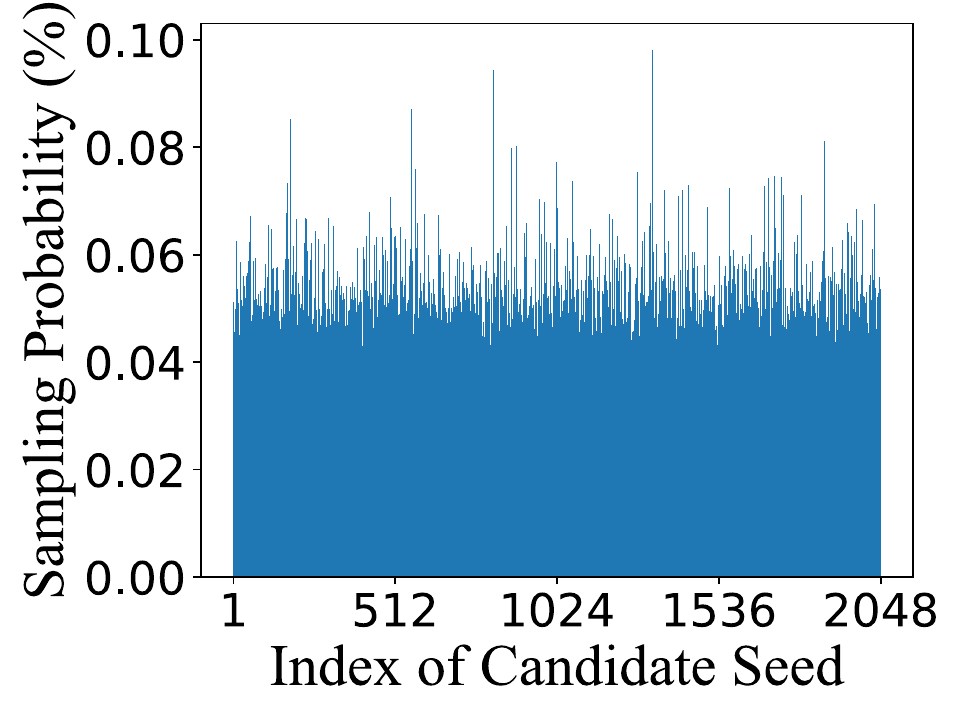}
  }
  \hspace{0.5cm}
  \subfigure[\modelllama on \datadolly ($\alpha=0.5$)]{
    \includegraphics[width=0.32\linewidth]{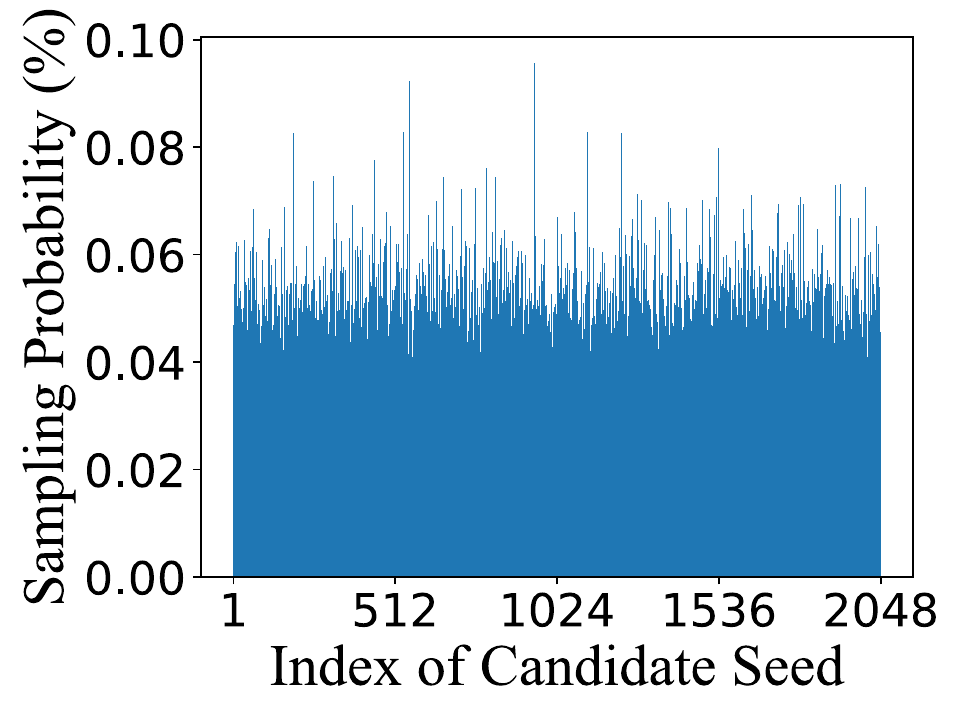}
  }
  \caption{Probabilities of candidate seeds calculated by \apppro after the last round.}
  \label{pic-seed-probabilities}
\end{figure}
\begin{figure*}[h]
  \centering
  \subfigure[\modeldatajuicer on \datani]{
    \includegraphics[width=0.486\linewidth]{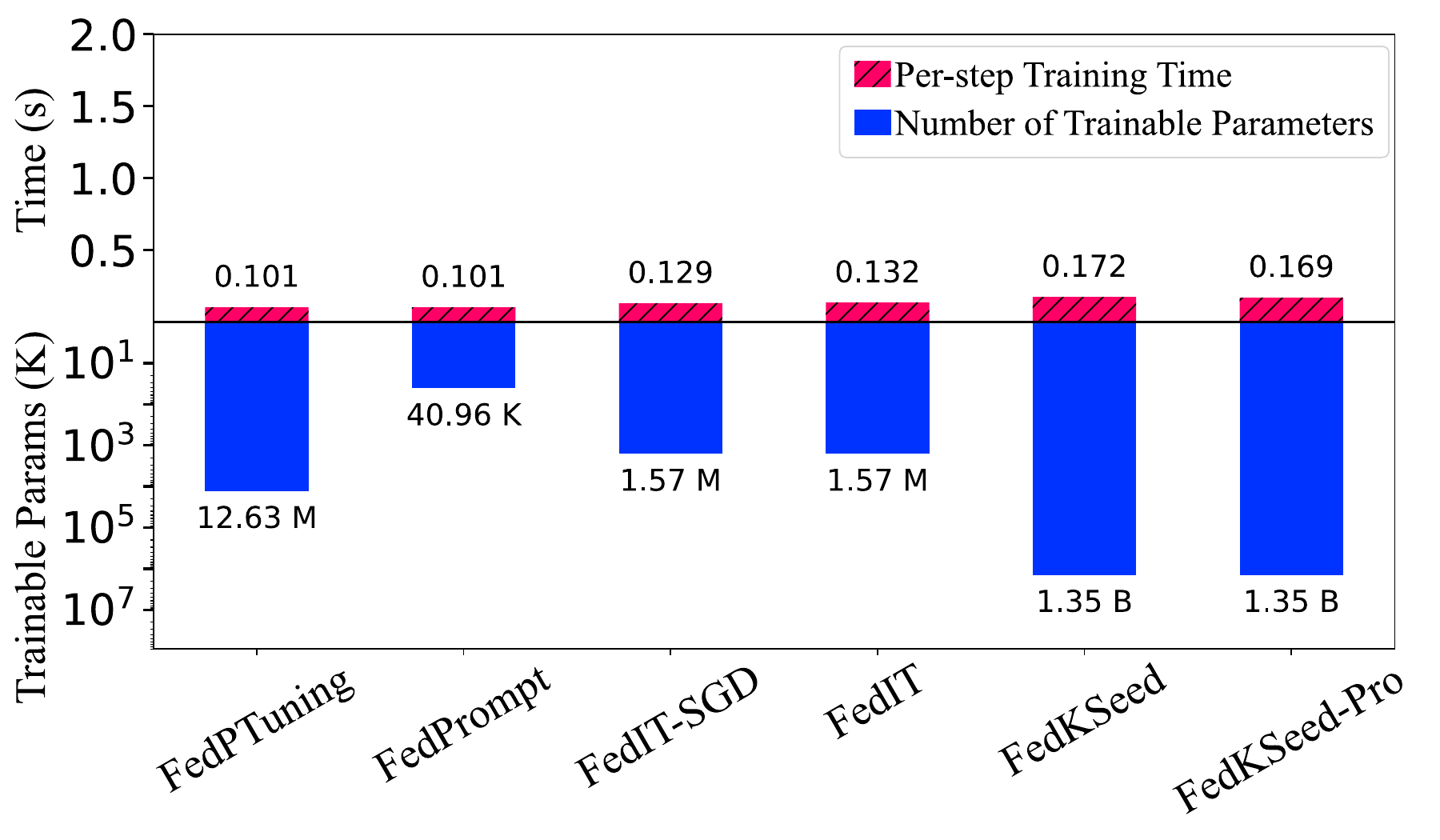}
  }
  \subfigure[\modelllama on \datani]{
    \includegraphics[width=0.486\linewidth]{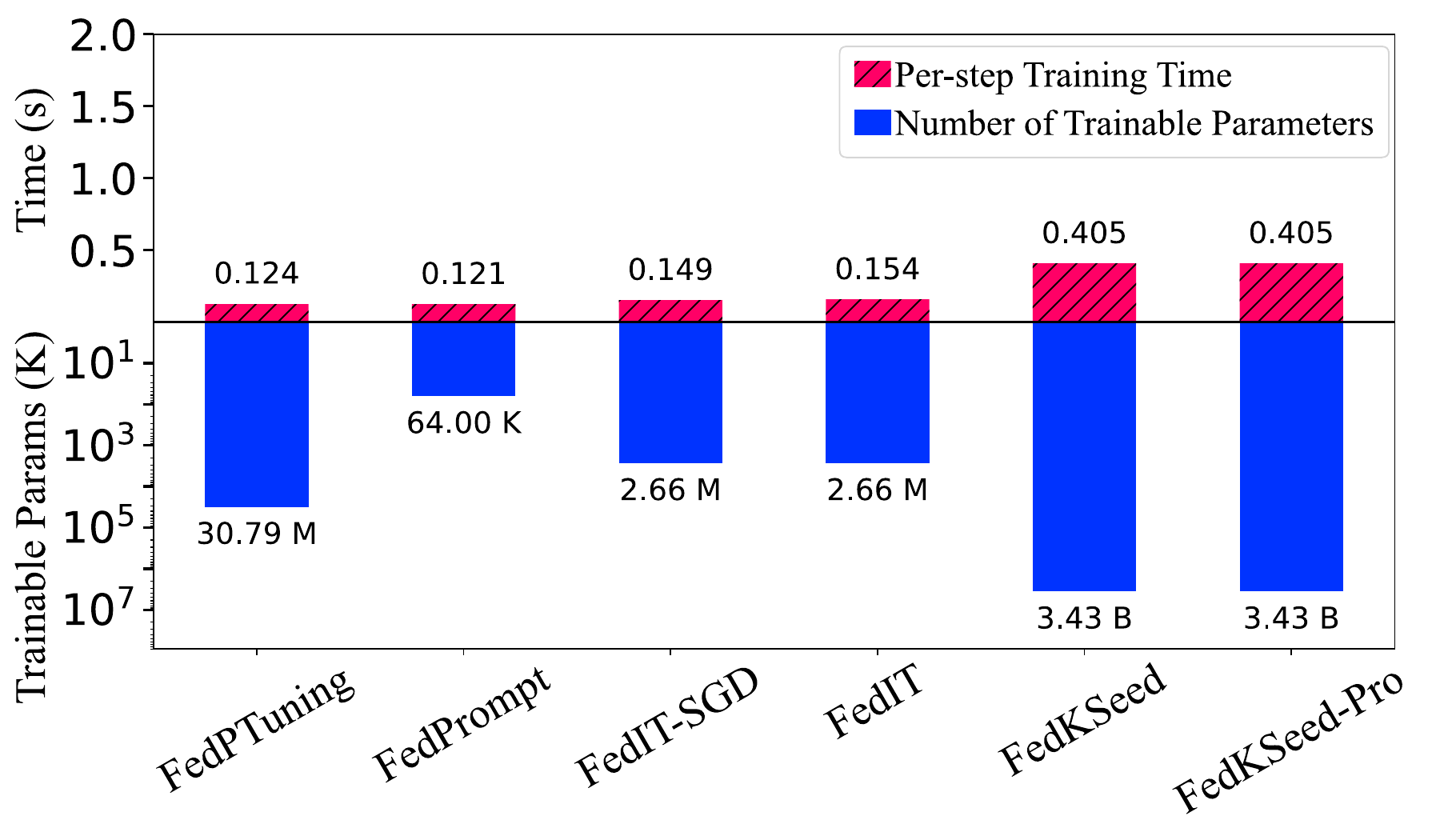}
  }
  \caption{
    Comparisons between the practical approaches in Table \ref{tab-performance} on per-step training time (measured with an NVIDIA V100 GPU) and on the number of trainable parameters.
  }
  \label{pic-training-efficiency}
\end{figure*}

Different from training from sketch, the pre-trained weights offer a good initial point for fine-tuning, leading to relatively quick convergence rates for these methods. 
As illustrated in Figure \ref{pic-convergence-all-instruct}, all of the approaches except for FedPrompt can converge within 40 rounds. 
It is worth noting that FedPrompt converges much more slowly, and as shown in Table \ref{tab-performance}, it also underperforms compared to the other baselines.
Although not as swift as the backpropagation (BP)-based approaches, the convergence rates for \app and \apppro are considerably faster than one might expect.
The reason is that fine-tuning is different from pre-training.  
As theoretically suggested by \citet{malladi2023mezo}, in centralized settings, based on adequate pre-training, the convergence of ZOO usually depends on the local effective rank instead of the number of full model parameters.
Therefore, it is reasonable and promising to apply ZOO to the federated full-parameter tuning of billion-sized LLMs.

\subsection{Illustrations of Seed Probabilities in \apppro}
\label{subsec-appendix-exp-seed-probabilities}
To demonstrate the variability in seed importance, we present the seed probabilities calculated in the last round by \app and \apppro on \datani and \datadolly ($\alpha=0.5$) with \modeldatajuicer and \modelllama, respectively in Figure \ref{pic-seed-probabilities}. 
As described in Section \ref{subsec-exp-setup}, we set $K$ to 1024 for \apppro with \modeldatajuicer and 2048 for \apppro with \modelllama, respectively. 

It can be observed that with non-uniform sampling, there is a significant disparity in the probabilities of seeds being sampled and a relatively small subset of the candidate seeds exhibits higher sampling probabilities.
Thus, we can conclude that the average amplitude corresponding to the scalar gradient of a seed is positively correlated with the importance of the corresponding perturbation to the model accuracy.

It is worth noting that quantifying the importance of seeds or perturbations through the amplitude of scalar gradients may be better than that through similarity evaluated by cosine or Euclidean distance, since vectors from high-dimensional Gaussian distributions tend to be orthogonal to each other, which results in the similarity distance of random perturbations with the same dimensionality as billion-sized LLMs typically being very close to 0, making it difficult to distinguish numerically.
Therefore, it is challenging for distance-based methods to assign significantly different importance to various seeds or perturbations. 
That is the reason why we adopt the amplitude of the scalar gradient to quantify the seed importance.

\subsection{Training Efficiency}
\label{subsec-appendix-exp-training-efficiency}
To have a clear view of the training efficiency of these approaches, we present their per-step training time together with the number of trainable parameters in Figure \ref{pic-training-efficiency}. 
Note that to ensure the comparability of the time consumptions across different LLMs with the same approach, these time consumptions are uniformly tested on the same platform equipped with an NVIDIA V100 GPU and an Intel(R) Xeon(R) Platinum 8163 CPU, with the required libraries the same as that have been described in Appendix \ref{sec-appendix-implementations-environments}. 

As shown in Figure \ref{pic-training-efficiency}, compared to the baselines, \app and \apppro only incur minimal additional per-step training time overhead. 
This limited extra per-step computational expense brings the benefit that allows for an expansion of trainable parameters by several orders of magnitude, thus improving the accuracy of the global model, as demonstrated in Table \ref{tab-performance}. 
Furthermore, the communication overhead and memory footprint have also been significantly reduced by \app and \apppro compared to the baselines as in Table \ref{tab-overhead}.
While \app and \apppro may take more time to perform one step of local training due to the significantly larger number of trainable parameters compared to PEFT-based techniques, the constraint on time consumption is not as strict as that on memory and communication. 
This is because the development of computing power has generally outpaced that of memory and communication resources.
Thus, the additional computational overheads of \app and \apppro may be worth it in comparison to the gains obtained by them in accuracy, memory footprint and communication consumption.

%% file: sections/15_appendix_calculation_overhead.tex
\section{Detailed Calculation of Communication Overhead}
\label{sec-appendix-calculation-overhead}
In this section, we provide a detailed analysis to demonstrate how the client-side per-round communication overheads of \app, \apppro and the baselines presented in Table \ref{tab-overhead} are calculated. 
Note that the communication overhead referred to here includes both the downlink overhead incurred from downloading data from the server and the uplink overhead associated with uploading data to the server. 

In Table \ref{tab-overhead}, for the baselines, we only account for the communication overhead brought about by the transmission of model parameters, ignoring the transmission of other information, including model architecture, request headers, etc., as these costs vary under different encoding schemes.
The model parameters in baselines and scalar gradients in \app and \apppro are encoded as 32-bit floating point numbers to prevent the loss of precision and overflow or underflow. 
For \app and \apppro, the random seeds are encoded as 32-bit integers. 
Note that the aforementioned encoding scheme with single-precision floating-point format is fair to all methods.
For all methods, half-precision encoding can be used to uniformly reduce communication overheads presented in Table \ref{tab-overhead} by half.

\para{Communication overheads of baselines.} 
For these baselines, their per-round client-side communication costs are accounted as the total requirements of uploading and downloading the trainable parameters, since only this part of the parameters in the model is updated. 

\para{Communication overhead of \app.}
In our experiments, $K$ is uniformly set to 4,096 for \app. 
Let $D$, $U$ and $C$ denote the per-round communication cost of each client for the downlink, uplink, and total, respectively.
At the start of each round, each client needs to download the candidate seeds $\SSS$ and scalar gradient accumulator $\cA$ from the server. 
Since the $\SSS$ can be encoded as one integer seed which only occupies 4 Bytes, $D$ of \app can be calculated as
\begin{align*}
D &= \underbrace{1 \times 4 \ \text{Bytes}}_{\text{one integer seed that encodes the candidate seeds in $\SSS$}} + \underbrace{4096 \times 4 \ \text{Bytes}}_{\text{$\cA$ that contains 4096 accumulated scalar gradients}} \\
  &= 16388\ \text{Bytes}.
\end{align*}
After local training, each client return $\HH_i$ that contains $\tau$ pairs of $(s_j, \hat{\varrho}_j)$, with $\tau$=200 as described in Section \ref{subsec-exp-setup}, $U$ can be calculated as
\begin{align*}
U &= \underbrace{200 \times \underbrace{(2 \times 4 \ \text{Bytes})}_{\text{each $( s_j, \hat{\varrho}_j )$ pair}}}_{\text{$\HH_i$ that contains 200 pairs when $\tau$ = 200}} \\
  &= 1600\ \text{Bytes}.
\end{align*}
Finally, we can derive the total communication overhead required by each client in each round when applying \app, i.e., $C = D + U = 17988 \ \text{Bytes}$.

\para{Communication overhead of \apppro.}
As described in Section \ref{subsec-exp-setup}, we set $K$ to different values, i.e., 1,024 for \modeldatajuicer and 2,048 for \modelllama. 
For \emph{\apppro with \modeldatajuicer}, $D$ is calculated as
\begin{align*}
D &= \underbrace{1 \times 4 \ \text{Bytes}}_{\text{one integer seed that encodes the candidate seeds in $\SSS$}} + \underbrace{1024 \times 4 \ \text{Bytes}}_{\text{$\cA$ that contains 1024 accumulated scalar gradients}} + \underbrace{1024 \times 4 \ \text{Bytes}}_{\text{probabilities corresponding to the 1024 seeds}}\\
    &= 8196\ \text{Bytes}.
\end{align*}
Similarly to \app, in \apppro, each client is also required to upload only the gradient history, such that for each client $i$, we have
\begin{align*}
U &= \underbrace{200 \times \underbrace{(2 \times 4 \ \text{Bytes})}_{\text{each $( s_j, \hat{\varrho}_j )$ pair}}}_{\text{$\HH_i$ that contains 200 pairs when $\tau$ = 200}} \\
  &= 1600\ \text{Bytes}.
\end{align*}
Thus, we can derive the total communication overhead required per client per round in \emph{\apppro with \modeldatajuicer}, i.e., $C = D + U = 9796 \ \text{Bytes}$.

For \emph{\apppro with \modelllama}, $K$ is set to 2,048. 
Thus, we have
\begin{align*}
D &= \underbrace{1 \times 4 \ \text{Bytes}}_{\text{one integer seed that encodes the candidate seeds in $\SSS$}} + \underbrace{2048 \times 4 \ \text{Bytes}}_{\text{$\cA$ that contains 2048 accumulated scalar gradients}} + \underbrace{2048 \times 4 \ \text{Bytes}}_{\text{probabilities corresponding to the 2048 seeds}}\\
  &= 16388\ \text{Bytes}.
\end{align*}
\begin{align*}
U &= \underbrace{200 \times \underbrace{(2 \times 4 \ \text{Bytes})}_{\text{each $( s_j, \hat{\varrho}_j )$ pair}}}_{\text{$\HH_i$ that contains 200 pairs when $\tau$ = 200}} \\
  &= 1600\ \text{Bytes}.
\end{align*}
Thus, we have the total communication cost required by \emph{\apppro with \modelllama} for each client per round, i.e., $C = D + U = 17988 \ \text{Bytes}$.

%% file: sections/16_appendix_discussion.tex
\section{Extended Benefits in Real-world Applications}
\label{sec-appendix-discussions}
In this section, we provide discussions on more benefits brought by \app and \apppro to existing FL systems.

\subsection{Alleviating the Burden of Aggregation}
Assuming there are $m$ active clients in each round, traditional FL aggregation is conducted on the server with the computation and communication complexity of $\cO(m d)$, where $d$ is very large when the global model possesses a huge number of parameters, and $m$ is also large when there are many clients such as in cross-device FL \cite{chen2023fsreal,bai2024federated}.
Thus, the FL organizer usually needs to host an FL server with abundant computation and communication resources. 
In \app and \apppro, the computation and communication complexity of the server are both reduced to $\cO(mK)$. 
In this case, only a few computational and communication resources are required by the server, such that even a mobile device can handle it.
Consequently, the financial burden of FL organizers is greatly alleviated with \app and \apppro.

\subsection{Enabling Possibility to Decentralized Federated Fine-Tuning of LLMs}
Due to the transmission delays and unstable connections caused by long-distance transmission, many organizations opt for decentralized FL by allowing some clients to perform aggregation \cite{qin2024blockdfl}. 
However, it can exacerbate the communication costs of FL, as each client might be required to transmit its model parameters to multiple recipients. 
\app and \apppro can significantly reduce communication costs, thus bringing the possibility of fine-tuning LLMs with decentralized FL.

\subsection{Alleviating the Burden of Saving Checkpoints}
The trajectory of LLM fine-tuning does not always proceed in the desired direction. 
At times, fine-tuning may become trapped in a local optimum or an unfavorable region, or even be jeopardized by malicious attacks. 
Thus, multiple snapshots of LLMs often need to be stored during the fine-tuning of LLMs.
Besides, saving checkpoints may also contribute to snapshot ensemble \cite{huang2016snapshot}.
Snapshot saving of LLMs may incur a huge storage overhead. 
However, with \app and \apppro, only the snapshots of accumulated scalar gradients need to be preserved for the potential rollback of models, each of which is an array containing $2K$ scalars.
Therefore, it significantly reduces the storage consumption for model snapshots.